\documentclass[preprint,sts]{imsart-custom}
\usepackage[]{graphicx}
\usepackage[]{color}

\usepackage[colorlinks,citecolor=blue,urlcolor=blue]{hyperref}
\usepackage[utf8]{inputenc} 
\usepackage[T1]{fontenc}    

\usepackage{rotating}
\usepackage{subcaption}

\usepackage{placeins}

\usepackage[square,numbers]{natbib}

\usepackage{booktabs}       
\usepackage{amsfonts}       
\usepackage{microtype}      

\usepackage{url} 

\usepackage{amsmath}
\usepackage{wasysym}

\usepackage{silence}
\usepackage{amssymb}

\usepackage{float}
\usepackage{multirow}

\newcommand{\bd}{\begin{description} }
\newcommand{\ed}{\end{description} } 

\newcommand{\be}{\begin{enumerate}[a)]}
\newcommand{\ee}{\end{enumerate}}
\newcommand{\bc}{\begin{lstlisting}}
\newcommand{\ec}{\end{lstlisting}}

\DeclareMathOperator*{\Exp}{\mathbb{E}}
\newcommand{\PH}{{\hat{P}}}
\newcommand{\QH}{{\hat{Q}}}
\newcommand{\UH}{{\hat{U}}}
\newcommand{\disc}{\text{disc}}
\newcommand{\MMD}{\text{MMD}}

\DeclareMathOperator*{\argmin}{arg\,min}

\newtheorem{theorem}{Theorem}

\newtheorem{corollary}{Corollary}

\newtheorem{definition}{Definition}

\begin{document} 

\begin{frontmatter}

\title{Nuclear Discrepancy \\for Active Learning} 
	\author{Tom J. Viering} 
	\and
	\author{Jesse H. Krijthe}
  \and
  \author{Marco Loog}

 \address{T. J. Viering is with the Pattern Recognition Laboratory, Delft University of Technology, Delft, The Netherlands.  
 (e-mail: \href{mailto:t.j.viering@tudelft.nl}{{t.j.viering@tudelft.nl}}).
 J. H. Krijthe is with the Data Science Group, Radboud University Nijmegen, Nijmegen, The Netherlands. 
 (e-mail: \href{mailto:jkrijthe@gmail.com}{{jkrijthe@gmail.com}}).
 M. Loog is with the Pattern Recognition Laboratory, Delft University of Technology, Delft, The Netherlands and the Image Section, University of Copenhagen, Copenhagen, Denmark. 
 (e-mail: \href{mailto:m.loog@tudelft.nl}{{m.loog@tudelft.nl}}).
}

\begin{abstract}
Active learning algorithms propose which unlabeled objects should be queried for their labels to improve a predictive model the most. 
We study active learners that minimize generalization bounds and uncover relationships between these bounds that lead to an improved approach to active learning. 
In particular we show the relation between the bound of the state-of-the-art Maximum Mean Discrepancy (MMD) active learner, the bound of the Discrepancy, and a new and looser bound that we refer to as the Nuclear Discrepancy bound. 
We motivate this bound by a probabilistic argument: we show it considers situations which are more likely to occur. 
Our experiments indicate that active learning using the tightest Discrepancy bound performs the 
worst in terms of the squared loss. Overall, our proposed loosest Nuclear Discrepancy generalization bound performs the best. 
We confirm our probabilistic argument empirically: the other bounds focus on more pessimistic scenarios that are rarer in practice. We conclude that tightness of bounds is not always of main importance and that active learning methods should concentrate on realistic scenarios in order to improve performance.
\end{abstract}


\end{frontmatter}

\section{Introduction}
\label{intro}

Supervised machine learning models require enough labeled data to obtain good generalization performance. 
For many practical applications such as medical diagnosis or video classification it can be expensive or time consuming to label data \cite{SettlesNieuw}.
Often in these settings unlabeled data is abundant, but due to high costs only a small fraction can be labeled. 
In active learning an algorithm chooses unlabeled samples for labeling \citep{Cohn1994}. The idea is that models can perform better with less labeled data if the labeled data is chosen carefully instead of randomly. This way active learning methods make the most of a small labeling budget or can be used to reduce labeling costs.

A lot of methods have been proposed for active learning \citep{SettlesNieuw}, among which several works have used generalization bounds to perform active learning \citep{Gu2012,Gu2012aNieuw,Ganti2012a,Gu2014}. We perform a theoretical and empirical study of active learners, that choose queries that explicitly minimize generalization bounds, to investigate how the relation of the bounds impacts active learning performance.
We use the kernel regularized least squares model \citep{Rifkin2003} and the squared loss, a popular active learning setting \citep{Huang2010a,Wang2013Nieuw}.

We study the state-of-the-art Maximum Mean Discrepancy (MMD) active learner of \citet{Chattopadhyay2012Nieuw2} that minimizes a generalization bound \citep{Wang2013Nieuw}. The MMD is a divergence measure \citep{Gretton2012Nieuw} which is closely related to the Discrepancy measure of \citet{Mansour2009}, both have been used in domain adaptation \citep{GrettonNieuw,Cortes2014}. 
Using the Discrepancy, we show that we can get a tighter bound than the MMD in the realizable setting.
Tighter bounds are generally considered better as they estimate the generalization error more accurately. 
One might therefore also expect them to lead to better labeling choices in active learning and so we 
 introduce an active learner that minimizes the Discrepancy.  
We show using a probabilistic analysis, however, that the Discrepancy and MMD active learners optimize their strategies for unlikely scenarios. 
This leads us to introduce the Nuclear Discrepancy whose bound is looser. The Nuclear Discrepancy considers an average case scenario that may occur more often in practice.

We show empirically that active learning using the Nuclear Discrepancy improves upon the MMD and Discrepancy. In fact, active learning using the tightest Discrepancy bound performs the worst. 
We show experimentally that the scenarios considered by our bound occurs more often, explaining these counter-intuitive results. 
Our study shows that tighter bounds do not guarantee improved active learning performance and that a probabilistic analysis is essential.

The rest of this paper is organized as follows. In Section \ref{sec_theory} we introduce two existing generalization bounds, the MMD and Discrepancy, and we present several novel theoretical results. We give an improved MMD bound applicable to active learning and we show how to choose the MMD kernel (contribution 1). Under these conditions the MMD and Discrepancy bound become comparable, and we show that the Discrepancy bound is tighter (contribution 2). We use these theoretical results in Section \ref{sec_nucl} to analyze these existing bounds probabilistically. In this section we explain why tighter bounds may not lead to improved active learning performance (contribution 3). This probabilistic analysis leads to our novel looser Nuclear Discrepancy bound (contribution 4). In Section \ref{sec_exp} we benchmark the active learners on several datasets. We show that indeed our bound improves upon the Discrepancy and MMD for active learning, and we verify our probabilistic argument empirically by computing a novel error decomposition (contribution 5). In Section \ref{sec_discussion} we give a brief discussion and in Section \ref{sec_conclusion} we give the conclusions of this work. All proofs are given in the supplementary material. First, however, we introduce the setting and necessary notation.

\section{Setting and Notation}\label{sec_setting}

Let $\mathcal{X} = \mathbb{R}^d$ denote the input space and $\mathcal{Y}$ the output space. 
We assume there exists a deterministic labeling function $f : \mathcal{X} \rightarrow \mathcal{Y}$. We study the binary classification setting but all our results are applicable to regression as well.
We assume there is an unknown distribution $P$ over $\mathcal{X}$ from which we get an independent and identically distributed (i.i.d.) unlabeled sample $\PH$. 
Initially the labeled set $\QH$ is empty. The active learner selects (queries) samples sequentially from the unlabeled pool $\UH = \PH \setminus \QH$,
these samples are labeled and added to $\QH$. The samples are not removed from the set $\PH$, but the set $\UH$ is updated after each query. 
On $\QH$ a kernel regularized least squares model is trained. 
We indicate such a model using $h$ and its output on a sample $x \in \mathcal{X}$ using $h(x)$. 

We take the kernel $K$ of the model to be positive definite symmetric (PDS). For these kernels a reproducing kernel Hilbert space (RKHS) exists and we indicate the RKHS by $\mathcal{H}$. 
We use the overloaded notation where $h$ also indicates the corresponding vector in the RKHS of the model or function $h$.
The norm of a vector $h$ in the RKHS is written as $||h||_K$. 
We use $K(x,x')$ to indicate the kernel function between object $x$ and $x'$. 
In this work we use the Gaussian kernel $K(x,x') = \exp\left(-\frac{||x - x'||_2^2}{2\sigma^2}\right)$
where $\sigma$, the bandwidth, is a hyperparameter of the kernel.
For the computations of the MMD we need a second kernel which we indicate with $K'$ and we indicate its RKHS and bandwidth by $\mathcal{H}'$ and $\sigma'$, respectively. We use the convention that all vectors are column vectors. $X_{\hat{P}}$ and $X_{\hat{Q}}$ are the $n_\PH$ by $d$ and $n_\QH$ by $d$ matrices of the sets $\hat{P}$ and $\hat{Q}$. 
We use the same convention as \citet{Cortes2014}, where kernel regularized least squares minimizes the following objective for an hypothesis $h \in \mathcal{H}$ when trained on the sample $\hat{Q}$: $L_{\hat{Q}}(h,f) + \lambda ||h||_K^2$.  
We use $L_{\hat{Q}}(h,f)$ as shorthand for the mean squared error of $h$ on the sample $\QH$ with labels given by $f$: $\sum_{x \in \QH} (h(x) - f(x))^2$. 
The parameter $\lambda > 0$ is a regularization parameter which controls the complexity of the model. 
Similar to \citep{Cortes2014} we choose a subset of $\mathcal{H}$ as our hypothesis set: $H = \{ h \in \mathcal{H}: ||h||_K \leq \Lambda = \frac{f_{\text{max}}}{\sqrt{\lambda}}\}$, where $f_{\text{max}} = \max_{x \in \mathcal{X}}{f(x)}$. 
In \citep[Lemma 11.1]{mohri2012foundations} it is shown that training of the kernel regularized least squares model always leads to a solution $h \in H$. 

We focus on the squared loss since the bounds give direct guarantees on this performance measure, and because the quantities in the bounds can be computed in closed form for this loss. The goal for the active learner is to choose queries in such a way as to minimize the expected loss of the model: $L_P(h,f) = \int_{\mathcal{X}} (h(x)-f(x))^2 P(x) dx$. 
We would actually want to train our model on $\PH$, since if the model complexity is chosen appropriately, small loss on $\PH$ will lead to small expected loss on $P$\footnote{This holds even in the active learning setting, since the samples in $\PH$ are i.i.d. samples. If desired, a standard Rademacher complexity bound can be used to bound the expected loss on $P$ in terms of the loss on $\PH$ \citep{Generalized}.}. However, since we do not have labels for the samples in $\PH$, we upperbound the loss on $\PH$ instead. This upperbound is minimized by the active learners. The studied bounds are of the form $L_{\PH}(h,f) \leq L_{\QH}(h,f) + \text{obj}(\PH,\QH) + \eta$.

Due to training $L_{\QH}(h,f)$ will be relatively small. The term $\eta$ is a constant that cannot be minimized during the active learning process since it depends on the labels of the set $\PH$. However, if the model mismatch is small, $\eta$ will be small. Therefore we ignore this term during active learning, this is also (sometimes implicitly) done in other works \citep{GrettonNieuw,Chattopadhyay2012Nieuw2,Cortes2014}. The active learners will therefore minimize some objective $\text{obj}(\PH,\QH)$ sequentially. This objective can be the $\MMD$, $\disc$ or $\disc_N$ which will be introduced in the next sections. These objectives estimate the similarity between the samples $\PH$ and $\QH$ and do not depend on labels. Note that the resulting active learners will be non-adaptive: the active learning strategy is independent from the observed labels during active learning. 

Sequential minimization by the active learner is done as follows. 
The active learner forms a candidate set $\QH \cup s$ for each possible query $s$, and computes the objective for each candidate set. The query of the candidate set with minimal objective is chosen for labeling: $s^* = \argmin_{s \in \UH} \text{obj} (\PH, \QH \cup s)$. Note that constants $\Lambda$ or $\Lambda'$ do not influence the query selection for any of the objectives. 
We consider two settings. In the {realizable setting} the labeling function $f \in H$, thus a model of our hypothesis set generates the labels and there is no model misspecification. In this case $\mathcal{Y} = \mathbb{R}$. In the {agnostic setting} we use binary labels, thus $\mathcal{Y} = \{+1,-1\}$, and we generally have that $f \notin H$. 

\section{Theoretical Analysis of Existing Bounds} \label{sec_theory}

\paragraph{Improved MMD Bound for Active Learning.} The MMD measures the similarity between the two samples $\QH$ and $\PH$. Using this criterion we give a generalization bound similar to the one given in \citep{Wang2013Nieuw} that is suitable for active learning. 
We use the empirical MMD quantity, defined as 
\begin{equation} 
\text{MMD}(\hat{P},\hat{Q}) = \max_{\tilde{g} \in H'} \left( \frac{1}{n_\PH} \sum_{x \in \PH} \tilde{g}(x) - \frac{1}{n_\QH} \sum_{x \in \QH} \tilde{g}(x) \right).
\end{equation}
Here $\tilde{g}$ is the worst-case function from a set of functions $H'$. 
We take the set $H'$ as $H' = \{ h \in \mathcal{H}': ||h||_{K'} \leq \Lambda'\}$, where $K'$ is a yet to be specified PDS kernel with RKHS $\mathcal{H}'$. 
We note that the MMD can be computed in practice using Equation 3 in \citep{Gretton2012Nieuw}, where we have to multiply by $\Lambda'$ since we consider the general case where $\Lambda' \neq 1$. The equation is given in the supplementary material. 
By using the function $\tilde{g}$ to approximate the worst-case loss function we can prove the following novel bound: 
\begin{theorem}[MMD Generalization bound]\label{mmd_bound_lastig}
Let $L$ be any loss function, and let $g(x) = L(h(x),f(x))$. Then for any hypothesis $h \in H$,
\begin{equation}
L_{\hat{P}}(h,f) \leq L_{\hat{Q}}(h,f) + \MMD(\PH,\QH) + \eta_{\text{MMD}}, \label{EquationMMDBound}
\end{equation}
where $\eta_{\text{MMD}}$ is given by $\eta_{\text{MMD}} = 2 \min_{\tilde{g} \in H'} \max_{h \in H, x \in \PH} |g(x) - \tilde{g}(x)|$.
\end{theorem}
The term $\eta_{\text{MMD}}$ appears because we may have that $g \notin H'$. 
Our MMD bound differs in two aspects from the bound of \citet{Wang2013Nieuw}. In \citet{Wang2013Nieuw} the MMD is estimated between the distributions $P$ and $Q$. However, to estimate the MMD between distributions i.i.d. samples are required \citep[Appendix A.2]{Gretton2012Nieuw}. The samples of $\QH$ are not i.i.d. since they are chosen by an active learner. We avoid this by using the MMD for empirical samples. 
The second novelty is that we measure the error of approximating the loss function $g$ using the quantity $\eta_{\text{MMD}}$. 
This formulation allows us to adapt the MMD to take the hypothesis set and loss into account, similar to the Discrepancy measure of \citet{Cortes2014}.

\begin{theorem}[Adjusted MMD]\label{th_MMD_adjusted}
Let $L$ be the squared loss and assume $f \in H$ (realizable setting). If $K'(x_i,x_j) = K(x_i,x_j)^2$ and $\Lambda' = 4 \Lambda^2$, then it is guaranteed that $g \in H'$ and thus $\eta_{\text{MMD}} = 0$.
\end{theorem} 

\begin{corollary} \label{col_mmd}
Let $f \in H$ and let $K$ be a Gaussian kernel with bandwidth $\sigma$. If $K'$ is a Gaussian kernel with bandwidth $\sigma' = \frac{\sigma}{\sqrt{2}}$ and $\Lambda' = 4\Lambda^2$ then $\eta_{\text{MMD}} = 0$.
\end{corollary} 

Compared to other works Theorem \ref{th_MMD_adjusted} gives a more principled way to choose the MMD kernel in the context of learning\footnote{Note that the MMD can also be used to determine whether or not two sets of samples are from the same distribution \citep{Gretton2012Nieuw}.}. %
Previously, often a Gaussian kernel was used for the MMD with $\sigma' = \sigma$. %
In particular, Corollary \ref{col_mmd} shows that if our model uses $\sigma$ as bandwidth and $\sigma' = \sigma$, we may have that $\eta_{\text{MMD}} \neq 0$ even in the realizable setting $f \in H$, since $\sigma'$ is too large. This is undesirable since $\eta_{\text{MMD}}$ cannot be minimized during active learning. Therefore our choice for $\sigma'$ which guarantees that $\eta_{\text{MMD}} = 0$ in the realizable setting is preferable. 

\paragraph{Discrepancy Bound for Active Learning. } We give a bound of \citet{Generalized} in terms of the Discrepancy.
The Discrepancy is defined as 
\begin{equation}
\text{disc}(\hat{P},\hat{Q}) = \max_{h,h' \in H} |L_{\hat{P}}(h',h) - L_{\hat{Q}}(h',h)|.
\end{equation}
Observe that the Discrepancy depends directly on the loss $L$ and the hypothesis set $H$. 

\begin{theorem}[Discrepancy generalization bound] \label{theorem_disc_bound}
Assume that for any $x \in \mathcal{X}$ and $h \in H$ that $L(h(x),f(x)) \leq C$ and that $L$ is the squared loss. 
Then given any hypothesis $h \in H$,
\begin{equation}
L_{\hat{P}}(h,f) \leq L_{\hat{Q}}(h,f) + \text{disc}(\hat{P},\hat{Q}) + \eta_{\text{disc}}, \label{EquationDiscBound}
\end{equation}
where $\eta_{\text{disc}}$ is given by $\eta_{\text{disc}} = 4 C \min_{\tilde{f} \in H} \max_{x \in \PH} |\tilde{f}(x) - f(x)|$.
\end{theorem}
Here $\eta_{\text{disc}}$ measures the model misspecification. If $f \in H$ the term $\eta_{\text{disc}}$ becomes zero. 

\paragraph{Eigenvalue Analysis. } The matrix $M$ is given by 
\begin{equation}
M = \frac{1}{n_\PH} X_\PH^T X_\PH - \frac{1}{n_\QH} X_\QH^T X_\QH. \label{EquationdefM}
\end{equation}
We study the bounds of the Discrepancy and the MMD using the eigenvalues of matrix $M$. This analysis is novel for the MMD and allows us to show that the Discrepancy bound is tighter. Furthermore, we need this analysis in the next section to motivate the Nuclear Discrepancy. 
\citet{Mansour2009} show that the Discrepancy, in terms of the eigenvalues $\lambda_i$ of $M$, is given by 
\begin{equation}
\disc(\hat{P},\hat{Q}) = 4 \Lambda^2 \max_i |\lambda_i| \equiv 4 \Lambda^2 \lambda_1 \label{EquationDiscComp}.
\end{equation}
in case $K$ is the linear kernel\footnote{The Discrepancy can also be computed for any arbitrary kernel by replacing $M$ by $M_K$ \citep{Cortes2014}, see the supplementary material for more details. All our theoretical results that follow are applicable to both $M$ and $M_K$ and are thus applicable to any kernel. For simplicity we use $M$ in the main text.}.
From this point on we assume that the eigenvalues $\lambda_i$ of $M$ are sorted by absolute value, where $\lambda_1$ is the largest absolute eigenvalue. 
The next original theorem shows how the MMD can be computed using $M$.
\begin{theorem}
Under the conditions of Theorem \ref{th_MMD_adjusted} we have 
\begin{equation}
\MMD(\PH,\QH) = 4 \Lambda^2 \sqrt{\sum_i \lambda_i^2} \label{EquationMMDM},
\end{equation}
\end{theorem}
In this case, $\eta_{\text{MMD}} = \eta_{\text{disc}} = 0$, and by comparing Equation \ref{EquationDiscComp} and Equation \ref{EquationMMDM} we can show that  $\disc(\PH,\QH) \leq \MMD(\PH,\QH)$. 
Thus the Discrepancy bound (Theorem \ref{theorem_disc_bound}) is tighter than the MMD bound (Theorem \ref{mmd_bound_lastig}) under the conditions of Theorem \ref{th_MMD_adjusted}.  
Since the Discrepancy bound is tighter, one can argue that it estimates the expected loss more accurately than the MMD bound. Therefore, one may expect that active learning by minimization of the Discrepancy may result in better active learning queries  than minimization of the MMD. 

\section{Nuclear Discrepancy} \label{sec_nucl}

Though it may seem obvious to expect better performance in active learning when tighter bounds are used, the probabilistic analysis given in this section indicates that the Discrepancy will perform worse than the MMD. This, in turn, will lead us to introduce the Nuclear Discrepancy. Our analysis suggests that this bound will improve upon the MMD and the Discrepancy when used for active learning. To start with, we require the following decomposition of the error:

\begin{theorem}
If $f,h \in H$, $L$ is the squared loss, and $L_{\QH}(h,f) \approx 0$ ($h$ is trained on $\QH$), then
\begin{equation}
L_{\PH}(h,f) \approx |u^T M u| = |\sum_i \bar{u}_i^2 \lambda_i|, \label{EquationErrorDecomposition}
\end{equation}
where $u = h-f$, and where $\bar{u}_i$ is the projection of $u$ on the normalized $i$th eigenvector of $M$. Note that $\sqrt{\sum_i \bar{u}_i^2} \leq 2\Lambda$.
\end{theorem}\label{th_errorDecomposition}
Observe that in the above theorem, the components of $\bar{u}_i^2$ weigh the contribution of each eigenvalue $\lambda_i$ to the error. Essentially, this means that if $\bar{u}$ points more in the direction of the eigenvector belonging to eigenvalue $\lambda_i$, this eigenvalue will contribute more to the error.

Active learning by minimization of the Discrepancy minimizes the largest absolute eigenvalue (Equation \ref{EquationDiscComp}). 
In view of the error decomposition above, we can see that the Discrepancy always considers the scenario where $u$ points in the direction of the eigenvector with largest absolute eigenvalue. One should realize that this is a very specific scenario, where all components of $\bar{u}_{i\neq1}$ are zero, that is very unlikely to occur. This is a worst case scenario because this maximizes the right hand side of Equation \ref{EquationErrorDecomposition} with respect to $u$. 

The MMD considers a less pessimistic and a more realistic scenario. The MMD active learner minimizes the squared eigenvalues (Equation \ref{EquationMMDM}) and thus assumes all eigenvalues contribute to the error. However, the MMD is biased towards minimizing large absolute eigenvalues because the eigenvalues are squared. This suggests that the MMD assumes that $u$ is more likely to point in the direction of eigenvectors with large absolute eigenvalues, since the objective indicates these eigenvalues are more important to minimize.  
This is also in a sense pessimistic, since large absolute eigenvalues can contribute more to the error. However, this scenario is less unlikely than the scenario considered by the Discrepancy. Because the MMD active learner optimizes its strategy for a scenario that we expect to occur more often in practice, we expect it to improve upon the Discrepancy.

In light of the foregoing, we now propose the more optimistic assumption that any $u$ is equally likely. This assumption is not true generally since $h$ is the result of a minimization problem. However, we expect this average case scenario to better reflect reality than always assuming a pessimistic scenario like the MMD where $u$ points in the direction of large eigenvalues. An active learner that optimizes its strategy for this more realistic scenario will therefore likely improve upon the MMD. Our optimistic assumption leads to the following theorem. 
\begin{theorem}[Probabilistic generalization bound]
Let $f,h \in H$ and $u = h-f$. Then $u \in U = \{u \in \mathcal{H} : ||u||_K \leq 2\Lambda\}$. Assuming each $u \in U$ is equally likely we can show that:
\begin{equation}
\Exp_{u \in U} L_\PH(h,f) \leq \Exp_{u \in U} L_\QH(h,f) + \frac{4}{3}\Lambda^2 \sum_i |\lambda_i| \label{EquationBoundProbabilistic},
\end{equation}
where $\lambda_i$ are the eigenvalues of the matrix $M$ (Equation \ref{EquationdefM}).
\end{theorem}
This bound indicates that we should minimize all absolute eigenvalues if all $u$ are equally likely to occur.
Inspired by this analysis, we define the Nuclear Discrepancy quantity as $\text{disc}_N(\PH,\QH) = 4 \Lambda^2 \sum_i |\lambda_i|$.
This is proportional to the so-called nuclear matrix norm of $M$. 
Observe that the Nuclear Discrepancy upperbounds the MMD (Equation \ref{EquationMMDM}) and the Discrepancy (Equation \ref{EquationDiscComp}). By upperbounding the Discrepancy in Equation \ref{EquationDiscBound} we obtain the following deterministic bound.

\begin{theorem}[Deterministic Nuclear Discrepancy bound] \label{theorem_disc_bound_nucl}
Assume that for any $x \in \mathcal{X}$ and $h \in H$ that $L(h(x),f(x)) \leq C$ and that $L$ is the squared loss. 
Then given any hypothesis $h \in H$,
\begin{equation}
L_{\hat{P}}(h,f) \leq L_{\hat{Q}}(h,f) + \text{disc}_N(\hat{P},\hat{Q}) + \eta_{\text{disc}}, \label{EquationDiscNuclearBound}
\end{equation}
where $\eta_{\text{disc}}$ is given in Theorem \ref{theorem_disc_bound}.
\end{theorem}

The Nuclear Discrepancy bound is looser in the realizable setting than the MMD and Discrepancy bounds. Yet the bound is more optimistic since it takes an average case scenario into account instead of an unlikely pessimistic scenario. These average case scenarios might occur more often in practice, and therefore we expect the Nuclear Discrepancy to improve upon the MMD and Discrepancy when minimized for active learning, since the Nuclear Discrepancy takes these scenarios explicitly into account in its strategy.

\section{Experiments} \label{sec_exp}

\paragraph{Experimental Setup and Baselines.}
A training set ($65\%$) and test set ($35\%$) are used in each experiment. The training set corresponds to $\PH$. $\QH$ is initially empty. 
After each query, the labeled set $\QH \subseteq \PH$ is updated and the model is trained and evaluated on the test set in terms of the mean squared error (MSE). 
We use the active learners to sequentially select 50 queries.
As baseline we use random sampling and a sequential version of the state-of-the-art MMD active learner \citep{Chattopadhyay2012Nieuw2,Wang2013Nieuw}. 
We compare the baselines with our novel active learners: the Discrepancy active learner and the Nuclear Discrepancy active learner. 

The active learning methods are evaluated on 12 datasets. Some datasets are provided by \citet{Cawley2004} and the other datasets originate from the UCI Machine Learning repository \citep{UCI}. See the supplementary material for the dataset characteristics. 
Similar to \citet{Huang2010a} we convert multiclass datasets into two class datasets. 
To ease computation times we subsampled datasets to contain a maximum of 1000 objects. All features were normalized to have zero mean and a standard deviation of one.  
To make datasets conform to the realizable setting we use the approach of \citet{Cortes2014}: we fit a model of our hypothesis set to the whole dataset and use its outputs as labels. 

To set reasonable hyperparameters, we repeat the following procedure multiple times. {This procedure makes sure the $\eta$'s in the bounds are small and that the model complexity is appropriate.} We randomly select $25$ examples from the dataset and label these. We train a model on these samples and evaluate the MSE on all unselected objects. The hyperparameters that result in the best performance after averaging are used in the active learning experiments. For reproducibility we give all hyperparameter settings in the supplementary material. 
The procedure above leads to the hyperparameter $\sigma$ of the Gaussian kernel and the regularization parameter $\lambda$ of the model. 
We set the hyperparameter of the Gaussian kernel of the MMD according to our analysis in Corollary \ref{col_mmd} as $\sigma' = \frac{\sigma}{\sqrt{2}}$. 

\paragraph{Realizable Setting.} \label{subsec_realizeable}
First we benchmark our proposed active learners in the realizable setting. In this setting we are assured that $\eta = 0$ in all bounds and therefore we eliminate effects that arise due to model misspecification. 
Also the relations between the bounds are guaranteed: the Discrepancy bound is the tightest, and the Nuclear Discrepancy bound is the loosest. 

\begin{figure}[tbh]
		\begin{subfigure}{0.48\textwidth}
        \includegraphics[width=\textwidth]{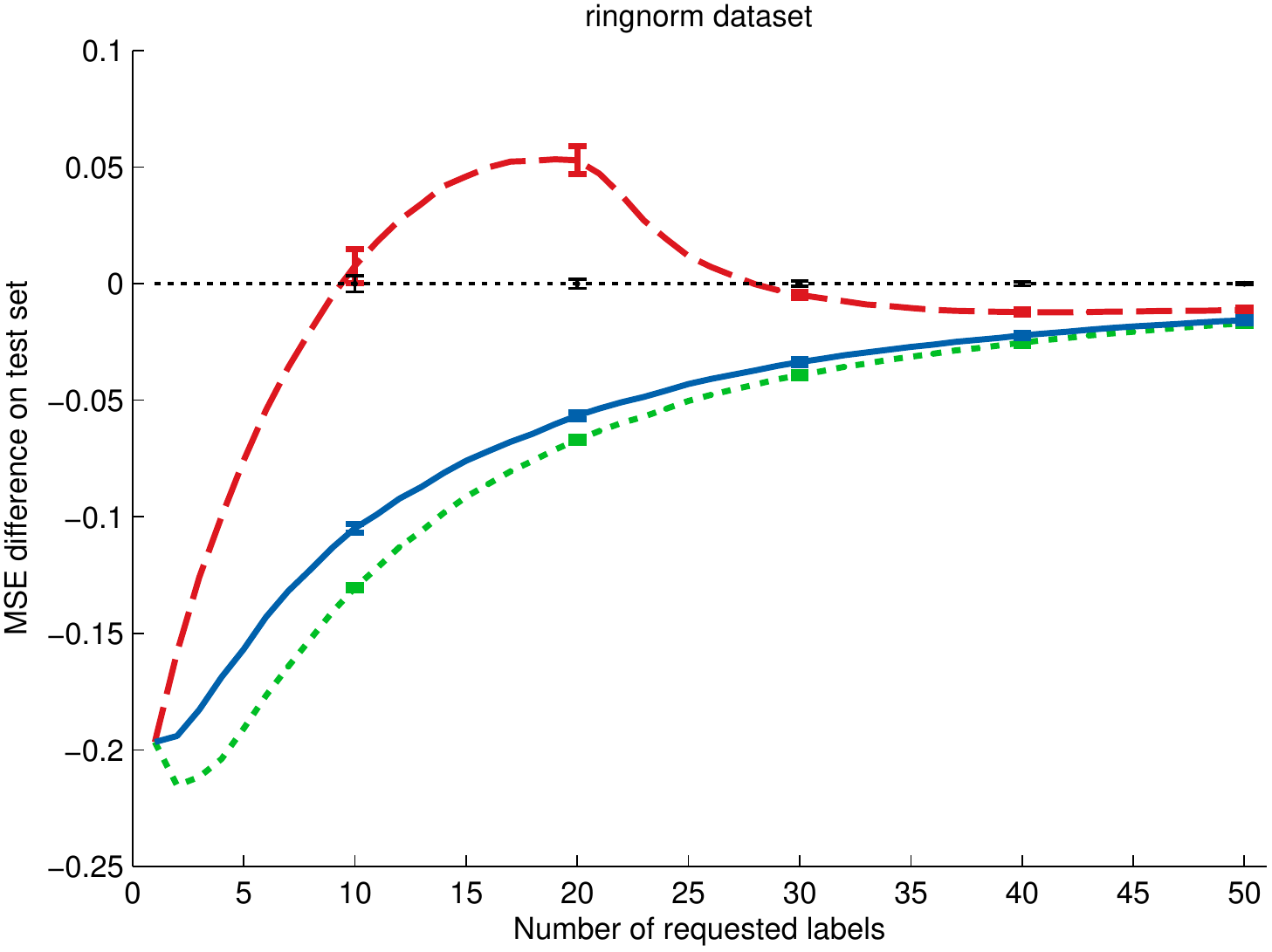}
    \end{subfigure}
		\begin{subfigure}{0.48\textwidth}
        \includegraphics[width=\textwidth]{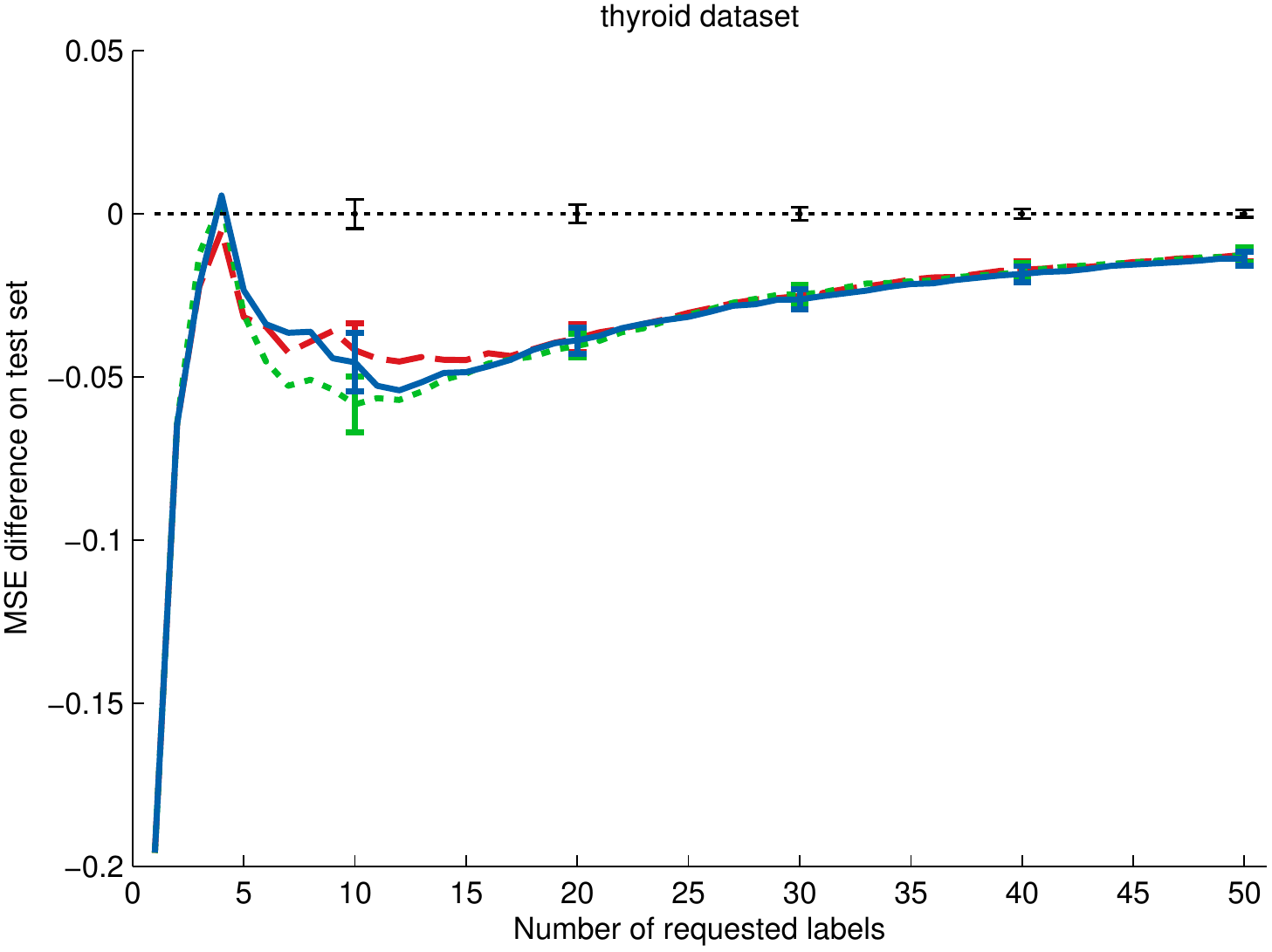}
    \end{subfigure}
		\begin{subfigure}{0.48\textwidth}
        \includegraphics[width=\textwidth]{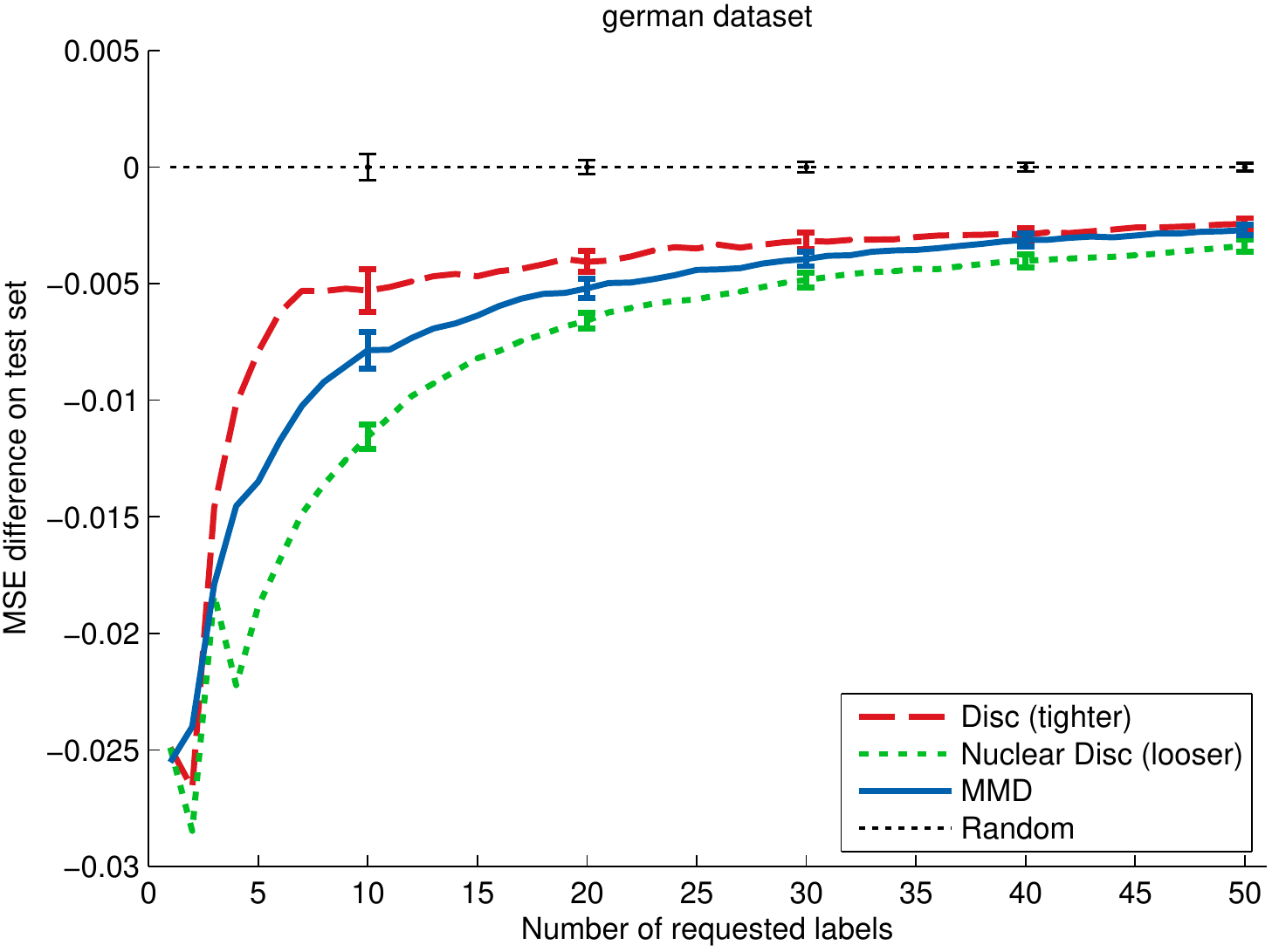}
    \end{subfigure}
    \caption{\small{Averaged learning curves over 100 runs in the realizable setting. The MSE is measured with respect to random sampling (lower is better). The Discrepancy with the tightest bound often performs the worst and the Nuclear Discrepancy with the loosest bound often performs the best.}}\label{fig_lc_realizeable}
\end{figure}

We plot several learning curves in Figure \ref{fig_lc_realizeable}.
The MSE of the active learner minus the mean performance (per query) of random sampling is displayed on the y-axis (lower is better). The curve is averaged over 100 runs. 
Error bars represent the $95\%$ confidence interval computed using the standard error. 
We summarize our results on all datasets in Table \ref{table_perf_realizeable} (see page \pageref{table_perf_realizeable}) as is usual in active learning \citep{Wang2013Nieuw,Gu2014}. To this end, each time after 5 queries, we compute a two tailed t-test (significance level $p = 0.05$) comparing the 200 MSE results of two active learning methods. If an active learner improves significantly upon another in terms of MSE we count this as a win, if there is no significant difference we count it as a tie, and if it performs significantly worse we count it as a loss. 

Observe that the Discrepancy active learner, with the tightest bound, in the majority of the cases performs worst. Especially the results on the ringnorm dataset are remarkable, here the Discrepancy performs worse than random sampling. In the majority of the cases the proposed Nuclear Discrepancy, with the loosest bound, indeed improves upon the MMD and the Discrepancy active learners. 

\newcommand{\datasetformat}[1]{{#1}}

\paragraph{Error Decomposition.}
In the realizable setting we have the advantage that we know the labeling function $f$. This allows us to compute the error decomposition of Equation \ref{EquationErrorDecomposition} explicitly. See the supplementary material for details how to compute the decomposition in case kernels are used. 
Using this decomposition we can explain the differences in performance between the active learners.

\begin{figure}[tbh]
    \begin{subfigure}{0.48\textwidth}
        \includegraphics[width=\textwidth]{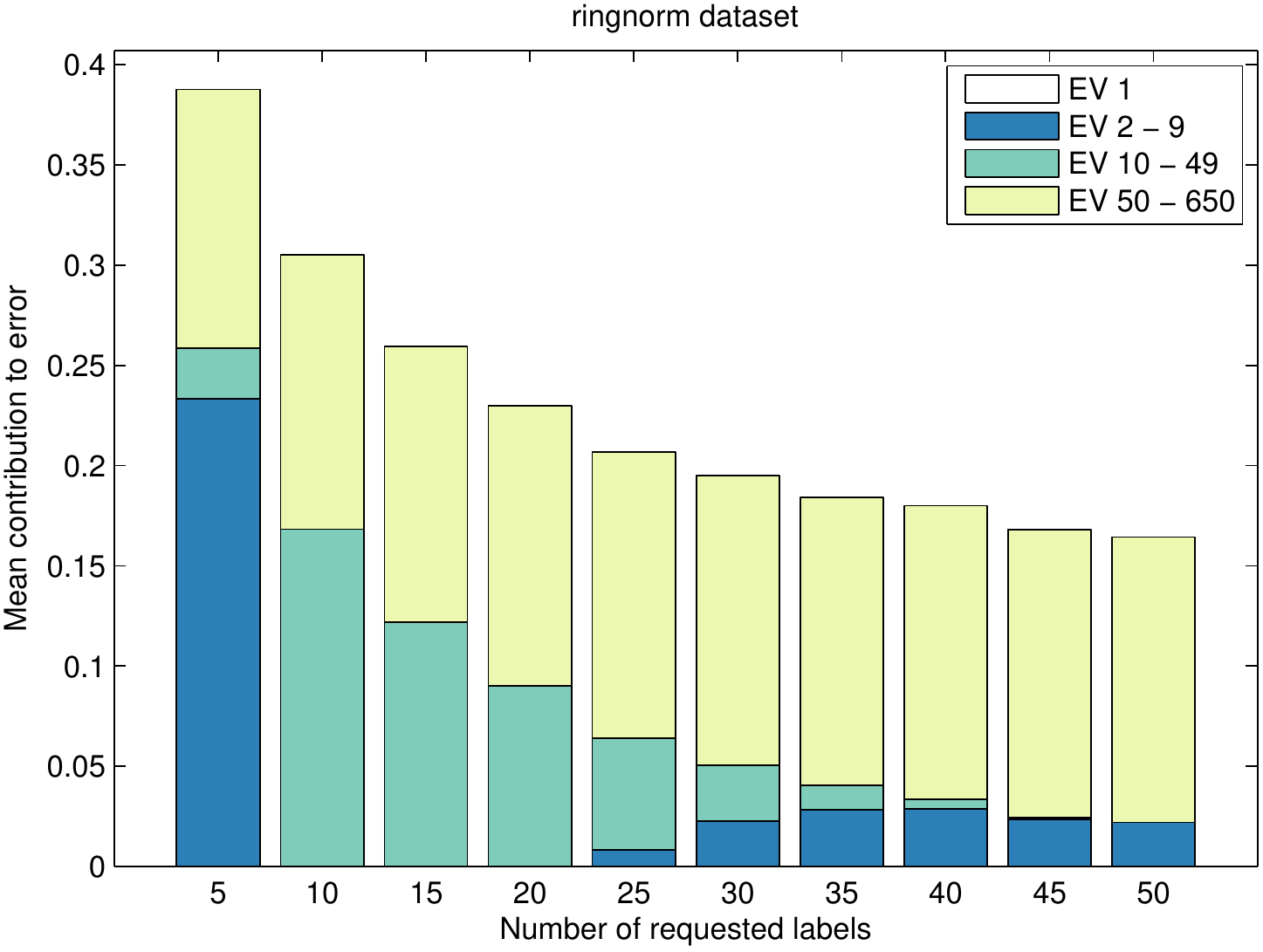}
    \end{subfigure}
		\begin{subfigure}{0.48\textwidth}
        \includegraphics[width=\textwidth]{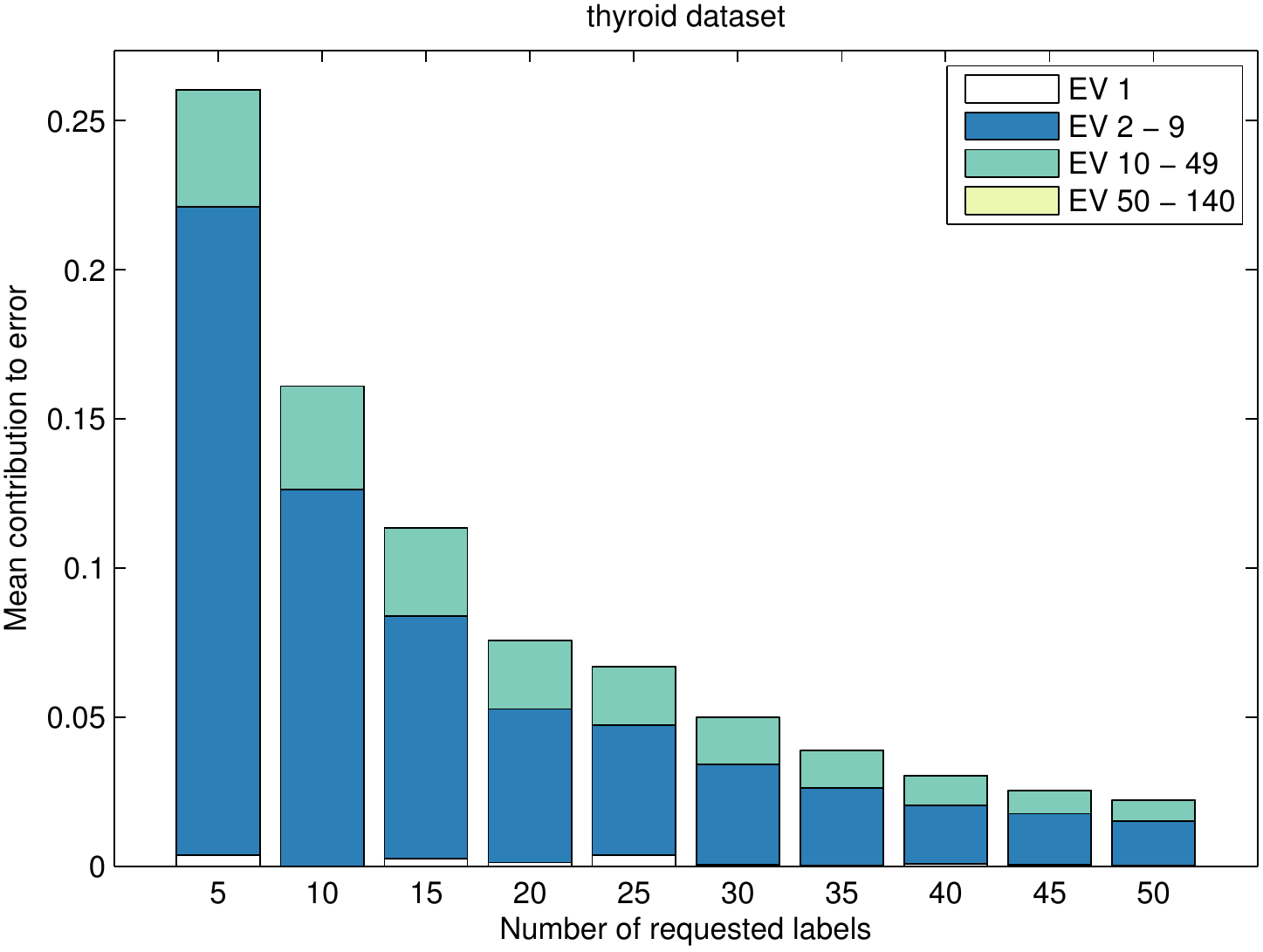}
    \end{subfigure}
		\begin{subfigure}{0.48\textwidth}
        \includegraphics[width=\textwidth]{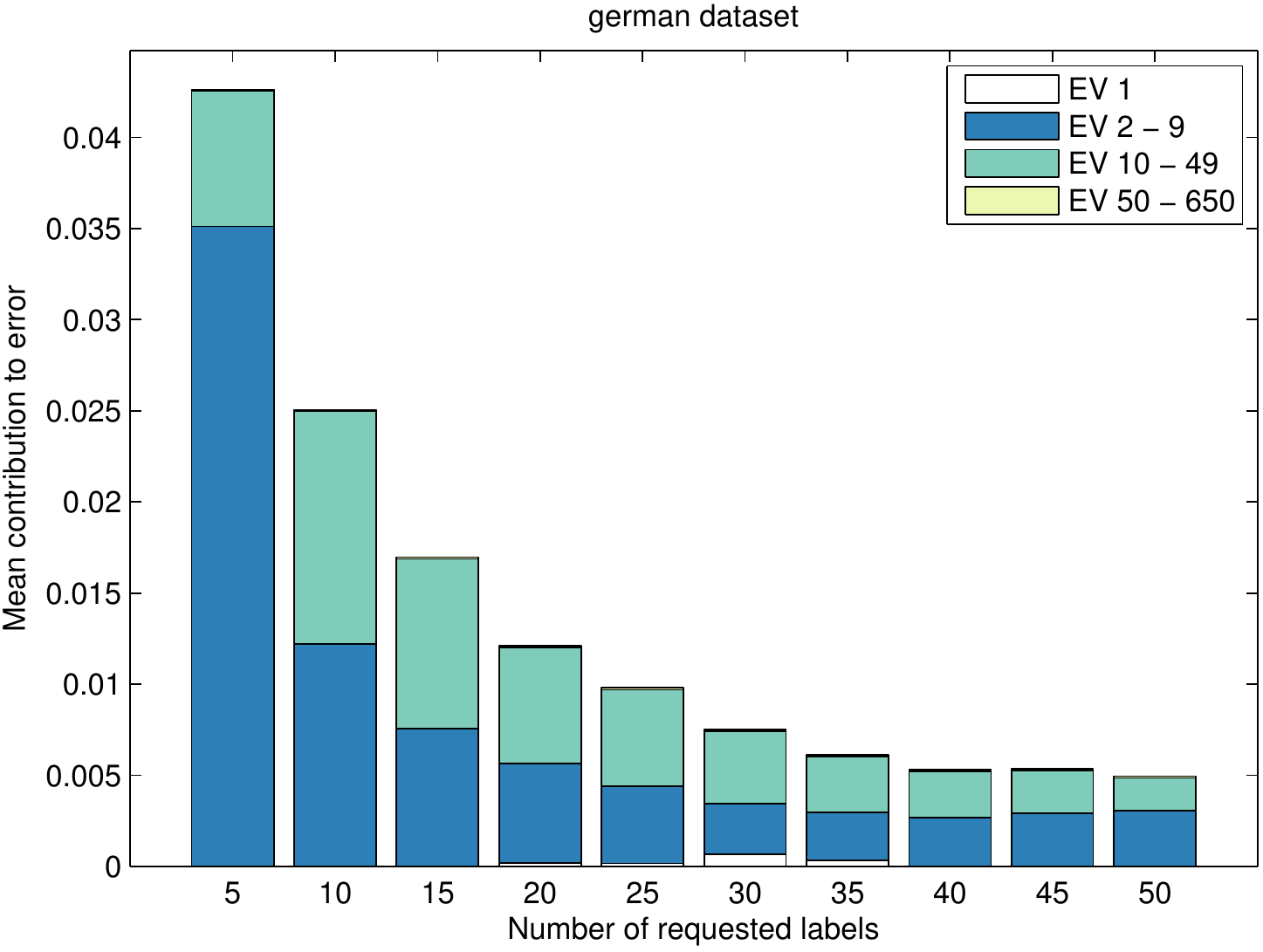}
    \end{subfigure}
    \caption{\small{Error decomposition during active learning. We plot the mean contribution of the absolute largest eigenvalue (EV1), $\bar{u}_1^2 \lambda_1$, and the mean summed contribution of several other sets of eigenvalues to $\PH$ (see Equation \ref{EquationErrorDecomposition}) over all 100 runs of the random active learner. The eigenvalues are ordered by absolute size. Observe that the absolute largest eigenvalue in some cases contributes little or nothing and that in general all eigenvalues contribute to the error. This confirms that the Nuclear Discrepancy considers more realistic scenarios.}}\label{fig_lc_umu}
\end{figure}

In Figure \ref{fig_lc_umu} we show this decomposition of the error using a stacked bar chart during several active learning experiments of the baseline active learner `Random'\footnote{The graphs for other active learners are qualitatively similar and for brevity we do not show them here.}. 
Recall that the eigenvalues of the matrix $M$ are sorted by absolute size, here we use EV1 to indicate the absolute largest eigenvalue. The contribution of EV1 is given by: $\bar{u}_1^2\lambda_1$ (see also Equation \ref{EquationErrorDecomposition}). We use the notation EV 2 - 9 to indicate the summed contribution of the eigenvalues: $\sum_{i=2}^9 \bar{u}_i^2\lambda_i$. The mean contribution over 100 runs is shown.

Observe that the contribution of the absolute largest eigenvalue to the error in practice often is extremely small: the bar of EV1 is hardly visible in Figure \ref{fig_lc_umu}. Note: EV1 is represented by the white bar that starts at the bottom which is only visible for the datasets thyroid and german. 
Therefore the Discrepancy active learner chooses suboptimal samples: its strategy is optimized for a worst-case scenario that is very rare in practice.
The MMD improves upon the Discrepancy since we observe that the scenario of the MMD is more likely. 
However, observe that small absolute eigenvalues can contribute substantially to the error, this is especially clear for the ringnorm dataset where EV 50 - 650 contribute the most to the error after 10 queries. 
In practice we did not find evidence that larger absolute eigenvalues are likely to contribute more to the error. 
This confirms our argument why the Nuclear Discrepancy can improve upon the MMD: the scenario considered by the Nuclear Discrepancy is more likely to occur in practice. 


\paragraph{Agnostic Setting.}
The experiments in the realizeable setting dealt with a somewhat artificial setting without model mismatch. Now we discuss the results of the agnostic setting where the original binary labels of the datasets are used. In this setting $\eta \neq 0$, but $\eta$ will be small due to our choice of hyperparameters, and therefore we ignore it during active learning. 
All results are summarized in Table \ref{table_perf_realizeable} (see page \pageref{table_perf_realizeable}). The learning curves are quite similar to the realizable setting, therefore we defer them to the supplementary material.
One notable difference is that the learning curves are less smooth and have larger standard errors. 
Therefore the active learning methods are harder to distinguish which is reflected in Table \ref{table_perf_realizeable} by more ties. 
However, the trends observed in the realizable setting are still observed in this setting: the Nuclear Discrepancy active learner improves more often on the MMD than the reverse, and the MMD improves more often upon the Discrepancy than the reverse. 

{
\begin{sidewaystable}[p]
  \vskip -0.1in
	\caption{\small{Win / tie / loss counts comparing the MMD, Discrepancy (D) and Nuclear Discrepancy (ND) active learners in the realizable and agnostic setting. Observe that the Discrepancy (tightest bound) often performs the worst, and that the proposed Nuclear Discrepancy (loosest bound) often performs the best.}} 
		\label{table_perf_realizeable} 
		\vskip 0.05in
		\small
		\centering

         \begin{tabular}{rccccccc}
					    \toprule
		                             & \multicolumn{3}{c}{Realizeable setting}                       & & \multicolumn{3}{c}{Agnostic setting}\\ \cmidrule{2-4}\cmidrule{6-8}
                             Dataset & D vs MMD            & ND vs D           & ND vs MMD       & & D vs MMD       & ND vs D           & ND vs MMD   \\
														 \midrule
							 
            \datasetformat{vehicles} & 0 / 10 / 0          & 0  / 10 / 0       & 0  / 10 / 0     & & 3 / 7  / 0     & 0 / 9  / 1        & 1 / 9  / 0  \\
               \datasetformat{heart} & 0 / 3  / 7          & 10 / 0  / 0       & 7  / 3  / 0     & & 0 / 10 / 0     & 0 / 10 / 0        & 0 / 10 / 0  \\
               \datasetformat{sonar} & 0 / 1  / 9          & 10 / 0  / 0       & 4  / 6  / 0     & & 0 / 3  / 7     & 8 / 2  / 0        & 3 / 7  / 0  \\
                                                                                                 
             \datasetformat{thyroid} & 0 / 10 / 0          & 1  / 9  / 0       & 1  / 9  / 0     & & 0 / 10 / 0     & 2 / 8  / 0        & 2 / 8  / 0  \\
            \datasetformat{ringnorm} & 0 / 0  / 10         & 10 / 0  / 0       & 10 / 0  / 0     & & 0 / 1  / 9     & 7 / 0  / 3        & 1 / 1  / 8  \\
          \datasetformat{ionosphere} & 0 / 0  / 10         & 10 / 0  / 0       & 10 / 0  / 0     & & 1 / 3  / 6     & 0 / 8  / 2        & 0 / 7  / 3  \\
            \datasetformat{diabetes} & 0 / 9  / 1          & 4  / 6  / 0       & 5  / 3  / 2     & & 1 / 9  / 0     & 0 / 7  / 3        & 0 / 8  / 2  \\
             \datasetformat{twonorm} & 0 / 1  / 9          & 10 / 0  / 0       & 10 / 0  / 0     & & 0 / 7  / 3     & 5 / 5  / 0        & 7 / 3  / 0  \\
              \datasetformat{banana} & 0 / 3  / 7          & 7  / 3  / 0       & 0  / 10 / 0     & & 2 / 8  / 0     & 2 / 8  / 0        & 6 / 4  / 0  \\
              \datasetformat{german} & 0 / 1  / 9          & 10 / 0  / 0       & 10 / 0  / 0     & & 1 / 9  / 0     & 1 / 9  / 0        & 2 / 8  / 0  \\
              \datasetformat{splice} & 1 / 9  / 0          & 9  / 1  / 0       & 8  / 2  / 0     & & 0 / 8  / 2     & 6 / 4  / 0        & 3 / 7  / 0  \\
              \datasetformat{breast} & 1 / 0  / 9          & 10 / 0  / 0       & 10 / 0  / 0     & & 0 / 6  / 4     & 2 / 8  / 0        & 1 / 9  / 0  \\
							   \midrule
						     Summary             & 2 / 47 / 71         & 91 / 29 / 0       & 75 / 43 / 2     & & 8 / 81 / 31    & 33 / 78 / 9       & 26 / 81 / 13 \\ 
							     \bottomrule
			\end{tabular}    
			\vskip -0.1in
\end{sidewaystable}
}
\FloatBarrier
\section{Discussion}\label{sec_discussion}

A first issue raised by this work is the following. Where the experiments in the realizable setting provide clear insights, the results concerning the agnostic setting are certainly not fully understood. Studying the approximation errors $\eta$ in the bounds may offer further insight. But such study is not trivial since the behavior of $\eta$ in this setting depends on the precise model mismatch and the setting of the hyperparameters. 

 A second issue of interest is whether our Nuclear Discrepancy bound can be helpful in other settings as well.
\citet{Ben-David2010} give the Discrepancy bound for the zero-one loss. Given our results we suspect this bound can be too pessimistic as well. A Nuclear Discrepancy type bound for the zero-one loss is therefore desirable. However such a bound is non-trivial to compute since it will involve an integral over the zero-one loss function, and therefore we defer this to future work.

 Aside for the implications for active learning, our results have implications for domain adaptation as well.
\citet{Cortes2014} observe that the Discrepancy outperforms the MMD in adaptation. We suspect that the MMD can improve for our suggested choice of $\sigma'$. Furthermore, our results suggest that the Nuclear Discrepancy is a promising bound for adaptation, however it poses a non-trivial optimization problem which we plan to address in the future. 

 Finally, in this work we have only considered non-adaptive active learners. Adaptive active learning strategies use label information to choose queries, this additional information may lead to improved performance. For domain adaptation, \cite{Generalized} considers such an adaptive approach to improve upon the Discrepancy. Their bound is, however, not trivial to apply to active learning because it is intrinsically designed for domain adaptation. Nevertheless, any successful adaptation of our approach in such direction would certainly be valuable.

\section{Conclusion} \label{sec_conclusion}

We proposed two novel active learning methods based on generalization bounds and compared them to the state-of-the-art MMD active learner. 
To investigate the relation between the bounds and active learning performance, we have shown that the Discrepancy bound is the tightest bound and our proposed Nuclear Discrepancy is the loosest.

 Even though the Discrepancy bound is tighter, the active learner performs worse compared to the MMD active learner, even in the realizable setting. Our proposed Nuclear Discrepancy, which has the loosest bound, improves significantly upon both the MMD and Discrepancy in terms of the mean squared error in the realizable setting.

 We explain this counter-intuitive result by showing that the Discrepancy and MMD focus too much on pessimistic scenarios that are unlikely to occur in practice. On the other hand, the Nuclear Discrepancy considers an average case scenario, which occurs more often, and therefore performs better. 
We show that a probabilistic approach is essential: active learners should optimize their strategy for scenarios that are likely to occur in order to perform well. 

\newpage

\small
\bibliography{librarythesis}

\begin{thebibliography}{20}
\providecommand{\natexlab}[1]{#1}
\providecommand{\url}[1]{\texttt{#1}}
\expandafter\ifx\csname urlstyle\endcsname\relax
  \providecommand{\doi}[1]{doi: #1}\else
  \providecommand{\doi}{doi: \begingroup \urlstyle{rm}\Url}\fi

\bibitem[Ben-David et~al.(2010)Ben-David, Blitzer, Crammer, Kulesza, Pereira,
  and Vaughan]{Ben-David2010}
Shai Ben-David, John Blitzer, Koby Crammer, Alex Kulesza, Fernando Pereira, and
  Jennifer~Wortman Vaughan.
\newblock A theory of learning from different domains.
\newblock \emph{Machine Learning}, 79\penalty0 (1):\penalty0 151--175, 2010.

\bibitem[Cawley and Talbot(2004)]{Cawley2004}
Gavin~C. Cawley and Nicola~L.C. Talbot.
\newblock Fast exact leave-one-out cross-validation of sparse least-squares
  support vector machines.
\newblock \emph{Neural Networks}, 17\penalty0 (10):\penalty0 1467 -- 1475,
  2004.

\bibitem[Chattopadhyay et~al.(2012)Chattopadhyay, Wang, Fan, Davidson,
  Panchanathan, and Ye]{Chattopadhyay2012Nieuw2}
Rita Chattopadhyay, Zheng Wang, Wei Fan, Ian Davidson, Sethuraman Panchanathan,
  and Jieping Ye.
\newblock {Batch Mode Active Sampling Based on Marginal Probability
  Distribution Matching}.
\newblock In \emph{Proceedings of the 18th ACM SIGKDD International Conference
  on Knowledge Discovery and Data Mining (KDD)}, pages 741--749, 2012.

\bibitem[Cohn et~al.(1994)Cohn, Atlas, and Ladner]{Cohn1994}
David Cohn, Les Atlas, and Richard Ladner.
\newblock {Improving generalization with active learning}.
\newblock \emph{Machine Learning}, 15\penalty0 (2):\penalty0 201--221, 1994.

\bibitem[Cortes and Mohri(2014)]{Cortes2014}
Corinna Cortes and Mehryar Mohri.
\newblock {Domain adaptation and sample bias correction theory and algorithm
  for regression}.
\newblock \emph{Theoretical Computer Science}, 519:\penalty0 103--126, 2014.

\bibitem[Cortes et~al.(forthcoming)Cortes, Mohri, and Medina]{Generalized}
Corinna Cortes, Mehryar Mohri, and Andres~Mu{\~{n}}oz Medina.
\newblock {Adaptation Based on Generalized Discrepancy}.
\newblock \emph{Machine Learning Research}, forthcoming.
\newblock URL \url{http://www.cs.nyu.edu/~mohri/pub/daj.pdf}.

\bibitem[Ganti and Gray(2012)]{Ganti2012a}
Ravi Ganti and Alexander Gray.
\newblock {UPAL: Unbiased Pool Based Active Learning}.
\newblock In \emph{Proceedings of the 15th International Conference on
  Artificial Intelligence and Statistics (AISTATS)}, pages 422--431, 2012.

\bibitem[Gretton et~al.(2012)Gretton, Borgwardt, Rasch, Sch{\"{o}}lkopf, and
  Smola]{Gretton2012Nieuw}
Arthur Gretton, Karsten~M Borgwardt, Malte~J Rasch, Bernhard Sch{\"{o}}lkopf,
  and Alexander Smola.
\newblock {A Kernel Two-sample Test}.
\newblock \emph{Machine Learning Research}, 13\penalty0 (1):\penalty0 723--773,
  2012.

\bibitem[Gu and Han(2012)]{Gu2012}
Quanquan Gu and Jiawei Han.
\newblock {Towards Active Learning on Graphs: An Error Bound Minimization
  Approach}.
\newblock In \emph{Proceedings of the 12th IEEE International Conference on
  Data Mining (ICDM)}, pages 882--887, 2012.

\bibitem[Gu et~al.(2012)Gu, Zhang, Han, and Ding]{Gu2012aNieuw}
Quanquan Gu, Tong Zhang, Jiawei Han, and Chris~H Ding.
\newblock {Selective Labeling via Error Bound Minimization}.
\newblock In \emph{Proceedings of the 25th Conference on Advances in Neural
  Information Processing Systems (NIPS)}, pages 323--331, 2012.

\bibitem[Gu et~al.(2014)Gu, Zhang, and Han]{Gu2014}
Quanquan Gu, Tong Zhang, and Jiawei Han.
\newblock {Batch-Mode Active Learning via Error Bound Minimization}.
\newblock In \emph{Proceedings of the 30th Conference on Uncertainty in
  Artificial Intelligence (UAI)}, 2014.

\bibitem[Huang et~al.(2007)Huang, Smola, Gretton, Borgwardt, and
  Sch{\"{o}}lkopf]{GrettonNieuw}
Jiayuan Huang, Alexander~J. Smola, Arthur Gretton, Karsten~M. Borgwardt, and
  Bernhard Sch{\"{o}}lkopf.
\newblock {Correcting sample selection bias by unlabeled data}.
\newblock In \emph{Proceedings of the 19th Conference on Advances in Neural
  Information Processing Systems (NIPS)}, pages 601--608, 2007.

\bibitem[Huang et~al.(2010)Huang, Jin, and Zhou]{Huang2010a}
Sheng-jun Huang, Rong Jin, and Zhi-hua Zhou.
\newblock {Active Learning by Querying Informative and Representative
  Examples}.
\newblock In \emph{Proceedings of the 23th Conference on Advances in Neural
  Information Processing Systems (NIPS)}, pages 892--900, 2010.

\bibitem[Lichman(2013)]{UCI}
M~Lichman.
\newblock {UCI Machine Learning Repository}, 2013.
\newblock URL \url{http://archive.ics.uci.edu/ml}.

\bibitem[Mansour et~al.(2009)Mansour, Mohri, and Rostamizadeh]{Mansour2009}
Yishay Mansour, Mehryar Mohri, and Afshin Rostamizadeh.
\newblock {Domain Adaptation: Learning Bounds and Algorithms}.
\newblock In \emph{Proceedings of the 22nd Annual Conference on Learning Theory
  (COLT)}, 2009.

\bibitem[Mohri et~al.(2012)Mohri, Rostamizadeh, and
  Talwalkar]{mohri2012foundations}
Mehryar Mohri, Afshin Rostamizadeh, and Ameet Talwalkar.
\newblock \emph{{Foundations of Machine Learning}}.
\newblock MIT press, Cambridge, Massachusetts, 2012.

\bibitem[Rifkin et~al.(2003)Rifkin, Yeo, and Poggio]{Rifkin2003}
Ryan Rifkin, Gene Yeo, and Tomaso Poggio.
\newblock {Regularized least-squares classification}.
\newblock \emph{Advances in Learning Theory: Methods, Model, and Applications},
  190:\penalty0 131--154, 2003.

\bibitem[Settles(2012)]{SettlesNieuw}
Burr Settles.
\newblock {Active Learning}.
\newblock \emph{Synthesis Lectures on Artificial Intelligence and Machine
  Learning}, 6\penalty0 (1):\penalty0 1--114, 2012.

\bibitem[Shawe-Taylor and Cristianini(2004)]{Shawe-Taylor2004}
J~Shawe-Taylor and N~Cristianini.
\newblock \emph{{Kernel Methods for Pattern Analysis}}.
\newblock Cambridge University Press, Cambridge, UK, 2004.

\bibitem[Wang and Ye(2013)]{Wang2013Nieuw}
Zheng Wang and Jieping Ye.
\newblock {Querying Discriminative and Representative Samples for Batch Mode
  Active Learning}.
\newblock In \emph{Proceedings of the 19th ACM SIGKDD International Conference
  on Knowledge Discovery and Data Mining (KDD)}, pages 158--166, 2013.

\end{thebibliography}


\begin{supplement}
\section{Background Theory}

\subsection{Computation of the MMD}

The MMD quantity can be computed in practice by rewriting it as follows:
\begin{align}
\max_{\tilde{g} \in H'} \frac{1}{n_{\hat{P}}} \sum_{x \in \hat{P}} \tilde{g}(x) - \frac{1}{n_{\hat{Q}}} \sum_{x \in \hat{Q}} \tilde{g}(x) =& \max_{\tilde{g} \in H'} \frac{1}{n_{\hat{P}}} \sum_{x \in \hat{P}} \langle\tilde{g},\psi(x)\rangle_{K'} - \frac{1}{n_{\hat{Q}}} \sum_{x \in \hat{Q}} \langle\tilde{g},\psi(x)\rangle_{K'}\label{rarekut}\\ 
=& \max_{\tilde{g} \in H'} \langle\tilde{g}, \frac{1}{n_{\hat{P}}} \sum_{x \in \hat{P}} \psi(x)\rangle_{K'} -  \langle\tilde{g},\sum_{x \in \hat{Q}} \frac{1}{n_{\hat{Q}}} \psi(x)\rangle_{K'}\\
=&  \max_{\tilde{g} \in H'} \langle\tilde{g}, \mu_{\hat{P}} - \mu_{\hat{Q}}\rangle_{K'} \label{MMD_def}\\
=& \Lambda' || \mu_{\hat{P}} - \mu_{\hat{Q}} ||_{K'} \label{MMD_combi1}.
\end{align}

In the first step we used that $\tilde{g}(x) = \langle\tilde{g},\psi(x)\rangle_{K'}$ due to the reproducing property \citep[p. 96]{mohri2012foundations}. Here $\psi$ is the mapping from $\mathcal{X}$ to the RKHS $\mathcal{H}'$ of $K'$. The other steps follow from the linearity of the inner product.  In Equation \ref{MMD_def} we defined $\mu_{\hat{P}} = \frac{1}{n_{\hat{P}}} \sum_{x \in \hat{P}} \psi(x)$ and similarly for $\mu_{\hat{Q}}$, note that these are vectors in the RKHS of $K'$. The last step follows from the fact that the vector in $H'$ maximizing the term in Equation \ref{MMD_def} is: 
\begin{equation}
\tilde{g} = \frac{\mu_{\hat{P}} - \mu_{\hat{Q}}}{|| \mu_{\hat{P}} - \mu_{\hat{Q}} ||_{K'}}\Lambda' \label{MMD_g_uitrekenen}
\end{equation}
This follows from the fact that the inner product between two vectors is maximum if the vectors point in the same direction. 
Because of the symmetry of $|| \mu_{\hat{P}} - \mu_{\hat{Q}} ||_{K'}$ with respect to $\hat{P}$ and $\hat{Q}$, it's straightforward to show that this derivation also holds if we switch $\hat{P}$ and $\hat{Q}$.

We can compute the MMD quantity in practice by working out the norm with kernel products:
\begin{align}
\Lambda' ||\mu_{\hat{Q}} - \mu_{\hat{P}}||_{K'} =& \Lambda' \sqrt{\langle\mu_{\hat{Q}},\mu_{\hat{Q}}\rangle_{K'} - 2\langle\mu_{\hat{P}},\mu_{\hat{Q}}\rangle_{K'} + \langle\mu_{\hat{Q}}, \mu_{\hat{P}}\rangle_{K'}}\\
=& \Lambda'\sqrt{\frac{1}{n_{\hat{Q}}^2} \sum_{x,x' \in \hat{Q}} K'(x,x') - 2 \frac{1}{n_{\hat{P}} n_{\hat{Q}}} \sum_{x \in \hat{P}} \sum_{x' \in \hat{Q}} K'(x,x') + \frac{1}{n_{\hat{P}}^2} \sum_{x,x' \in \hat{P}} K'(x,x')}\label{MMD_compute}
\end{align}

\subsection{Computation of the Discrepancy} \label{subsec_disc}

In this section we calculate the discrepancy analytically for the squared loss in the linear kernel following the derivation of \citep[p. 8]{Mansour2009}. At the end of this section we extend the computation to any arbitrary kernel following the derivation of \citep[Section 5.2]{Cortes2014}.
In our setting $\QH \in \PH$ and all samples have equal weights, therefore this derivation is slightly adapted from \citep{Mansour2009} and \citep{Cortes2014}.
We first rewrite the discrepancy for the linear kernel. As in Theorem 5 we take $u = h - h'$. We directly use the results of Theorem 5. 
\begin{align}
\disc(\hat{P},\hat{Q}) &= \max_{h,h' \in H} |L_{\hat{P}}(h',h) - L_{\hat{Q}}(h',h)|\\
                       &= \max_{||u|| \leq 2\Lambda} |u^T M u|\\
											 &= \max_{||\bar{u}|| \leq 2\Lambda} |\sum_i^d \bar{u}_i^2 \lambda_i|\\
											 &= \max\left(\max_{||\bar{u}|| \leq 2\Lambda} \sum_i^d \bar{u}_i^2 \lambda_i, \max_{||\bar{u}|| \leq 2\Lambda} \sum_i^d \bar{u}_i^2 -\lambda_i\right) \label{eq_dubbele_max}
\end{align}
First we solve the first term:
\begin{equation}
\max_{||\bar{u}|| \leq 2\Lambda} \sum_i^d \bar{u}_i^2 \lambda_i \label{eq_max_me}
\end{equation}
Observe that $\bar{u}$ provides a weighted combination of eigenvalues. To maximize this equation, we therefore need to put all the weight of $\bar{u}$ on the largest eigenvalue.
Thus the vector $u$ that maximizes this is given by a multiple of the eigenvector $e_{\text{max}}$ corresponding to the maximum eigenvalue $\lambda_{\text{max}}$:
\begin{equation}
u = e_{max} 2 \Lambda
\end{equation}
Note $||u|| = 2 \Lambda$ (since the eigenvector $e_{max}$ is orthonormal) to maximize the quantity in Equation \ref{eq_max_me}.
Substituting the solution of $u$ and using that the eigendecomposition is orthogonal we obtain:
\begin{align}
\max_{||\bar{u}|| \leq 2\Lambda} \sum_i^d \bar{u}_i^2 \lambda_i = 4 \Lambda^2 \max_i \lambda_i
\end{align}
Now for the maximization of the second term of Equation \ref{eq_dubbele_max}, observe that the solution changes sign (since now we want to place all the weight on the minimum eigenvalue. Thus we find that:
\begin{align}
\disc(\hat{P},\hat{Q}) &= 4 \Lambda^2 \max(\lambda_{\text{max}},-\lambda_{\text{min}})\\
&= 4 \Lambda^2 \max_i |\lambda_i| = 4 \Lambda^2 |\lambda_1|\\
&= 4 \Lambda^2 ||M||_2
\end{align}
Where $||M||_2$ is also known as the spectral norm of the matrix $M$, which is given by the largest absolute eigenvalue $|\lambda_1|$.

Now we can compute the discrepancy in a linear kernel. In an arbitrary kernel we cannot easily compute the covariance matrices of the sets $\PH$ and $\QH$, since the RKHS of $K$ may be very large or infinite, such as for the Gaussian kernel. In the following we rewrite the spectral norm of $M$ in terms of kernel innerproducts, so the discrepancy can be computed in any arbitrary kernel.

First we introduce the set $\UH = \PH \setminus \QH$. We assume in the following that the matrix $X_\PH$ is structured as:
\begin{equation}
X_\PH = \begin{bmatrix}X_\QH\\X_\UH \end{bmatrix}
\end{equation}
It can be shown that $M$ can be rewritten as \citep{Cortes2014}:
\begin{equation}
M = X_{\hat{P}}^T D X_{\hat{P}}
\end{equation}
Where $D$ is an $n_{\hat{P}} \times n_{\hat{P}}$ diagonal matrix. The matrix $D$ reweights all objects and is given by:
\begin{equation}
D = \begin{bmatrix} (\frac{1}{n_{\hat{P}}}-\frac{1}{n_{\hat{Q}}}) I & 0 \\ 0 & \frac{1}{n_\PH} I \end{bmatrix}
\end{equation}
Where $(\frac{1}{n_{\hat{P}}}-\frac{1}{n_{\hat{Q}}}) I$ is a diagonal matrix of size $n_\QH \times n_\QH$, and $\frac{1}{n_\PH} I$ is a diagonal matrix of size $n_\UH \times n_\UH$. 

Since the matrix product $AB$ and $BA$ have the same eigenvalues \citep{Cortes2014}, and since $||M||_2$ only depends on the eigenvalues, we can permute the matrices in $M$ to obtain a new matrix $M_K$ while $||M||_2 = ||M_K||_2$:
\begin{align}
M &= (X_{\hat{P}}^T D) X_{\hat{P}} \\
M_K &= X_{\hat{P}} (X_{\hat{P}}^T D) = K_{\hat{P}\hat{P}} D \label{def_MK}
\end{align}
Here $K_{\hat{P}\hat{P}}$ is the kernel matrix of the set $\PH$, meaning it contains the kernel products $K(x,x')$ for all objects in $\PH$.
Now $M_K$ only depends on the kernel matrix of $\PH$. Note that the kernel matrix should be ordered the same as $X_\PH$, thus the kernel matrix is given by:
\begin{equation}
K_{\hat{P}} = \left[ \begin{array}{ll}
K_{\hat{Q}\hat{Q}} & K_{\hat{Q}\hat{U}}\\
K_{\hat{U}\hat{Q}} & K_{\hat{U}\hat{U}}
\end{array} \right]
\end{equation}
Now the discrepancy can be computed in any arbitrary kernel using:
\begin{equation}
\text{disc}(\PH,\QH) = 4 \Lambda^2 ||M_K||_2 \label{disc_arb_kernel}
\end{equation}
Note that then we have to compute the largest absolute eigenvalue of the matrix $M_K$ to compute the discrepancy.

\section{Proofs}

\subsection{Proof of Theorem 1}

Let $L$ be any loss function, let $g(x) = L(h(x),f(x))$ and let the function $\tilde{g}$ be any arbitrary function in $H' = \{ h \in \mathcal{H}': ||h||_{K'} \leq \Lambda'\}$, where $K'$ is a PDS kernel with RKHS $\mathcal{H}'$. We indicate the empirical average of the functions $g$ and $\tilde{g}$ on a set of samples $\hat{S}$ by $g_{\hat{S}}$ and $\tilde{g}_{\hat{S}}$, respectively. 
We aim to bound the quantity:
\begin{equation}
|L_{\hat{P}}(h,f) - L_{\hat{Q}} (h,f)| = |g_\PH - g_\QH|
\end{equation}
Observe that:
\begin{equation}
|g_\PH - g_\QH| = |g_\PH - g_\QH + \tilde{g}_\PH - \tilde{g}_\PH + \tilde{g}_\QH - \tilde{g}_\QH|
\end{equation}
By reordering the terms and applying the triangle inequality twice we can show that:
\begin{equation}
|g_\PH - g_\QH| \leq |\tilde{g}_\PH - \tilde{g}_\QH| + |g_\PH - \tilde{g}_\PH| + |g_\QH - \tilde{g}_\QH| 
\end{equation}
The first term on the right hand side can be bounded by the MMD quantity:
\begin{equation}
|\tilde{g}_\PH - \tilde{g}_\QH| \leq \max_{\tilde{g} \in H'} |\tilde{g}_\PH - \tilde{g}_\QH| = \MMD(\PH,\QH)
\end{equation}
Then we obtain:
\begin{equation}
|g_\PH - g_\QH| \leq \MMD(\PH,\QH) + |g_\PH - \tilde{g}_\PH| + |g_\QH - \tilde{g}_\QH| \label{mmdstep3}
\end{equation}
Now we simplify the remaining two terms on the right. These terms appear because we may have that $g \notin H'$.
Observe that due to the triangle inequality we have that:
\begin{equation}
|g_\PH - \tilde{g}_\PH| \leq \frac{1}{n_\PH} \sum_{x \in \PH} |g(x) - \tilde{g}(x)| \label{mmdstep1}
\end{equation}
By maximizing over $x \in \PH$ we can show that:
\begin{equation}
\frac{1}{n_\PH} \sum_{x \in \PH} |g(x) - \tilde{g}(x)| \leq \max_{x \in \PH} |g(x) - \tilde{g}(x)| \label{mmdstep2}
\end{equation}
Combining Equations \ref{mmdstep1} and \ref{mmdstep2} we have that:
\begin{equation}
|g_\PH - \tilde{g}_\PH| \leq \max_{x \in \PH} |g(x) - \tilde{g}(x)| 
\end{equation}
Thus we have shown that:
\begin{equation}
|g_\PH - \tilde{g}_\PH| \leq \max_{x \in \PH} |g(x) - \tilde{g}(x)| 
\end{equation}
This same result can be derived for the second term:
\begin{equation}
|g_\QH - \tilde{g}_\QH| \leq \max_{x \in \QH} |g(x) - \tilde{g}(x)| 
\end{equation}
Now in our setting $\QH \subseteq \PH$, thus we can bound this using the term of $\PH$:
\begin{equation}
|g_\QH - \tilde{g}_\QH| \leq \max_{x \in \PH} |g(x) - \tilde{g}(x)| 
\end{equation}
Therefore, we have that:
\begin{equation}
|g_\PH - \tilde{g}_\PH| + |g_\QH - \tilde{g}_\QH| \leq 2 \max_{x \in \PH} |g(x) - \tilde{g}(x)| 
\end{equation}
Combining this with Equation \ref{mmdstep3} we find:
\begin{equation}
|g_\PH - g_\QH| \leq \MMD(\PH,\QH) + 2 \max_{x \in \PH} |g(x) - \tilde{g}(x)| 
\end{equation}
Now we make the bound independent of $h$ by maximizing over all $h \in H$: 
\begin{equation}
|g_\PH - g_\QH| \leq \MMD(\PH,\QH) + 2 \max_{h \in H} \max_{x \in \PH} |g(x) - \tilde{g}(x)| 
\end{equation}
Up to now our results hold for any $\tilde{g} \in H'$. 
Now to make the bound tight, we minimize with respect to $\tilde{g}$:
\begin{equation}
|g_\PH - g_\QH| \leq \MMD(\PH,\QH) + 2 \min_{\tilde{g} \in H} \max_{h \in H} \max_{x \in \PH} |g(x) - \tilde{g}(x)| 
\end{equation}
Because of the absolute value, the equation below also holds:
\begin{equation}
g_\PH - g_\QH \leq \MMD(\PH,\QH) + 2 \min_{\tilde{g} \in H} \max_{h \in H} \max_{x \in \PH} |g(x) - \tilde{g}(x)| 
\end{equation}
Rewriting we obtain:
\begin{equation}
L_\PH(h,f) \leq L_\QH(h,f) + \MMD(\PH,\QH) + 2 \min_{\tilde{g} \in H} \max_{h \in H}\max_{x \in \PH} |g(x) - \tilde{g}(x)| 
\end{equation}
For clarity, we now plug in $g(x) = L(h(x),f(x))$:
\begin{equation}
L_\PH(h,f) \leq L_\QH(h,f) + \MMD(\PH,\QH) + 2 \min_{\tilde{g} \in H} \max_{h \in H, x \in \PH} |L(h(x),f(x)) - \tilde{g}(x)| 
\end{equation}

\subsection{Proof of Theorem 2}

Let $L$ be the squared loss and we assume the realizable setting $f,h \in H$. 

\subsubsection{Proof in the Linear Case}

First we show the theorem for the case where $K$ is the linear kernel: $K(x,x') = x^T x'$.

Let $z(x) = h(x)-f(x)$.
Then $||z||_K = ||h - f||_K \leq 2 \Lambda$, since $||h||_K \leq \Lambda$ and $||f||_K \leq \Lambda$. The loss function is $g(x) = z(x)^2 = (h(x) - f(x))^2$. 

We define the squared kernel of $K$ as $K'(x,x') = \langle x,x' \rangle_{K}^2 = (x^T x')^2 = K(x,x')^2$.
The featuremap $\psi$ of $K'$ that maps from $\mathcal{X} = \mathcal{H}$ to $\mathcal{H}'$ is given by \citep[chap.~9.1]{Shawe-Taylor2004}\footnote{Note that actually in \citep{Shawe-Taylor2004} this kernel is defined as a polynomial kernel. In our case for this polynomial kernel we have that $R = 0$ and $d = 2$, resulting in the featuremap given in Equation \ref{eq_fmap}. This is often referred to as the squared kernel.}: 
\begin{equation}
\psi(x) = (x_1^2,x_2^2,\sqrt{2} x_1 x_2, x_3^2, \sqrt{2} x_1 x_3, \sqrt{2} x_2 x_3, x_4^2, \sqrt{2} x_1 x_4, \sqrt{2} x_2 x_4, \sqrt{2} x_3 x_4, \ldots) \label{eq_fmap}
\end{equation}
Note the kernel $K'$ is a PSD kernel since its featuremap exists.
The function $z(x)$ can be described as $z(x) = \langle z, x \rangle_{K} = z^T x$. Thus the function $g(x) = z(x)^2 = \langle z, x \rangle_{K}^2 = K'(z,x) = \langle \psi(z), \psi(x) \rangle_{K'}$, thus $g \in H'$ with $g = \psi(z)$. Furthermore we have that $||g||_{K'} = \langle \psi(z), \psi(z) \rangle_{K'} = K'(z,z) = \langle z, z \rangle_{K}^2 = ||z||_K^2 \leq 4 \Lambda^2$, since $||z||_K \leq 2 \Lambda$. Thus we have shown that $||g||_{K'} \leq 4 \Lambda^2$.

In conclusion, we have that:
\begin{equation}
g \in H' = \{ \forall h \in \mathcal{H}': ||h||_{K'} \leq \Lambda' = 4 \Lambda^2\}
\end{equation}
This will ensure $\eta_{\text{MMD}} = 0$.

\subsubsection{Proof for any Kernel}

Now we prove the more general case where $K$ is any kernel. 
\begin{table}
\centering
\caption{This table illustrates the notation used when 2 kernels are involved.}
\label{table_kernel}
\begin{tabular}{l|lclcl}
Transformation &               & $\psi_{K}$           &               & $\psi_{K'}$          &                \\ \hline
Space          & $\mathcal{X}$ & $\rightarrow$        & $\mathcal{H}$ & $\rightarrow$        & $\mathcal{H'}$ \\ \hline
Kernel         &               & \multicolumn{1}{l}{} & $K$           & \multicolumn{1}{l}{} & $K'$          
\end{tabular}
\end{table}
First we introduce some notation. We define the squared kernel $K'$ as:
\begin{equation}
K'(f,h) = \langle f,h \rangle^2_{K} \label{eq_def_K_accent}
\end{equation}
Where $f \in \mathcal{H}$ and $g \in \mathcal{H}$, where $\mathcal{H}$ is the RKHS of $K$. We indicate $\mathcal{H}'$ as the RKHS of $K'$. We assume $K$ is a PSD kernel. By definition of $K'$ the kernel $K'$ is a PSD kernel since a squared kernel is known to be PSD\citep[Theorem 5.3]{mohri2012foundations}.
Now we have two kernels we have two featuremaps: $\psi_K(x)$ which maps the input space $\mathcal{X}$ to the RKHS of $K$ (this map exists since we assume $K$ is a PSD kernel), and $\psi_{K'}(h)$ which maps a vector from the RKHS of $K$ to the RKHS of $K'$. Note that the second featuremap $\psi_{K'}(h)$ remains the same quadratic featuremap as before but now maps from $\mathcal{H}$ to $\mathcal{H}'$. See Table \ref{table_kernel} for an overview of the notation used.

Recall that because $K$ is a PSD kernel we have that:
\begin{equation}
K(x,x') = \langle \psi_K(x), \psi_K(x') \rangle_K \label{eq_map_to_RKHS_K}
\end{equation}
For $x,x' \in \mathcal{X}$.
Similarly for the kernel $K'$ which is also PSD we have that:
\begin{equation}
K'(f,g') = \langle \psi_{K'}(f), \psi_{K'}(g) \rangle_{K'} \label{eq_map_to_RKHS_Kaccent}
\end{equation}
For $f,g \in \mathcal{H}$.
 We define $z$ as:
\begin{equation}
z = h - f
\end{equation}
The function $g$ is given by:
\begin{equation}
g(x) = z(x)^2
\end{equation}
We have that $||h||_K \leq \Lambda$ and $||f||_K \leq \Lambda$ since we assumed the realizable setting. Then it is straightforward to show that: 
\begin{equation}
||z||_K = ||h - f||_K \leq 2 \Lambda \label{eq_bound_op_z}
\end{equation}
Since $h$ and $f$ are in the RKHS of $K$, $z$ is also in the RKHS of $K$. Thus we can write $z$ as an innerproduct in the RKHS of $K$:
\begin{equation}
z(x) = \langle z, \psi_K(x) \rangle_{K}
\end{equation}
Now we show that the function $g \in \mathcal{H}'$, in other words we show that $g$ is in the RKHS of $K'$.
By definition we have that:
\begin{equation}
g(x) = z(x)^2 = \langle z, \psi_K(x) \rangle_{K}^2
\end{equation}
Now we can easily recognize our definition of $K'$ in this equation (compare with Equation \ref{eq_def_K_accent}), thus we note that:
\begin{equation}
g(x) = K'(z,\psi_K(x))
\end{equation}
Now since $K'$ is a PSD kernel, each kernel product corresponds to an innerproduct in its RKHS. Note that the vectors $z$ and $\psi_K(x)$ are vectors in $\mathcal{H}$. To map these vectors to the RKHS of $K'$ we need to use the featuremap $\psi_{K'}$. We thus apply Equation \ref{eq_map_to_RKHS_Kaccent} resulting in:
\begin{equation}
g(x) = \langle \psi_{K'}(z), \psi_{K'}(\psi_K(x)) \rangle_{K'}
\end{equation}
We observe that $g$ corresponds to the vector $\psi_{K'}(z) \in \mathcal{H}'$, and thus we have that $g \in \mathcal{H}'$.

Now we show that $||g||_{K'} \leq 4 \Lambda^2$ to show that $g \in H'$. Since $g = \psi_{K'}(z) \in \mathcal{H}'$ the norm of $g$ in $K'$ is given by:
\begin{equation}
||g||_{K'}^2 = \langle \psi_{K'}(z), \psi_{K'}(z) \rangle_{K'}
\end{equation}
Now we can use Equation \ref{eq_map_to_RKHS_Kaccent} to rewrite this in terms of $K'$. We obtain:
\begin{equation}
||g||_{K'}^2 = K'(z,z) \label{app_temp1}
\end{equation} 
Using the definition of $K'$ we find:
\begin{equation}
K'(z,z) = \langle z, z \rangle_{K}^2 = ||z||_K^4 \label{app_temp2}
\end{equation}
Now recall we showed earlier in Equation \ref{eq_bound_op_z} that $||z||_K \leq 2 \Lambda$. Combining this with Equations \ref{app_temp1} and \ref{app_temp2} we find that:
\begin{equation}
||g||_{K'} = ||z||_K^2 \leq 4 \Lambda^2
\end{equation}
Thus we have shown that for any arbitrary kernel $K$ that $g \in H' = \{ \forall h \in \mathcal{H}': ||h||_{K'} \leq \Lambda' = 4 \Lambda^2\}$.

Finally, we show that to compute the MMD we can use the kernel $K''(x,x') = K(x,x')^2$. Since we require that kernel products are computed between objects in $\mathcal{X}$ to compute the MMD (see Equation \ref{MMD_compute}). As of now we defined the kernel $K'(f,h)$ so that it operates on $f,g \in \mathcal{H}$ and therefore we cannot use it to compute the MMD empirically using Equation \ref{MMD_compute}. To this end we show that $g \in H'' = \{ \forall h \in \mathcal{H}'': ||h||_{K''} \leq \Lambda' = 4 \Lambda^2\}$ by showing that $\mathcal{H}' = \mathcal{H}''$, where $\mathcal{H}''$ is the RKHS of $K''$. Then we can use $K''$ to compute the MMD using Equation \ref{MMD_compute} since $K''$ satisfies the assumptions of the MMD bound and $K''$ operates on objects in $\mathcal{X}$.

By definition of $K''$ we have that:
\begin{equation}
K''(x,x') = K(x,x')^2
\end{equation}
Now using Equation \ref{eq_map_to_RKHS_K} we can show that:
\begin{equation}
K''(x,x') = K(x,x')^2 = \langle\psi_K(x),\psi_K(x')\rangle_K^2
\end{equation}
Observe that this coincides with the definition of $K'$ (Equation \ref{eq_def_K_accent}), thus we can write this as:
\begin{equation}
K''(x,x') = \langle\psi_K(x),\psi_K(x')\rangle_K^2 = K'(\psi_K(x),\psi_K(x'))
\end{equation}
Now using Equation \ref{eq_map_to_RKHS_Kaccent} we can write this as:
\begin{equation}
K''(x,x') = K'(\psi_K(x),\psi_K(x')) = \langle\psi_{K'}(\psi_K(x)),\psi_{K'}(\psi_K(x'))\rangle_{K'}
\end{equation}
in other words, we see that the kernel product of $K''$ can be computed in the RKHS of the kernel $K'$. Thus, the RKHS of $K'$ and $K''$ coincide! Thus we have that $\mathcal{H}' = \mathcal{H}''$ and we can generalize all results in terms of $K'$ to the kernel $K''$. Therefore, $g$ is also in the RKHS of $K''$, and in particular we have that $g \in H'' = \{ \forall h \in \mathcal{H}'': ||h||_{K''} \leq \Lambda' = 4 \Lambda^2\}$. This could also be observed by noting that the featuremap of $K''$ is given by $\psi_{K''}(x) = \psi_{K'}(\psi_K(x))$ and thus maps to the space $\mathcal{H}'$, and from this it follows that $\mathcal{H}' = \mathcal{H}''$.

In conclusion, we have that:
\begin{equation}
g \in H'' = \{ \forall h \in \mathcal{H}'': ||h||_{K''} \leq \Lambda' = 4 \Lambda^2\}
\end{equation}
Where $K''(x,x') = K(x,x')^2$. In the statement of the theorem and in the rest of the main text we have renamed $K''$ to $K'$ for simplicity. 

Note that in both these cases (linear and arbitrary kernel) we have that $\eta_{\text{MMD}} = 0$, since $g' \in H$ the minimization over $g'$ will yield $g' = g$, and thus $\eta_{\text{MMD}} = 0$.

\subsection{Proof of Corollary 1}

We apply Theorem 2 to a Gaussian kernel $K$. As in Theorem 2 we let $f \in H$.
If we use a Gaussian kernel with bandwidth $\sigma$ for the learning algorithm $h$: 
\begin{equation}
K(x,x') = \exp\left(-\frac{||x - x'||^2}{2 \sigma^2}\right)
\end{equation}
We obtain for the kernel $K'$ that:
\begin{equation}
K'(x,x') = K(x,x')^2 = \exp\left(-\frac{2 ||x - x'||^2}{2 \sigma^2}\right) = \exp\left(-\frac{||x - x'||^2}{2 \sigma'^2}\right)
\end{equation}
Where we absorbed the factor of $2$ in the exponent in $\sigma'$, we obtain another Gaussian kernel with $\sigma' = \frac{\sigma}{\sqrt{2}}$. Thus, if one uses a Gaussian kernel with bandwidth $\sigma$ for the learning algorithm, $\sigma' = \frac{\sigma}{\sqrt{2}}$ for the kernel of the MMD together with the choice $\Lambda' = 4 \Lambda^2$ will ensure that $\eta_{\text{MMD}} = 0$.

\subsection{Proof of Theorem 3}

This proof is slightly adapted from the proof in \citep{Generalized}. Let $f$ be any deterministic labeling function and $h \in H$. We assume that for any $h$ and $f$ that: $L(h(x),f(x)) \leq C$.
Furthermore, we assume that the loss $L$ is $\mu$-admissible \citep{Generalized}.

\begin{definition}
A loss function $L$ is $\mu$-admissible if there exists a $\mu > 0$ such that the inequality
\begin{equation}
|L(h(x),f(x) - L(h'(x),f(x))| \leq \mu |h(x) - h'(x)|
\end{equation}
holds for all $(x,f(x)) \in \mathcal{X} \times \mathcal{Y}$ and $h,h' \in H$. 
\end{definition}
We assume $L$ is the squared loss, and for the squared loss it is possible to show that $\mu = 2C$ \citep{Generalized}. 

In case of the realizable setting it is straightforward to show the discrepancy bound: if $f \in H$ we simply maximize over $h$ and $f$ to obtain the bound:
\begin{equation}
|L_{\hat{P}}(h,f) - L_{\hat{Q}}(h,f)| \leq \max_{h,f \in H} |L_{\hat{P}}(h,f) - L_{\hat{Q}}(h,f)| = \disc(\PH,\QH)
\end{equation} 
To extend this bound to the agnostic case where $f \notin H$ this will require us to approximate the agnostic function $f$ by a function $\tilde{f} \in H$. We give the proof below.

Observe that the following equation holds for all $\tilde{f} \in H$:
\begin{equation}
|L_\PH(h,f) - L_\QH(h,f)| = |L_\PH(h,f) - L_\QH(h,f) + L_\PH(h,\tilde{f}) - L_\PH(h,\tilde{f}) - L_\QH(h,\tilde{f}) + L_\QH(h,\tilde{f})|
\end{equation}
If we rearrange the terms on the right hand side and apply the triangle inequality twice, we can show that:
\begin{equation}
|L_\PH(h,f) - L_\QH(h,f)| \leq |L_\PH(h,\tilde{f}) - L_\QH(h,\tilde{f})| + |L_\PH(h,f) - L_\PH(h,\tilde{f})| + |L_\QH(h,\tilde{f}) - L_\QH(h,f)|
\end{equation}
Now we can bound the first term on the right hand side by maximizing over all $h$ and $\tilde{f} \in H$:
\begin{equation}
|L_\PH(h,\tilde{f}) - L_\QH(h,\tilde{f})| \leq \max_{h,\tilde{f} \in H} |L_\PH(h,\tilde{f}) - L_\QH(h,\tilde{f})| = \disc(\PH,\QH)
\end{equation}
Then we obtain the following generalization bound which holds for all $\tilde{f} \in H$:
\begin{equation}
|L_\PH(h,f) - L_\QH(h,f)| \leq \disc(\PH,\QH) + |L_\PH(h,f) - L_\PH(h,\tilde{f})| + |L_\QH(h,\tilde{f}) - L_\QH(h,f)| \label{discstep2}
\end{equation}
Similar to the MMD bound we will bound the two last terms on the right hand side. These will form the approximation term.

Then by the $\mu$-admissibility of the squared loss we have (see the proof of Proposition 8 in \citep{Generalized}):
\begin{equation}
|L_\PH(h,f) - L_\PH(h,\tilde{f})| \leq \mu \frac{1}{n_\PH} \sum_{x \in \PH} |f(x) - \tilde{f}(x)|
\end{equation}
Maximizing $x$ over $\PH$ results in:
\begin{equation}
|L_\PH(h,f) - L_\PH(h,\tilde{f})| \leq \mu \max_{x \in \PH} |f(x) - \tilde{f}(x)|
\end{equation}
The same holds for the last term:
\begin{equation}
|L_\QH(h,f) - L_\QH(h,\tilde{f})| \leq \mu \max_{x \in \QH} |f(x) - \tilde{f}(x)|
\end{equation}
Now, since in our setting $\QH \subseteq \PH$ we have that:
\begin{equation}
\mu \max_{x \in \QH} |f(x) - \tilde{f}(x)| \leq \mu \max_{x \in \PH} |f(x) - \tilde{f}(x)|
\end{equation}
Thus we can bound both terms as:
\begin{equation}
|L_\PH(h,f) - L_\PH(h,\tilde{f})| + |L_\QH(h,\tilde{f}) - L_\QH(h,f)| \leq 2 \mu \max_{x \in \PH} |f(x) - \tilde{f}(x)|
\end{equation}

Combining this result with Equation \ref{discstep2} we find:
\begin{equation}
|L_\PH(h,f) - L_\QH(h,f)| \leq \disc(\PH,\QH) + 2 \mu \max_{x \in \PH} |f(x) - \tilde{f}(x)|
\end{equation}
This bound holds for all $\tilde{f} \in H$. Now we minimize over $\tilde{f}$ to obtain a tight bound:
\begin{equation}
|L_\PH(h,f) - L_\QH(h,f)| \leq \disc(\PH,\QH) + 2 \mu \min_{\tilde{f} \in H} \max_{x \in \PH} |f(x) - \tilde{f}(x)|
\end{equation}

Now we can rewrite the bound without absolute value as:
\begin{equation}
L_\PH(h,f) \leq L_\QH(h,f) + \disc(\PH,\QH) + 2 \mu \min_{\tilde{f} \in H} \max_{x \in \PH} |f(x) - \tilde{f}(x)|
\end{equation}
In particular for the squared loss $\mu = 2 C$ so we obtain:
\begin{equation}
L_\PH(h,f) \leq L_\QH(h,f) + \disc(\PH,\QH) + 4 C \min_{\tilde{f} \in H} \max_{x \in \PH} |f(x) - \tilde{f}(x)|
\end{equation}

\subsection{Proof of Theorem 4}

The goal of this section is to show that we can compute the MMD with the matrix $M$, where $M$ is given by:
\begin{equation}
M = \frac{1}{n_\PH} X_\PH^T X_\PH - \frac{1}{n_\QH} X_\QH^T X_\QH. \label{EquationdefM2}
\end{equation}
Let  $\lambda_i$ be the eigenvalues of $M$. Then we wish to show that:
\begin{equation}
\MMD(\PH,\QH) = 4 \Lambda^2 \sqrt{\sum_i \lambda_i^2}
\end{equation}
We require that the MMD is computed using the squared kernel $K'(x,x') = K(x,x')^2$.

To this end we will require the Frobenius norm:

\begin{definition}
The Frobenius norm for the $m$ by $n$ matrix $A$ is given by:
\begin{equation}
||A||_F^2 = \sum_{i = 1}^m \sum_{j = 1}^n |a_{ij}|^2
\end{equation}
\end{definition}

\subsubsection{Linear Kernel 2 Dimensions}

We first will show this result in $d = 2$ dimensions in the linear kernel $K(x,x') = x^T x'$. 

Let us introduce some notation. The $j$th component of object $i$ will be denoted by $x_{ij}$. The covariance matrix for set $\PH$ becomes:
\begin{equation}
\frac{1}{n_\PH} X_\PH^T X_\PH = \frac{1}{n_\PH}
\begin{pmatrix}\sum_{i \in \PH} x_{1i}^2 & \sum_{i \in \PH} x_{1i} x_{2i} \\
\sum_{i \in \PH} x_{1i} x_{2i} & \sum_{i \in \PH} x_{2i}^2 \end{pmatrix}
\end{equation}
The matrix $M$ is given by:
\begin{align}
M =& \frac{1}{n_\PH} X_\PH^T X_\PH - \frac{1}{n_\QH} X_\QH^T X_\QH\\ 
=& \begin{pmatrix} \frac{1}{n_\PH}\sum_{i \in \PH} x_{1i}^2 - \frac{1}{n_\QH}\sum_{i \in \QH} x_{1i}^2 & \frac{1}{n_\PH}\sum_{i \in \PH} x_{1i} x_{2i} - \frac{1}{n_\QH}\sum_{i \in \QH} x_{1i} x_{2i} \\
\frac{1}{n_\PH}\sum_{i \in \PH} x_{1i} x_{2i} - \frac{1}{n_\QH}\sum_{i \in \QH} x_{1i} x_{2i} & \frac{1}{n_\PH}\sum_{i \in \PH} x_{2i}^2 - \frac{1}{n_\QH}\sum_{i \in \QH} x_{2i}^2 \end{pmatrix}\\
=& \begin{pmatrix} \Delta_{11} & \Delta_{12}\\
\Delta_{21} & \Delta_{22} \end{pmatrix} \label{M_deltas}
\end{align}
We define $\Delta_{kl}$ to ease notation:
\begin{equation}
\Delta_{kl} \equiv \frac{1}{n_\PH}\sum_{i \in \PH} x_{ki} x_{li} - \frac{1}{n_\QH}\sum_{i \in \QH} x_{ki} x_{li}
\end{equation}
Note that $\Delta_{kl} = \Delta_{lk}$.

The MMD can be computed using:
\begin{equation}
\MMD(\PH,\QH) = \Lambda' ||\mu_\PH - \mu_\QH||_{K'} = 4 \Lambda^2 ||\mu_\PH - \mu_\QH||_{K'}
\end{equation}
Where $\mu_\PH$ and $\mu_\QH$ are the means of the sets $\PH$ and $\QH$ in the RKHS $H'$:
\begin{equation}
\mu_\PH = \sum_{x \in \PH} \psi(x)
\end{equation}
The mean of $\QH$ is defined similarly. 
Here $\psi(x)$ is the featuremap from $\mathcal{X} = \mathcal{H}$ to $\mathcal{H}'$ which was given in Equation \ref{eq_fmap}.

The mean of $\PH$ in the RKHS of $K'$ is given by:
\begin{equation}
\mu_\PH = \frac{1}{n_\PH} \begin{pmatrix} \sum_{i \in \PH} x_{1i}^2\\ \sum_{i \in \PH} x_{2i}^2 \\ \sum_{i \in \PH} \sqrt{2} x_{1i} x_{2i} \end{pmatrix}
\end{equation}
Note that this is in 3 dimensions, where we obtained these $x$ values by using the featuremap of the quadratic kernel: $\psi(x) = (x_1^2, x_2^2, \sqrt{2} x_1 x_2)$.
The difference between the means $\mu_\PH - \mu_\QH$ can be written as:
\begin{equation}
\mu_\PH - \mu_\QH = \begin{pmatrix} \frac{1}{n_\PH} \sum_{i \in \PH} x_{1i}^2 - \frac{1}{n_\QH} \sum_{i \in \QH} x_{1i}^2 \\
 \frac{1}{n_\PH} \sum_{i \in \PH} x_{2i}^2 - \frac{1}{n_\QH} \sum_{i \in \PH} x_{2i}^2\\
\sqrt{2}(\frac{1}{n_\PH} \sum_{i \in \PH} x_{1i} x_{2i} - \frac{1}{n_\QH} \sum_{i \in \QH} x_{1i} x_{2i}) \end{pmatrix} =
\begin{pmatrix} \Delta_{11} \\ \Delta_{22} \\ \sqrt{2} \Delta_{12}\end{pmatrix}
\end{equation}
The norm of $\mu_\PH - \mu_\QH$ is given by:
\begin{equation}
||\mu_\PH - \mu_\QH||^2_{K'} = \Delta_{11}^2 + \Delta_{22}^2 + 2 \Delta_{12}^2
\end{equation}
Note that the Frobenius norm of $M$ is given by the same expression (see Equation \ref{M_deltas}, and note that $\Delta_{12} = \Delta_{21}$):
\begin{equation}
||M||_F^2 = \Delta_{11}^2 + 2 \Delta_{12}^2 + \Delta_{22}^2 = || \mu_\PH - \mu_\QH ||^2_{K'} 
\end{equation}
Thus the MMD can be written as:
\begin{equation}
\MMD(\PH,\QH) = 4 \Lambda^2 || \mu_\PH - \mu_\QH ||_{K'}  = 4 \Lambda^2 ||M||_F
\end{equation}
It can be shown that the Frobenius norm is equal to the square root of the sum of squared singular values. However, since $M$ is a real symmetric matrix, the singular values $\sigma_i$ of $M$ are equal (except for the sign) to the eigenvalues $\lambda_i$ of $M$, and therefore here the Frobenius norm is equal to the square root of the sum of squared eigenvalues of $M$:
\begin{equation}
\MMD(\PH,\QH) = 4 \Lambda^2 \sqrt{\sum_i \lambda_i^2}
\end{equation}

\subsubsection{Linear Kernel d Dimensions}

This can be generalized to any arbitrary number of dimensions as follows.
In $d$ dimensions, the entries of the matrix $M$ become $M_{kl} = \Delta_{kl}$. Therefore the Frobenius norm of $M$ is given by:
\begin{equation}
||M||_F^2 = \sum_{i = 1\ldots d} \sum_{j = 1 \ldots d} \Delta_{ij}^2 = \sum_{i = 1\ldots d} \Delta_{ii}^2 + \sum_{j = i+1, \ldots, d} 2 \Delta_{ij}^2
\end{equation}
Correspondingly, the vector $\mu_\PH - \mu_\QH$ becomes:
\begin{equation}
\mu_\PH - \mu_\QH = \begin{pmatrix}\Delta_{11} \\ \Delta_{22} \\ \sqrt{2}\Delta_{12} \\ \Delta_{33} \\ \sqrt{2}\Delta_{13} \\  \sqrt{2}\Delta_{23} \\ \Delta_{44} \\ \sqrt{2}\Delta_{14} \\ \sqrt{2}\Delta_{24} \\ \sqrt{2}\Delta_{34} \\ \vdots \end{pmatrix} 
\end{equation}
The norm is given by:
\begin{align}
||\mu_\PH - \mu_\QH||_{K'}^2 =& \sum_{i = 1,\ldots,d} \Delta_{ii}^2 + \sum_{i = 1,\ldots,d} \sum_{j = i+1, \ldots, d} 2 \Delta_{ij}^2\\
=& \sum_{i = 1,\ldots,d} \Delta_{ii}^2 + \sum_{i \neq j} \Delta_{ij}^2\\
=& \sum_{i = 1\ldots d} \sum_{j = 1 \ldots d} \Delta_{ij}^2 = ||M||_F^2
\end{align}
And thus in this case it still holds that:
\begin{equation}
||M||_F = || \mu_\PH - \mu_\QH ||_{K'} 
\end{equation}
And thus:
\begin{equation}
\MMD(\PH,\QH) = 4 \Lambda^2 \sqrt{\sum_i \lambda_i^2}
\end{equation}

Note that due to our careful ordering of the featuremap of the squared kernel (Equation \ref{eq_fmap}) we have that this featuremap is still properly defined even if the dimension of $d \to \infty$. 

\subsubsection{Extension to Arbitrary Kernels}
The same result as above can be obtained for any arbitrary kernel $K$, only we have to work in the RKHS of $K$, thus everywhere $x$ needs to be replaced by $\psi(x)_K$. Furthermore $x_{ij}$ will become the $j$th component of $\psi(x_i)_K$, where $x_i$ will be object $i$. This proof still holds, since the featuremap $\psi(x)_{K'}$ is still given by the featuremap of the squared kernel, however in this case the featuremap is with respect to the RKHS of $K$: this does not influence the proof. We showed that this holds for any arbitrary dimension, thus our results hold for a kernel with arbitrary dimension of the RKHS of $K$.
For the Gaussian kernel we cannot compute the matrix $M$. Instead, since $M$ and $M_K$ have the same eigenvalues (where $M_K$ is defined in Equation \ref{def_MK}), we can compute the MMD instead using the eigenvalues of $M_K$.

So we have that:
\begin{equation}
\MMD(\PH,\QH) = 4 \Lambda^2 \sqrt{\sum_i \lambda_i^2} \label{imply}
\end{equation}
where $\lambda_i$ are the eigenvalues of $M_K$. Note that 
\begin{equation}
\MMD(\PH,\QH) \neq 4 \Lambda^2 ||M_K||_F.
\end{equation}
This does not hold, since the matrix $M_K$ is not symmetric. Therefore, the eigenvalues of $M_K$ are not the same as the singular values of $M_K$ (as was the case for $M$). Therefore, Equation \ref{imply} does not imply that the MMD can be computed using the Frobenius norm of $M_K$.

\subsection{Proof of Theorem 5}
Since $f \in H$, we have that $f = f^T x$. 
Thus wee have that:
\begin{equation}
L_\PH(h,f) = \frac{1}{n_\PH} (X_\PH h - X_\PH f)^T (X_\PH h - X_\PH f)
\end{equation}
Define $u = h - f$. Then we can show that:
\begin{equation}
L_\PH(h,f) = \frac{1}{n_\PH} (X_\PH u)^T (X_\PH u) = \frac{1}{n_\PH} u^T X_\PH^T X_\PH u
\end{equation}
We can show the same result for $L_\QH(h,f)$:
\begin{equation}
L_\QH(h,f) = \frac{1}{n_\QH} u^T X_\QH^T X_\QH u
\end{equation}
Rewriting:
\begin{equation}
L_\PH(h,f) - L_\QH(h,f) = \frac{1}{n_\PH} u^T X_\PH^T X_\PH u - \frac{1}{n_\QH} u^T X_\QH^T X_\QH u
\end{equation}
We can rewrite the above as:
\begin{equation}
L_\PH(h,f) - L_\QH(h,f) =  u^T \left(\frac{1}{n_\PH} X_\PH^T X_\PH - \frac{1}{n_\QH} X_\QH^T X_\QH\right) u
\end{equation}
Now we define the $d$ times $d$ matrix $M$ as:
\begin{equation}
M = \frac{1}{n_\PH} X_\PH^T X_\PH - \frac{1}{n_\QH} X_\QH^T X_\QH
\end{equation}
And thus we have that:
\begin{equation}
L_\PH(h,f) - L_\QH(h,f) =  u^T M u \label{app_combine_me}
\end{equation}
Since $M$ is a real symmetric matrix, $M$ is a normal matrix and admits an orthonormal eigendecomposition with real eigenvalues:
\begin{equation}
M = \sum_i^d e_i \lambda_i e_i^T
\end{equation}
Where $\lambda_i$ indicate the eigenvalues and $e_i$ are the corresponding eigenvectors with unit length. Since the eigendecomposition is orthonormal we have that $e_i^T e_j = 1$ only if $i = j$, otherwise $e_i^T e_j = 0$. 
Since $M$ is a normal matrix its orthonormal eigenvectors span the entire space $\mathbb{R}^d$, and thus form an orthonormal basis for $\mathbb{R}^d$. Because of this we can express the vector $u$ in terms of the eigenvectors of $M$:
\begin{equation}
u = \sum_i^d \bar{u}_i e_i
\end{equation}
Where $\bar{u}_i$ is the projection of $u$ on $e_i$:
\begin{equation}
\bar{u}_i = e_i^T u
\end{equation}

Now we can rewrite $u^T M u$:

\begin{align}
u^T M u 
        =& \sum_i^d u^T e_i \lambda_i e_i^T u \\
				=& \sum_i^d \bar{u}_i^2 \lambda_i
\end{align}
Combining with Equation \ref{app_combine_me} yields:
\begin{equation}
L_\PH(h,f) - L_\QH(h,f) = \sum_i^d \bar{u}_i^2 \lambda_i \label{EquationNewCombineMe}
\end{equation}
Taking the absolute value we have that:
\begin{equation}
|L_\PH(h,f) - L_\QH(h,f)| = |\sum_i^d \bar{u}_i^2 \lambda_i|
\end{equation}
Now if we assume $h$ is trained on $\QH$, $L_\QH(h,f) \approx 0$, and we have that:
\begin{equation}
L_\PH(h,f) \approx |\sum_i^d \bar{u}_i^2 \lambda_i|
\end{equation}

Since $||h||_K \leq \Lambda$ and $||f||_K \leq \Lambda$ we have that $||u||_K \leq 2\Lambda$. The vector $\bar{u}$ is equal up to $u$ except for a rotation (since $M$ is a normal matrix). Thus we have that, $||\bar{u}||_K = ||u||_K \leq 2\Lambda$. In the linear kernel this means that:
\begin{equation}
||\bar{u}||_K = \sqrt{\sum_i \bar{u}_i^2} \leq 2 \Lambda
\end{equation}

In Subsection \ref{subsecerrordecomp} we show how to compute $\bar{u}_i^2$ in this error decomposition explicitly.

\subsection{Proof of Theorem 6}

Let $h,f \in H$. Then $u = h-f$ has $||u||_K \leq 2 \Lambda$ since $||h||_K \leq \Lambda$ and $||f||_K \leq \Lambda$.
Thus $u \in U = \{u \in \mathcal{H} : ||u||_K \leq 2\Lambda\}$.
Since $\bar{u}$ is equal to $u$ up to a rotation, we have that $||\bar{u}||_K \leq 2 \Lambda$ as well.
We assume each $u$ in $U$ is equally likely, similarly each $\bar{u}$ in $U$ is equally likely.

We start from Equation \ref{EquationNewCombineMe} and take the absolute value on both sides:
\begin{equation}
|L_\PH(h,f) - L_\QH(h,f)| = |\sum_i^d \bar{u}_i^2 \lambda_i|
\end{equation}
We wish to bound the right hand side in expectation over $u$.
First, we bound the right hand side using the triangle inequality:
\begin{equation}
|\sum_i^d \bar{u}_i^2 \lambda_i| \leq \sum_i^d \bar{u}_i^2 |\lambda_i|
\end{equation}
Now we compute the expectation of the right hand side with respect to $u$, which is the same as in expectation with respect to $\bar{u}$. 
\begin{equation}
\Exp_u \sum_i^d \bar{u}_i^2 |\lambda_i| = \int_{U} \sum \bar{u}_i^2 |\lambda_i| p(\bar{u}) d\bar{u} \label{hard_integral}
\end{equation}
Where:
\begin{equation}
p(\bar{u}) = \frac{1}{\int_U d\bar{u}} = \frac{1}{vol(U)}
\end{equation}
The integral over the set $U$ in Equation \ref{hard_integral} is hard to compute. Therefore, we bound this integral by an integral over $V$, where the set $V = \{\bar{u} \in \mathcal{H} : \max_i |\bar{u}_i| \leq 2\Lambda\}$. Note that $V$ is the enclosing box around $U$ and thus $U \in V$.
The integral over $V$ can be solved in closed form.
Furthermore, take $D = V \setminus U$. We introduce shorthand notation $t(\bar{u}) = \sum {u}_i^2 |\lambda_i|$. 

Since the function $t(\bar{u})$ is a parabola that increases farther away from the origin, that the mean of $t$ on $D$ is larger than the mean of $t$ on $U$:
\begin{equation}
\frac{1}{vol(D)} \int_D t(\bar{u}) d\bar{u} \geq \frac{1}{vol(U)} \int_U t(\bar{u}) d\bar{u}
\end{equation}
We can rewrite this result above:
\begin{align}
vol(U) \int_D t(\bar{u}) d\bar{u} \geq&~ vol(D) \int_U t(\bar{u}) d\bar{u}\\
vol(U) \left(\int_D t(\bar{u}) d\bar{u} + \int_U t(\bar{u}) d\bar{u}\right) \geq&~ (vol(D) + vol(U)) \int_U t(\bar{u}) d\bar{u}\\
\frac{1}{vol(U) + vol(D)} \left(\int_D t(\bar{u}) d\bar{u} + \int_U t(\bar{u}) d\bar{u}\right) \geq&~ \frac{1}{vol(U)} \int_U t(\bar{u}) d\bar{u}\\
\frac{1}{vol(V)} \int_V t(\bar{u}) d\bar{u} \geq&~ \frac{1}{vol(U)} \int_U t(\bar{u}) d\bar{u} \label{last_step}
\end{align}
Now, we compute the integral of $t(\bar{u})$ over $V$:
\begin{align}
\frac{1}{vol(V)} \int_V t(\bar{u}) d\bar{u} &= \frac{1}{vol(V)} \int_V \sum_i {u}_i^2 |\lambda_i| d\bar{u}\\
                                            &= \frac{1}{vol(V)} \sum_i |\lambda_i| \int_V {u}_i^2  d\bar{u}\\
																						&= \frac{1}{vol(V)} \sum_i |\lambda_i| (4\Lambda)^{d-1} \int_{-2\Lambda}^{2\Lambda} {u}_i^2  d\bar{u_i}\\
																						&= \frac{1}{vol(V)} \sum_i |\lambda_i| (4\Lambda)^{d-1} \frac{2}{3} (2 \Lambda)^3
\end{align}
Note that $vol(V) = (4 \Lambda)^d$, thus:
\begin{equation}
\frac{1}{vol(V)} \int_V t(\bar{u}) d\bar{u} = \frac{4}{3} \Lambda^2 \sum_i |\lambda_i|
\end{equation}
Combining with Equation \ref{last_step} and Equation \ref{hard_integral} we have that:
\begin{equation}
\Exp_u \sum_i^d \bar{u}_i^2 |\lambda_i| \leq \frac{4}{3} \Lambda^2 \sum_i |\lambda_i|
\end{equation}
Thus, in expectation over $u \in U$ we have that:
\begin{equation}
\Exp_{u \in U} |L_\PH(h,f) - L_\QH(h,f)| \leq \frac{4}{3} \Lambda^2 \sum_i |\lambda_i|
\end{equation}
In particular, this also holds without absolute value. Thus we have that the following holds in expectation over $u \in U$:
\begin{equation}
\Exp_{u \in U} L_\PH(h,f) \leq \Exp_{u \in U} L_\QH(h,f) + \frac{4}{3} \Lambda^2 \sum_i |\lambda_i|
\end{equation}

\subsection{Proof of Theorem 7}

This theorem can directly be shown by observing that the Nuclear Discrepancy upperbounds the Discrepancy, and then upperbounding the Discrepancy  in the bound of Theorem 3 by the Nuclear Discrepancy. 

\subsection{Computation of the Error Decomposition} \label{subsecerrordecomp}

The computation of $\bar{u}_i$, the projection of $u$ onto the eigenvector $v_i$ of $M$ is non-trivial. Observe that here $v_i$ is the $i$th eigenvector and not a component. Here we assume $v_i$ is not normalized to unit norm. We give a detailed description in this appendix how to compute $\bar{u}_i$. In this case the equation for $\bar{u}_i$ is:
\begin{equation}
\bar{u}_i = \frac{u^T v_i}{\sqrt{v_i^T v_i}} \label{app_comp_ui}
\end{equation}
The difficulty in this derivation is finding the vector $v_i$ in case kernels are used. In that case we need to find $v_i$ expressed in terms of the datamatrix $X$. Then we can apply the `kernel trick' to compute Equation \ref{app_comp_ui}.

Note that in the linear kernel we have:
\begin{equation}
M v_i = \lambda_i v_i \label{eig_eq}
\end{equation}
In case of the linear kernel it is straightforward to compute $v_i$. To compute $v_i$ when kernels are used, first we show that $v_i$ can be expressed in terms of the datamatrix $X$, and afterward we find this expression of $v_i$ in terms of $X$.
Note that:
\begin{equation}
M v_i = \sum_{j=1}^{n_\PH} d_j x_j x_j^T v_i = \sum_{j=1}^{n_\PH} (x_j^T v_i) d_j x_j = \lambda_i v_i
\end{equation}
Thus we have that:
\begin{equation}
\sum_{j=1}^{n_\PH} \frac{(x_j^T v_i) d_j}{\lambda_i} x_j = v_i
\end{equation}
Thus we have that each eigenvector $v_i$ is a linear combination of the vectors $x_j$. Here the sum is taken over all objects $x \in \PH$. Since $\QH \subseteq \PH$, this includes all data the active learner has access to. 
Then we can write each eigenvector $v_i$ as:
\begin{equation}
v_i = X_\PH^T \alpha_i \label{app_eq_vi}
\end{equation}
Thus we can express each vector $v_i$ using the datamatrix $X_\PH$.
Now we will have to find the vector $\alpha_i$ to find $v_i$.
We substitute the equation above in equation \ref{eig_eq} to obtain:
\begin{equation}
M X_\PH^T \alpha_i = \lambda_i  X_\PH^T \alpha_i
\end{equation}
Now we multiply left with $D X_\PH$ on both sides to obtain:
\begin{equation}
D X_\PH M X_\PH^T \alpha_i = \lambda_i  D X_\PH X_\PH^T \alpha_i
\end{equation}
Observe that this is equal to:
\begin{equation}
M_K^T M_K^T \alpha_i = \lambda_i  M_K^T \alpha_i
\end{equation}
Where $M_K$ was defined in Equation \ref{def_MK}.
Now we define $\beta_i = M_K^T \alpha_i$. Then we find:
\begin{equation}
M_K^T \beta_i = \lambda_i \beta_i \label{app_mult_me}
\end{equation}
We can compute the eigenvectors $\beta$ by computing the eigendecomposition of $M_K^T$. This is possible even when using kernels, since $M_K$ is expressed in terms of the kernel matrix. However we require the vector $\alpha_i$ to compute the eigenvector $v_i$. Thus now we will aim to express $\alpha_i$ in terms of $\beta_i$.
Observe that if we multiply equation \ref{app_mult_me} by $(M_K^T)^{-1}$ on both sides we obtain:
\begin{equation}
\beta_i = \lambda_i (M_K^T)^{-1} \beta_i \label{app_eq_2}
\end{equation}
Now observe that due to the definition of $\beta_i$ we have that:
\begin{equation}
\beta_i (M_K^T)^{-1} =  \alpha_i \label{app_eq_1}
\end{equation}
Combining equation \ref{app_eq_2} and \ref{app_eq_1} we find that:
\begin{equation}
\alpha_i = \frac{\beta_i}{\lambda}
\end{equation}
Substituting this in equation \ref{app_eq_vi} we find the vector $v_i$:
\begin{equation}
v_i = X_\PH^T \frac{\beta_i}{\lambda} \label{app_eq_vi2}
\end{equation}
Now we have found $v_i$. Now we can proceed to compute $u_i$.

Note that due to the representer theorem we have that the hyperplane of each model can be written as a linear combination of the data:
\begin{equation}
u = f - h = X_{\hat{D}}^T c' - X_\QH^T c \equiv X_{\hat{D}}^T \tilde{c} \label{app_eq_u}
\end{equation}
Here $f$ is given as a linear combination of $X_{\hat{D}}$, which we define as the complete datamatrix. This datamatrix includes the training and test set, since $f$ in our experiments was obtained by training on the whole dataset where the original binary labels of the dataset are used. However note that for any $f \in H$ the model $f$ can be written in this way. Similarly, since $h$ is trained on the dataset $\QH$, we can write $h$ as a linear combination of objects in $\QH$.
Combining equation \ref{app_eq_vi2} and \ref{app_eq_u} with equation \ref{app_comp_ui} we find that:
\begin{equation}
u_i = \frac{\tilde{c} X_{\hat{D}} X_{\PH}^T \frac{\beta_i}{\lambda_i}}{\sqrt{\frac{\beta_i^T}{\lambda_i} X_{\PH} X_{\PH}^T \frac{\beta_i}{\lambda_i}}} = \frac{\tilde{c} K_{\hat{D}\hat{P}} \beta_i}{\sqrt{\beta_i K_{\PH\PH} \beta_i}}
\end{equation}

\FloatBarrier

\newpage
\section{Experimental Settings and Dataset Characteristics}

The active learning methods are evaluated on the datasets shown in Table \ref{table_data}. The datasets marked with $^*$ were provided by \citet{Cawley2004}. Other datasets originate from the UCI Machine Learning repository \citep{UCI}. 

\newcommand{\datasettable}[1]{\small{#1}}
\begin{table}[h] 
\begin{center}
		\caption{Basic characteristics of evaluation datasets.}\label{table_data}
		\vskip 0.1in
    \begin{tabular}{lccc}
    Dataset &               \#obj. & \#pos. & \#dim. \\ \hline
            \datasettable{vehicles} & 435  & 218  & 18 \\     
               \datasettable{heart} & 297  & 137  & 13 \\  
               \datasettable{sonar} & 208  & 97   & 60 \\  
             \datasettable{thyroid}$^*$ & 215  & 65   & 5  \\  
            \datasettable{ringnorm}$^*$ & 1000 & 503  & 20 \\  
          \datasettable{ionosphere} & 351  & 126  & 33 \\ 
					  \datasettable{diabetes} & 768  & 500  & 8  \\  
					   \datasettable{twonorm}$^*$ & 1000 & 500  & 20 \\ 
			        \datasettable{banana}$^*$ & 1000 & 439  & 2  \\  
			     		\datasettable{german} & 1000 & 700  & 20 \\ 
				    	\datasettable{splice} & 1000 & 541  & 60 \\ 
					    \datasettable{breast} & 699 & 458 & 9    
    \end{tabular} 
		\end{center}
		\vskip -0.1in
\end{table}

The parameter settings used are displayed in Table \ref{app_table_benchmark}.

\begin{table}[h]
\center
\caption{Table with parameters used for the benchmark datasets} \label{app_table_benchmark}
\begin{tabular}{l|l|l}
Dataset & $\sigma$ & $\log_{10}(\lambda)$ \\ \hline
\verb-vehicles- & 5.270 & -3.0 \\
\verb-heart- & 5.906 & -1.8 \\
\verb-sonar- & 7.084 & -2.6 \\
\verb-thyroid- & 1.720 & -2.6 \\
\verb-ringnorm- & 1.778 & -3.0 \\
\verb-ionosphere- & 4.655 & -2.2 \\
\verb-diabetes- & 2.955 & -1.4 \\
\verb-twonorm- & 5.299 & -2.2 \\
\verb-banana- & 0.645 & -2.2 \\
\verb-german- & 4.217 & -1.4 \\
\verb-splice- & 9.481 & -2.6 \\
\verb-breast- & 4.217 & -1.8 \\
\end{tabular}
\end{table}

\FloatBarrier
\newpage
\section{Learning Curves of Agnostic Setting}

\begin{figure}[h]
    \centering
    \begin{subfigure}[b]{0.3\textwidth}
        \includegraphics[width=\textwidth]{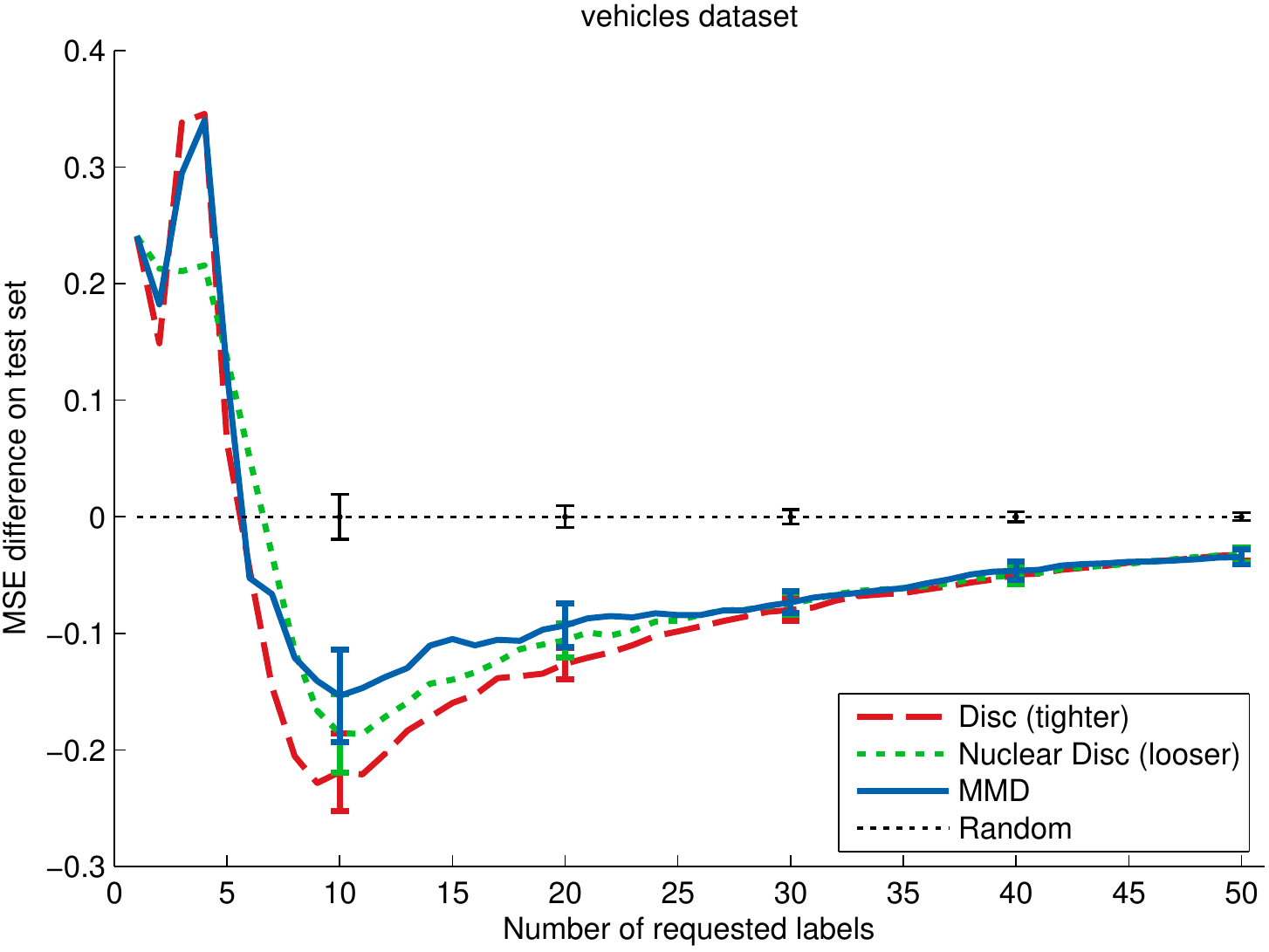}
    \end{subfigure}
    ~ 
    \begin{subfigure}[b]{0.3\textwidth}
        \includegraphics[width=\textwidth]{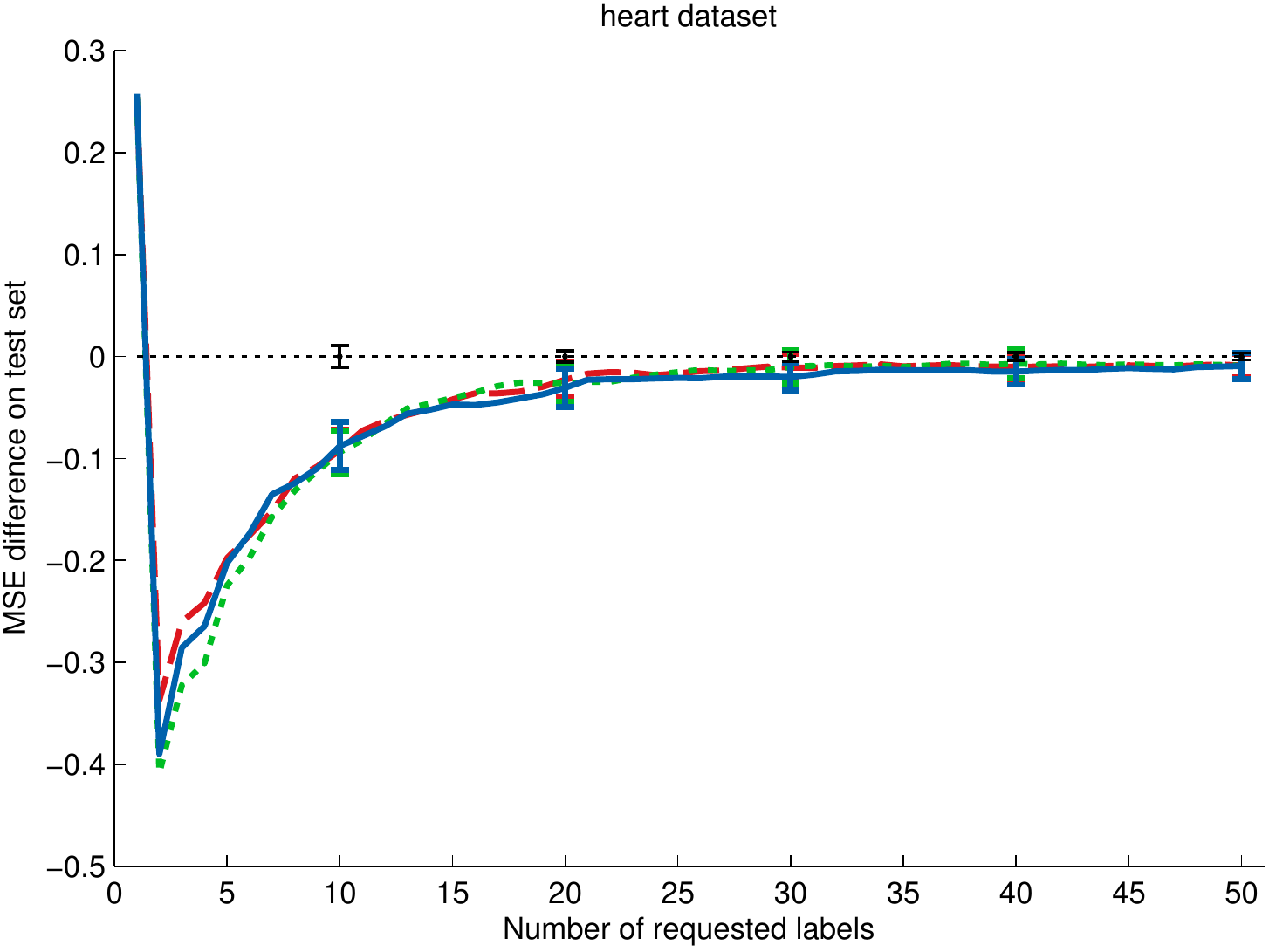}
    \end{subfigure}
		~
		\begin{subfigure}[b]{0.3\textwidth}
        \includegraphics[width=\textwidth]{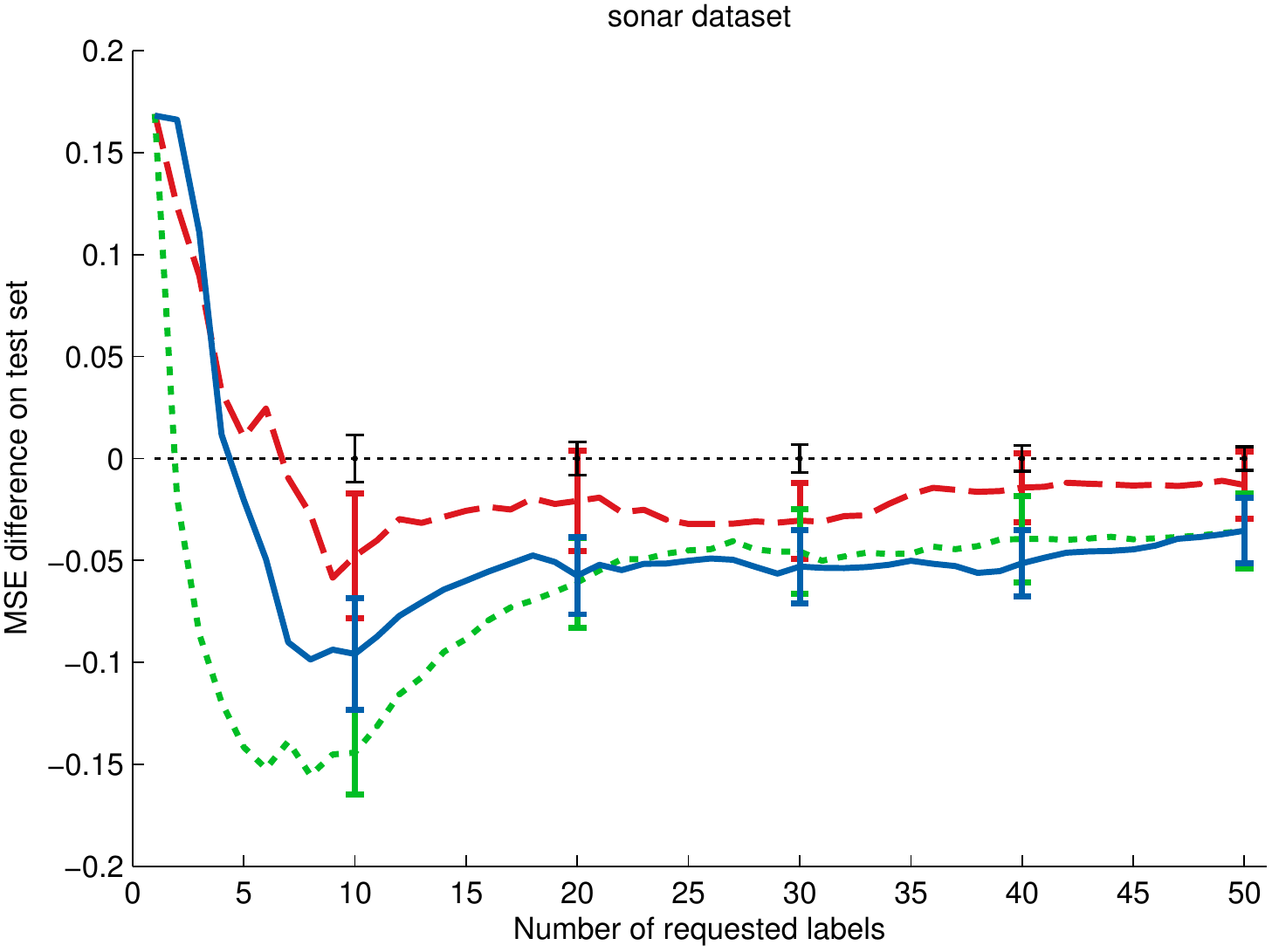}
    \end{subfigure}
		~
		\begin{subfigure}[b]{0.3\textwidth}
        \includegraphics[width=\textwidth]{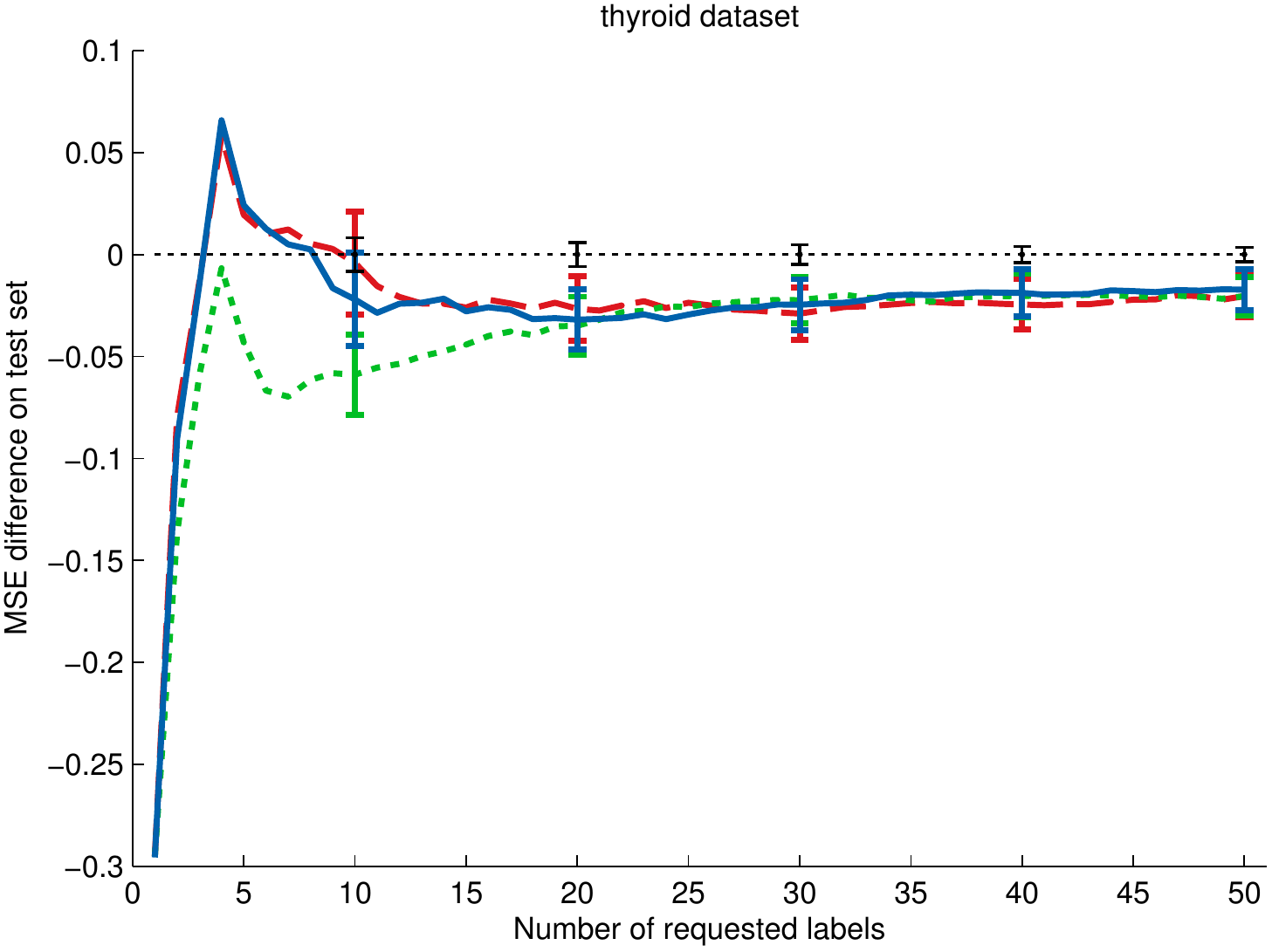}
    \end{subfigure}
		~
		\begin{subfigure}[b]{0.3\textwidth}
        \includegraphics[width=\textwidth]{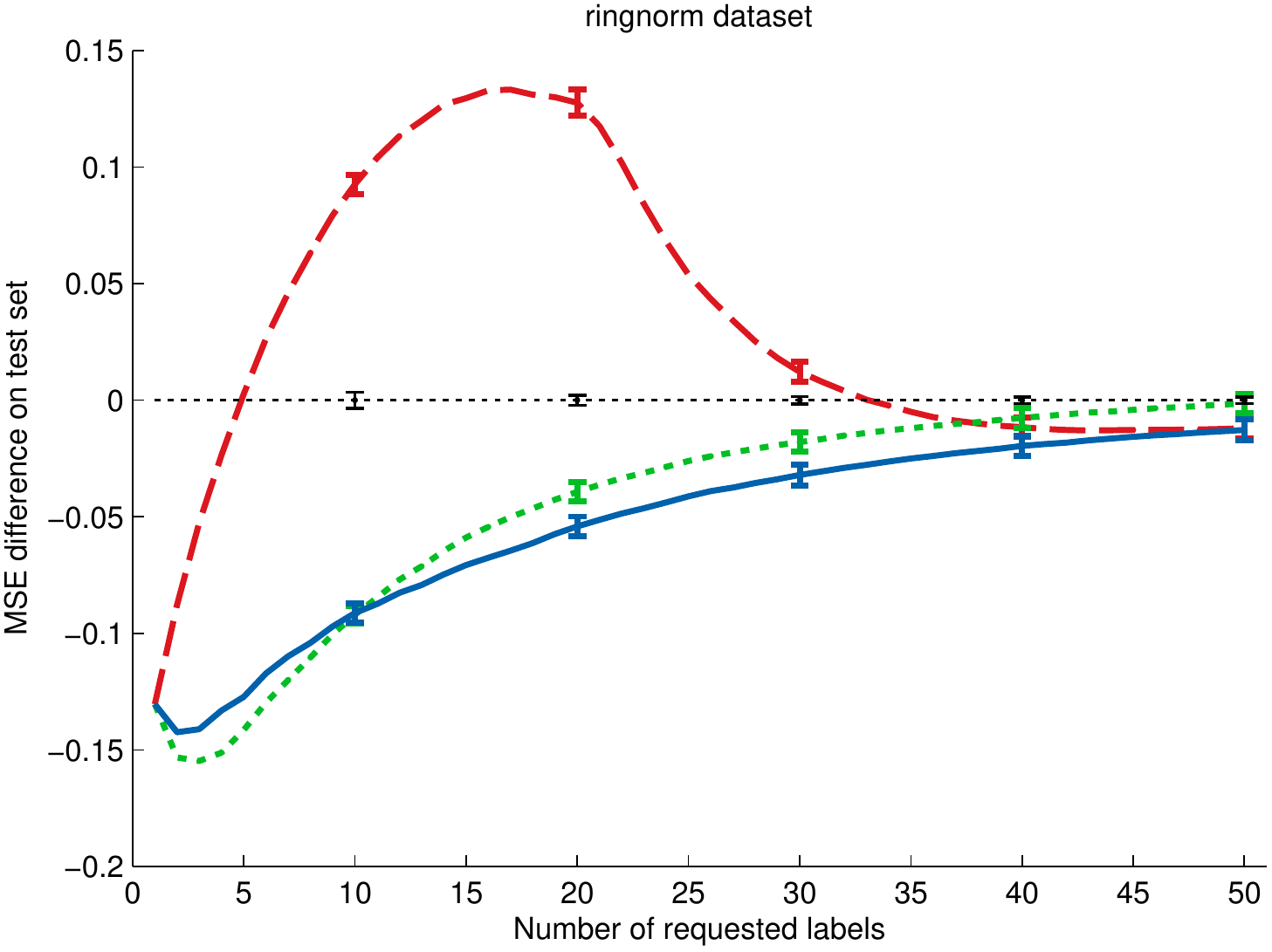}
    \end{subfigure}
		~
		\begin{subfigure}[b]{0.3\textwidth}
        \includegraphics[width=\textwidth]{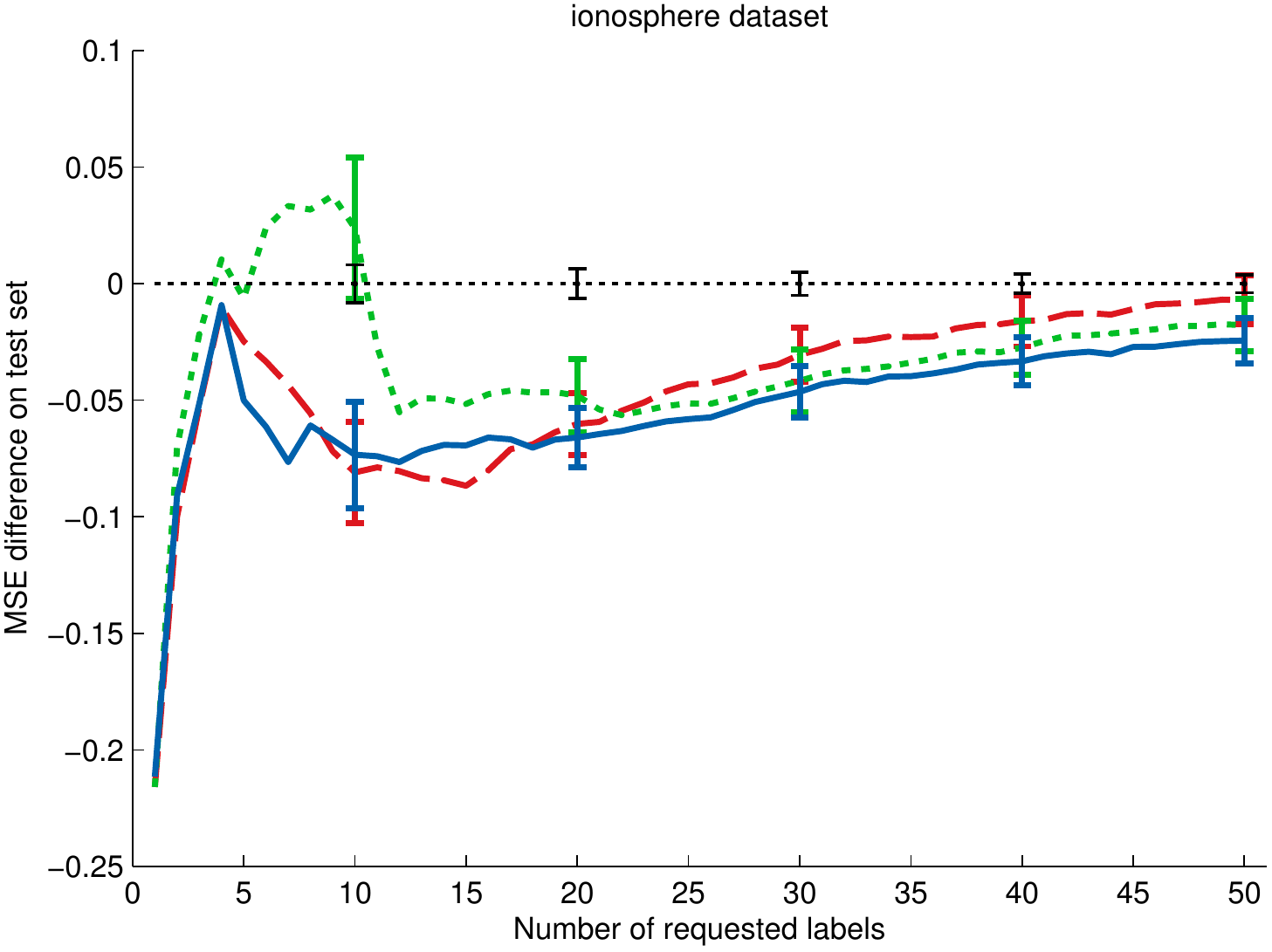}
    \end{subfigure}
		~
		\begin{subfigure}[b]{0.3\textwidth}
        \includegraphics[width=\textwidth]{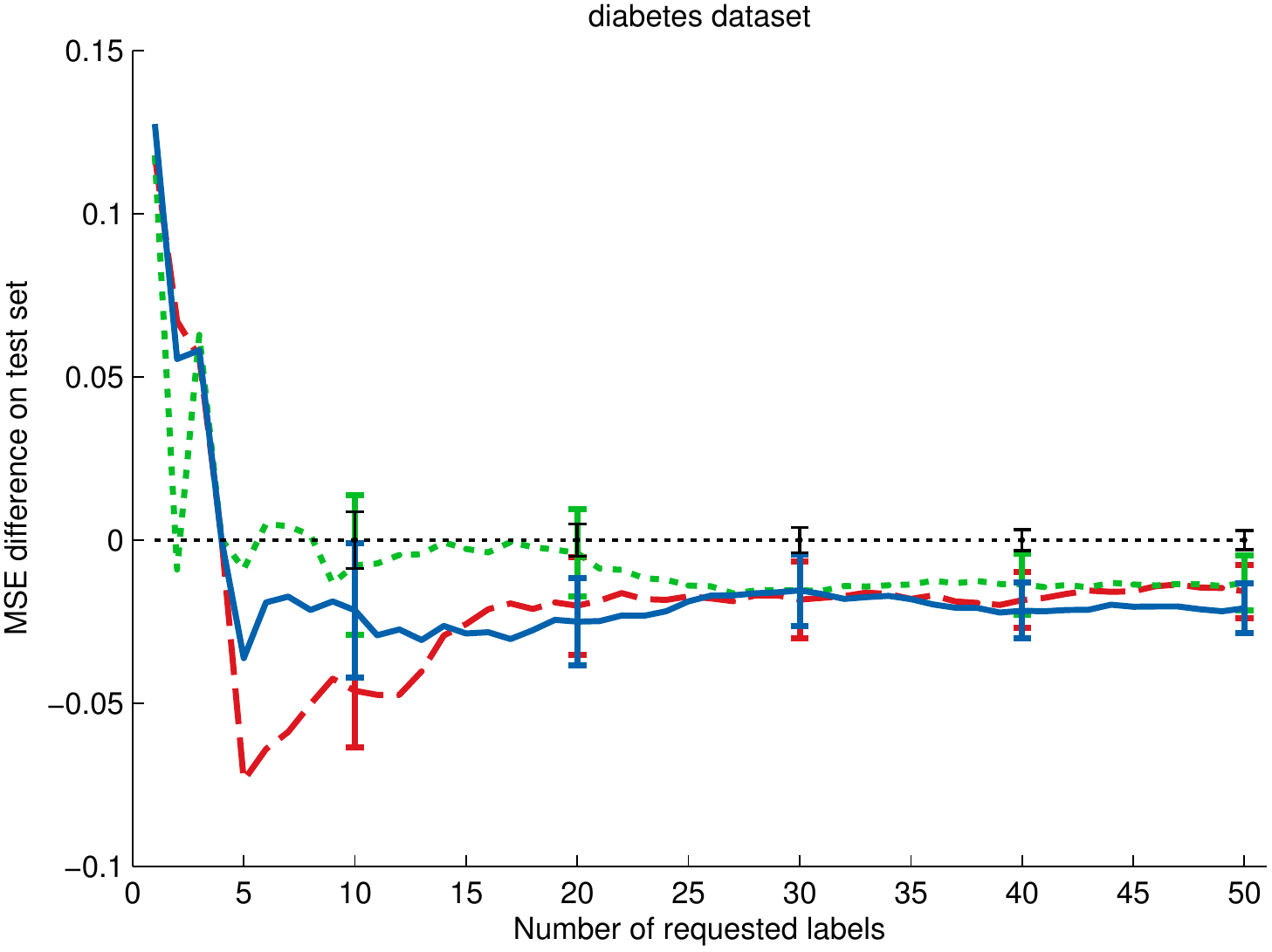}
    \end{subfigure}
		~
		\begin{subfigure}[b]{0.3\textwidth}
        \includegraphics[width=\textwidth]{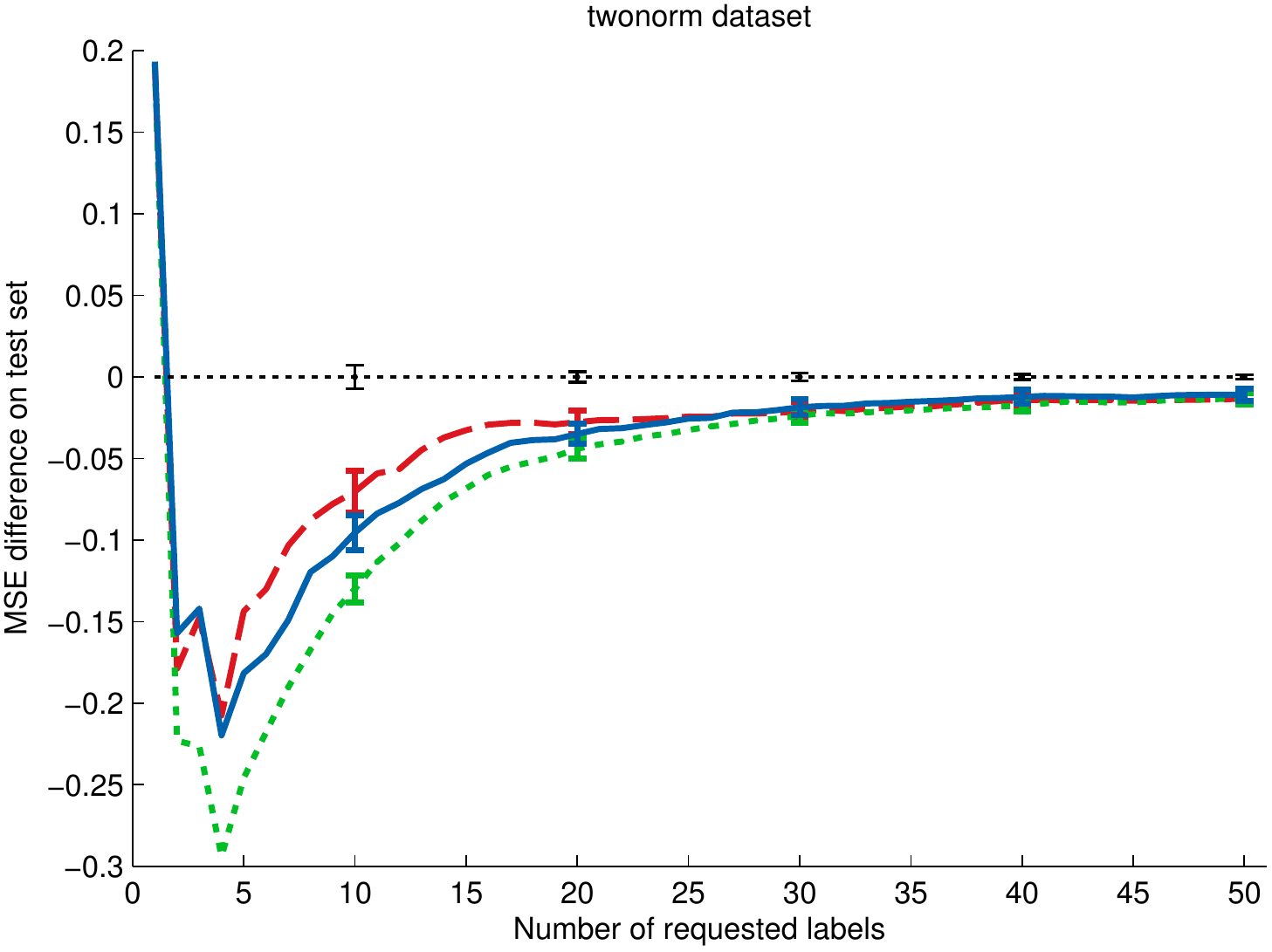}
    \end{subfigure}
		~
		\begin{subfigure}[b]{0.3\textwidth}
        \includegraphics[width=\textwidth]{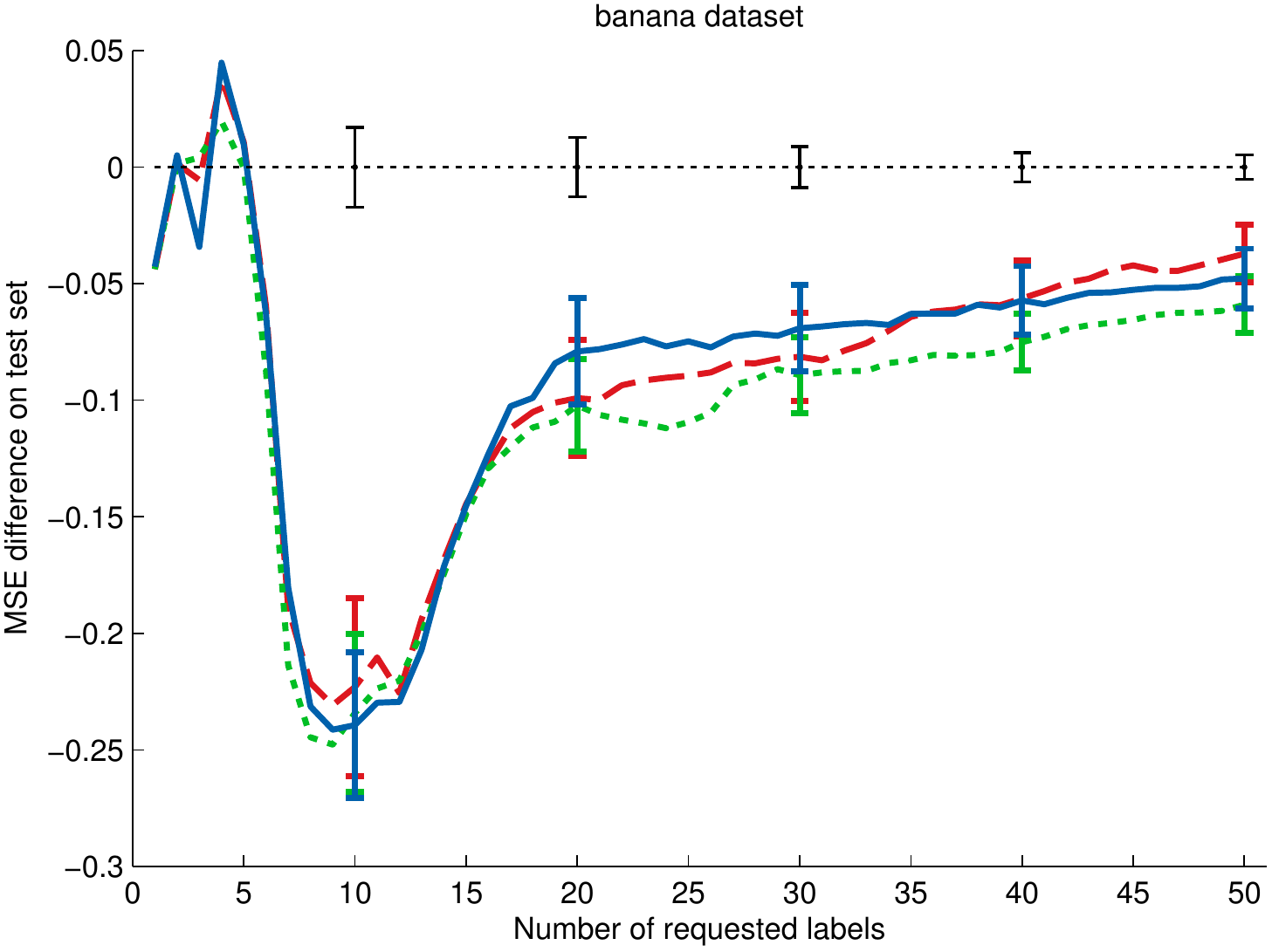}
    \end{subfigure}
		~
		\begin{subfigure}[b]{0.3\textwidth}
        \includegraphics[width=\textwidth]{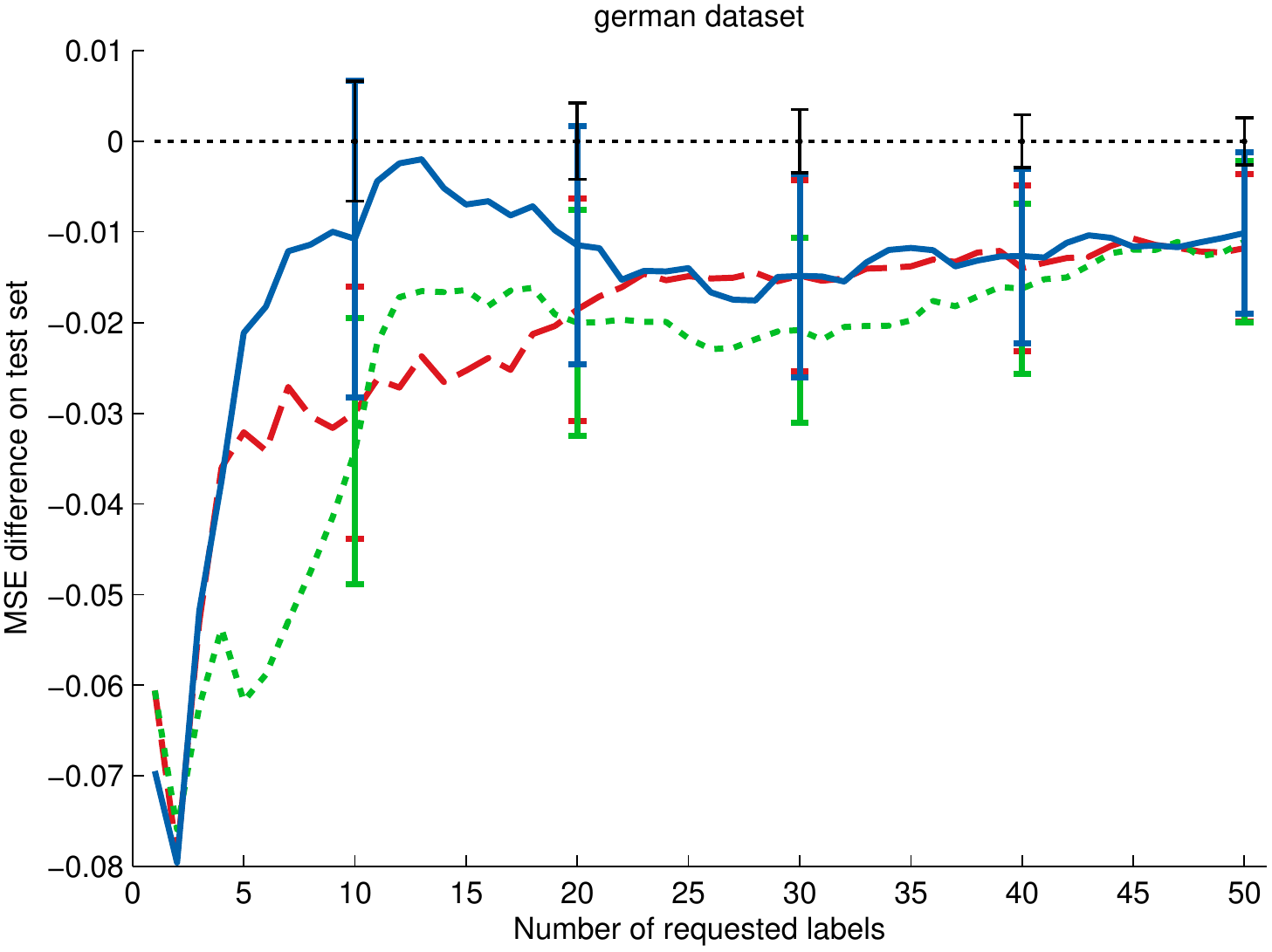}
    \end{subfigure}
		~
		\begin{subfigure}[b]{0.3\textwidth}
        \includegraphics[width=\textwidth]{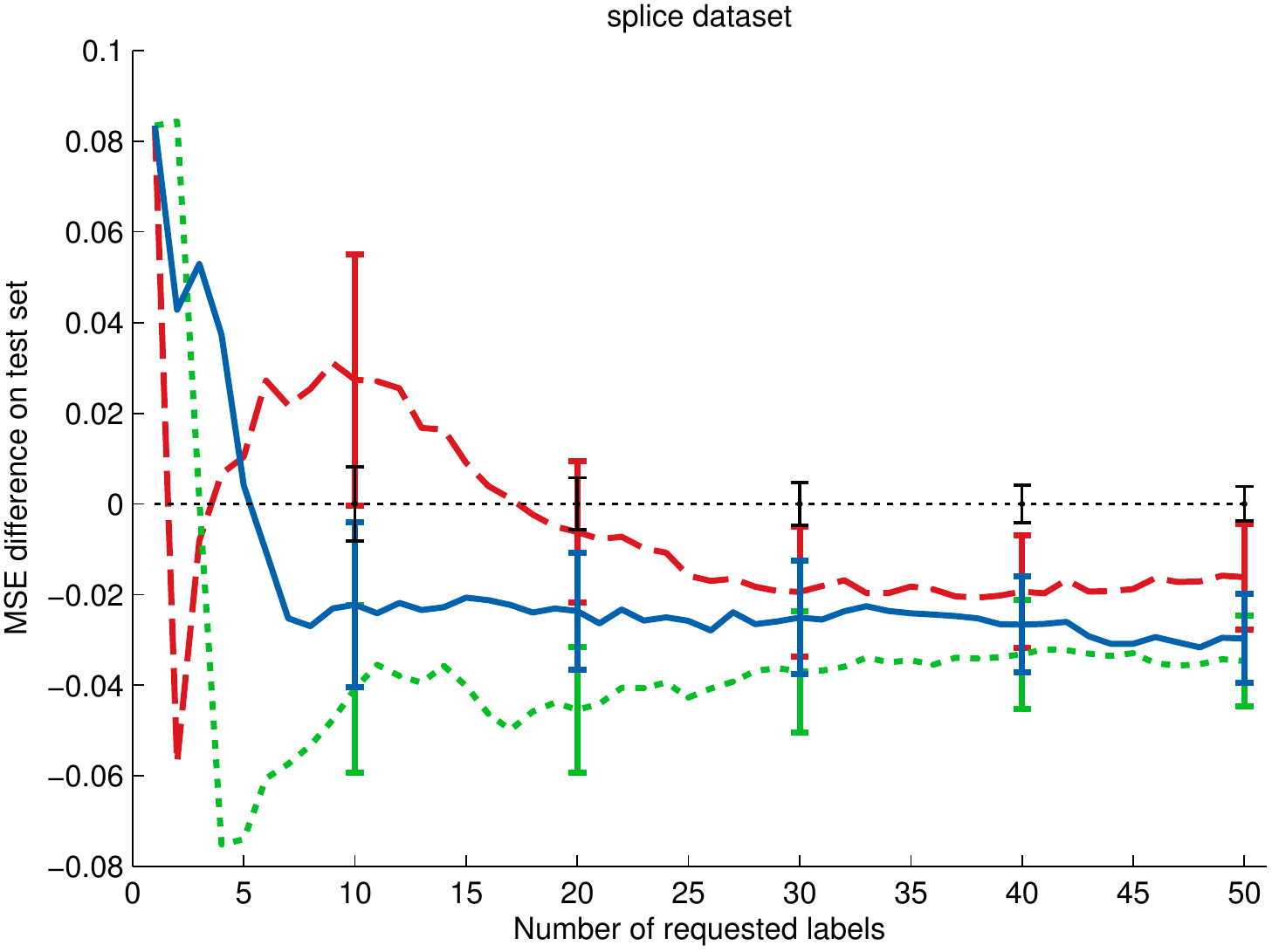}
    \end{subfigure}
		~
		\begin{subfigure}[b]{0.3\textwidth}
        \includegraphics[width=\textwidth]{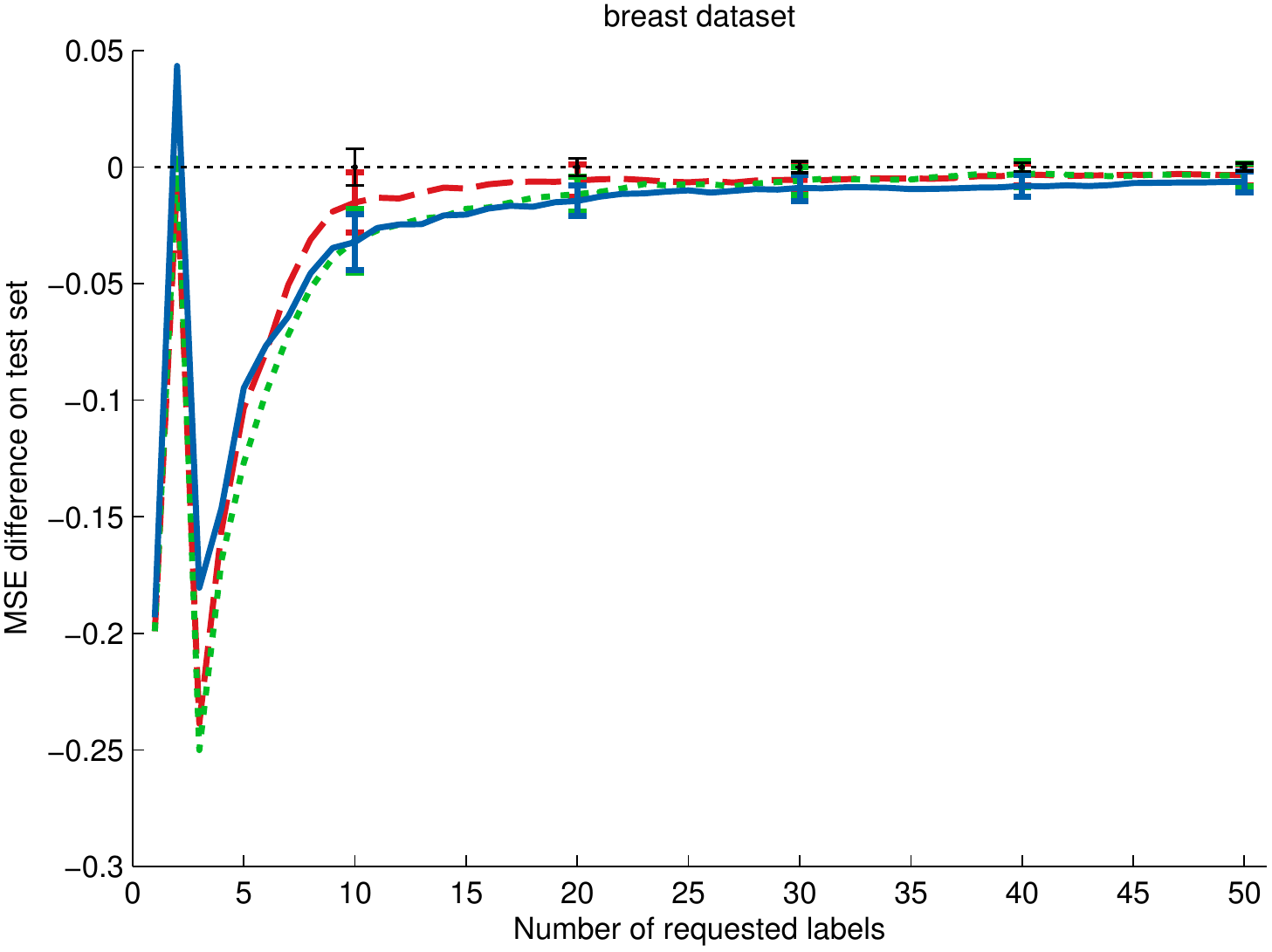}
    \end{subfigure}
    \caption{Results on all benchmark datasets for the agnostic setting.}\label{lc_agnostic}
\end{figure}

\FloatBarrier
\newpage
\section{Additional Results on All Datasets for the Realizeable Setting}
~
\begin{figure}[h]
    \centering
    \begin{subfigure}[b]{0.3\textwidth}
        \includegraphics[width=\textwidth]{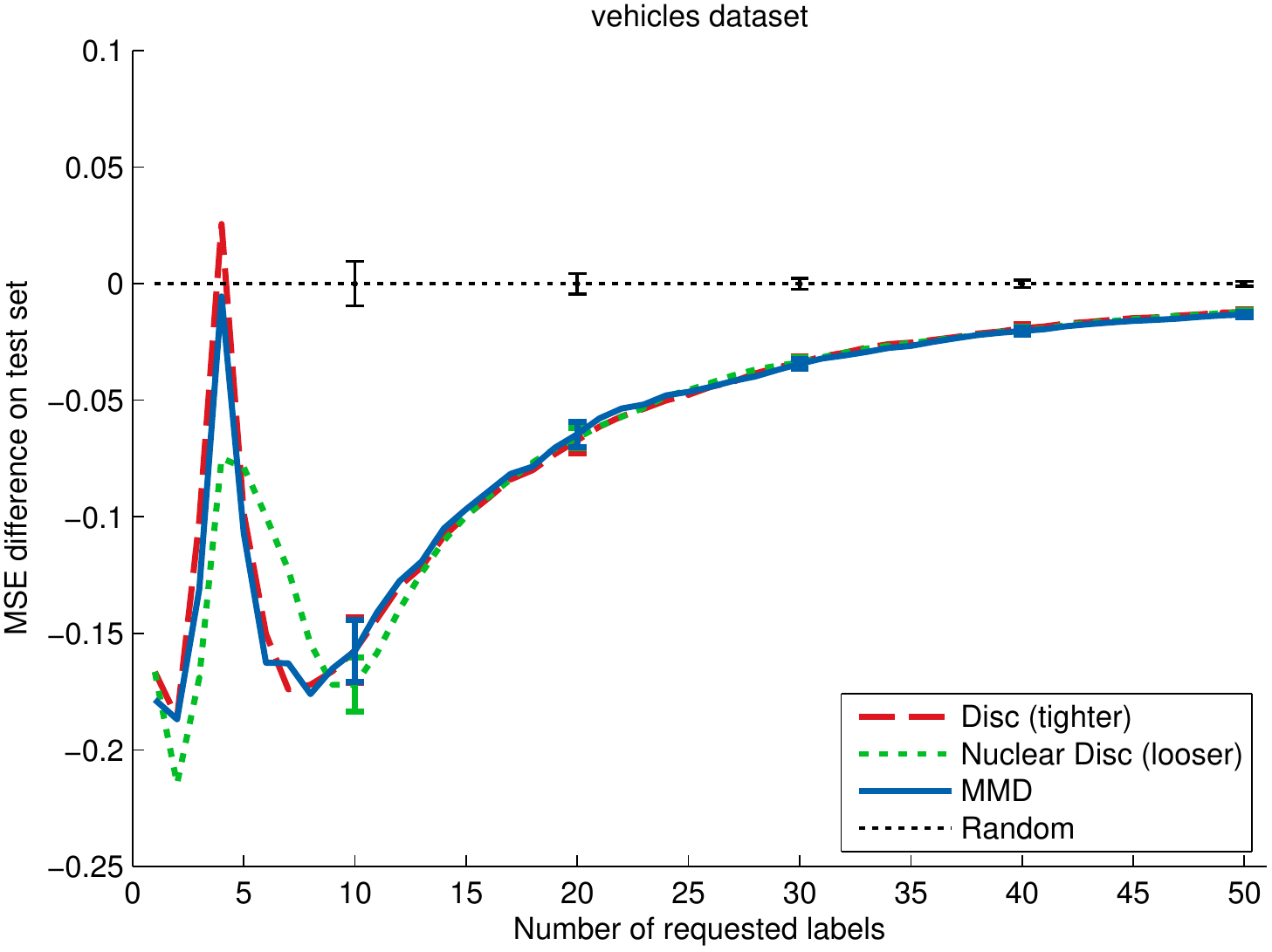}
    \end{subfigure}
    ~ 
    \begin{subfigure}[b]{0.3\textwidth}
        \includegraphics[width=\textwidth]{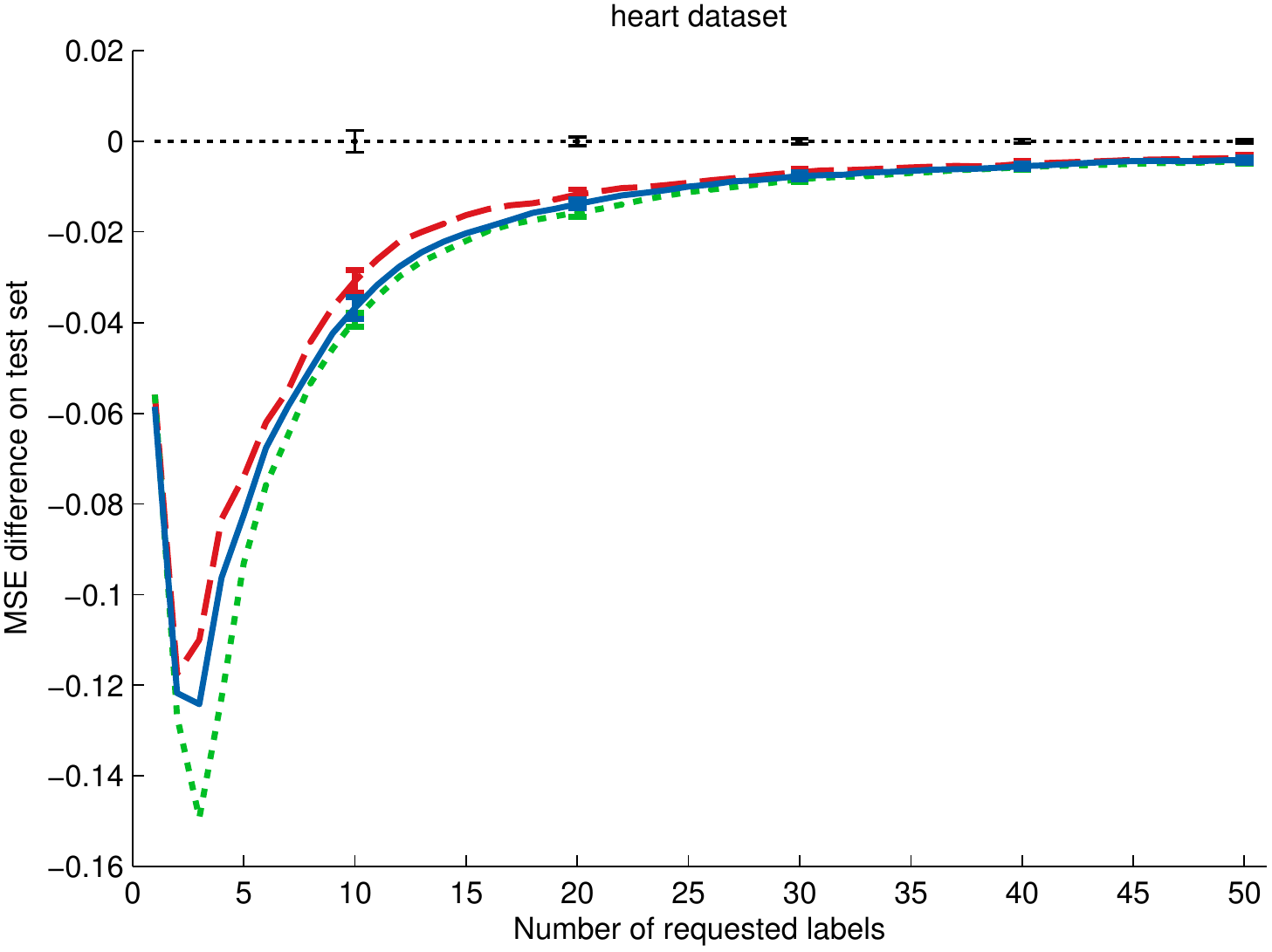}
    \end{subfigure}
		~
		\begin{subfigure}[b]{0.3\textwidth}
        \includegraphics[width=\textwidth]{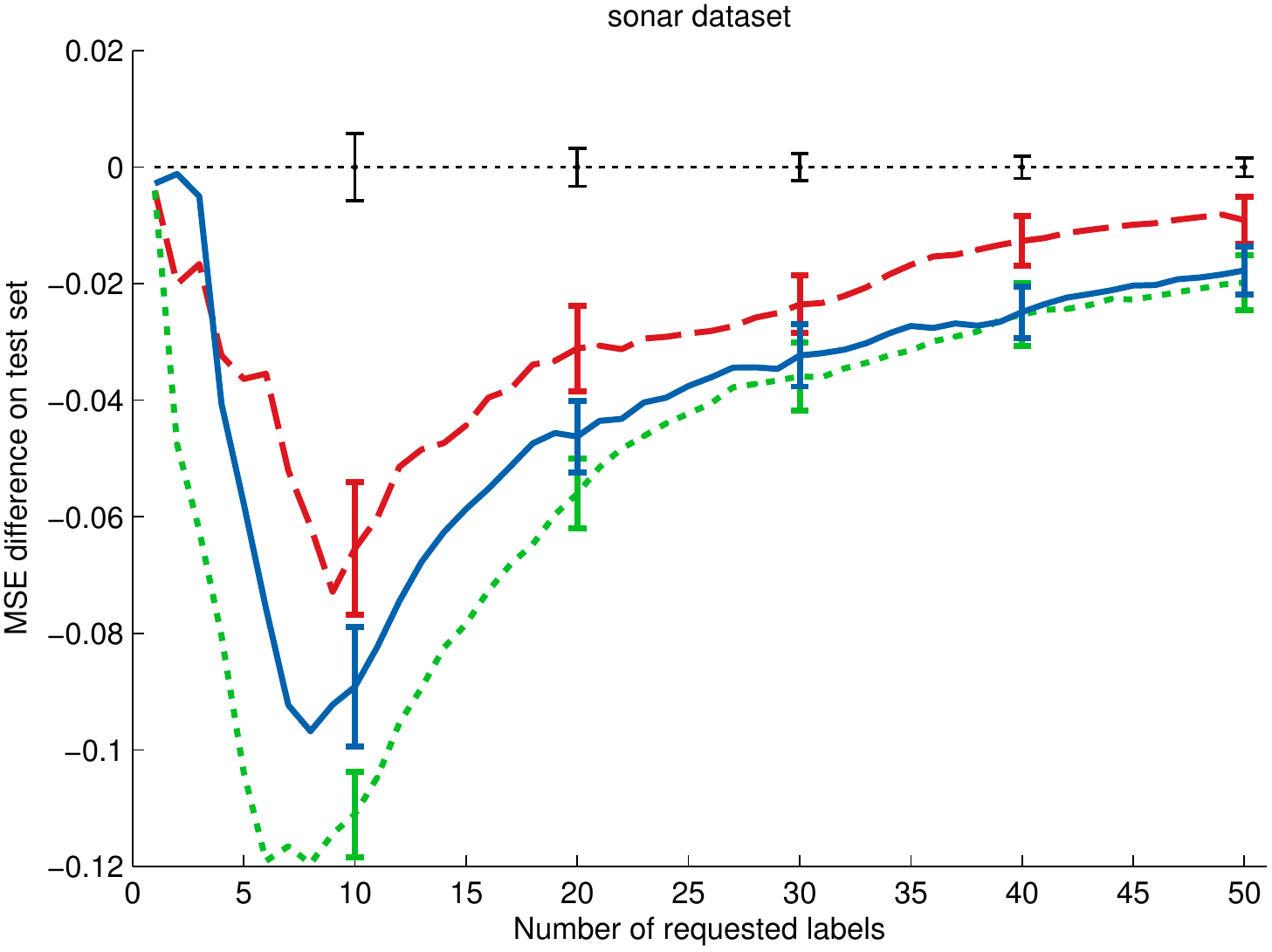}
    \end{subfigure}
		~
		\begin{subfigure}[b]{0.3\textwidth}
        \includegraphics[width=\textwidth]{lc_new/lc_nuclear1_agnostic0_5_legend0.pdf}
    \end{subfigure}
		~
		\begin{subfigure}[b]{0.3\textwidth}
        \includegraphics[width=\textwidth]{lc_new/lc_nuclear1_agnostic0_6_legend0.pdf}
    \end{subfigure}
		~
		\begin{subfigure}[b]{0.3\textwidth}
        \includegraphics[width=\textwidth]{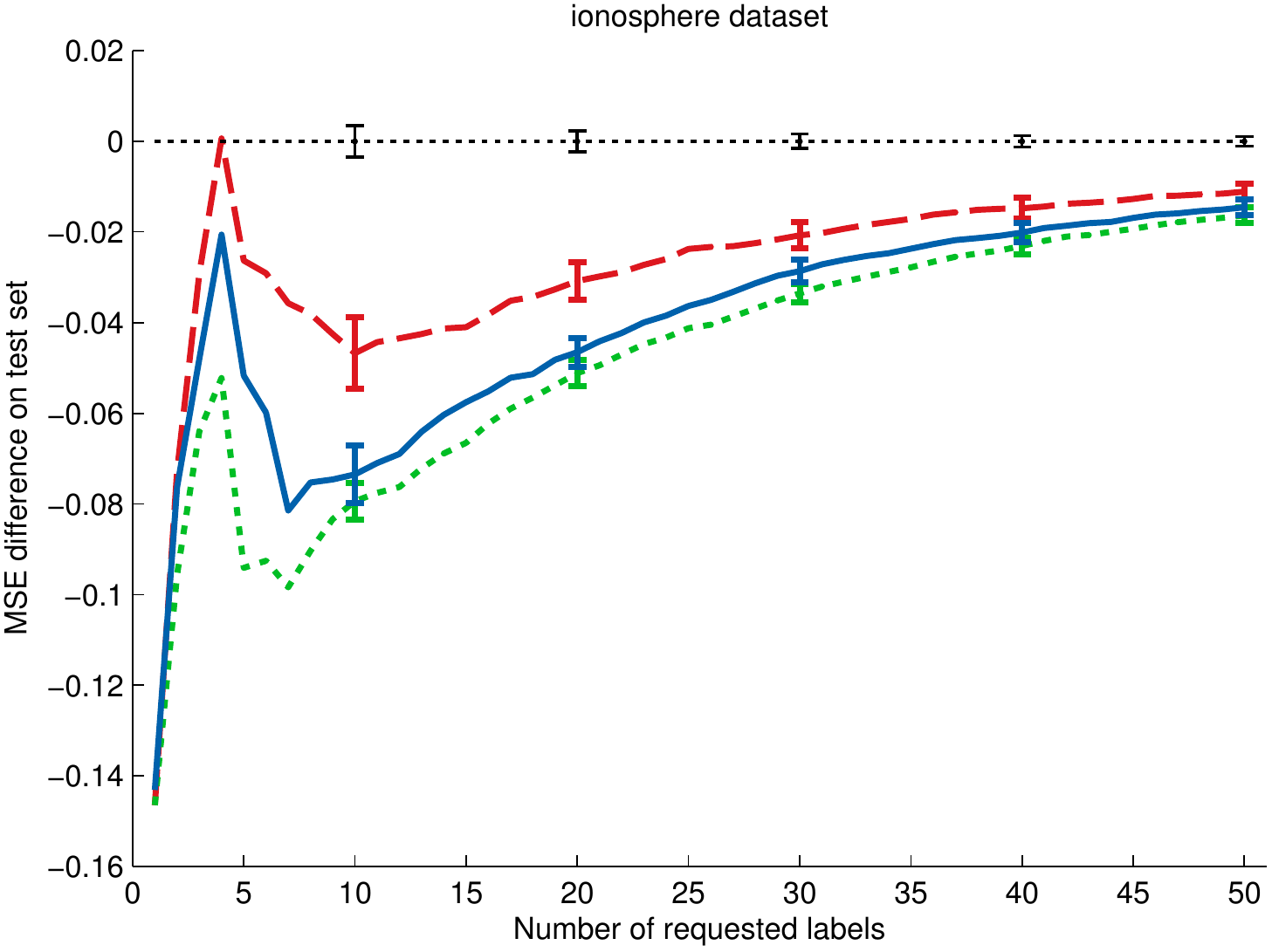}
    \end{subfigure}
		~
		\begin{subfigure}[b]{0.3\textwidth}
        \includegraphics[width=\textwidth]{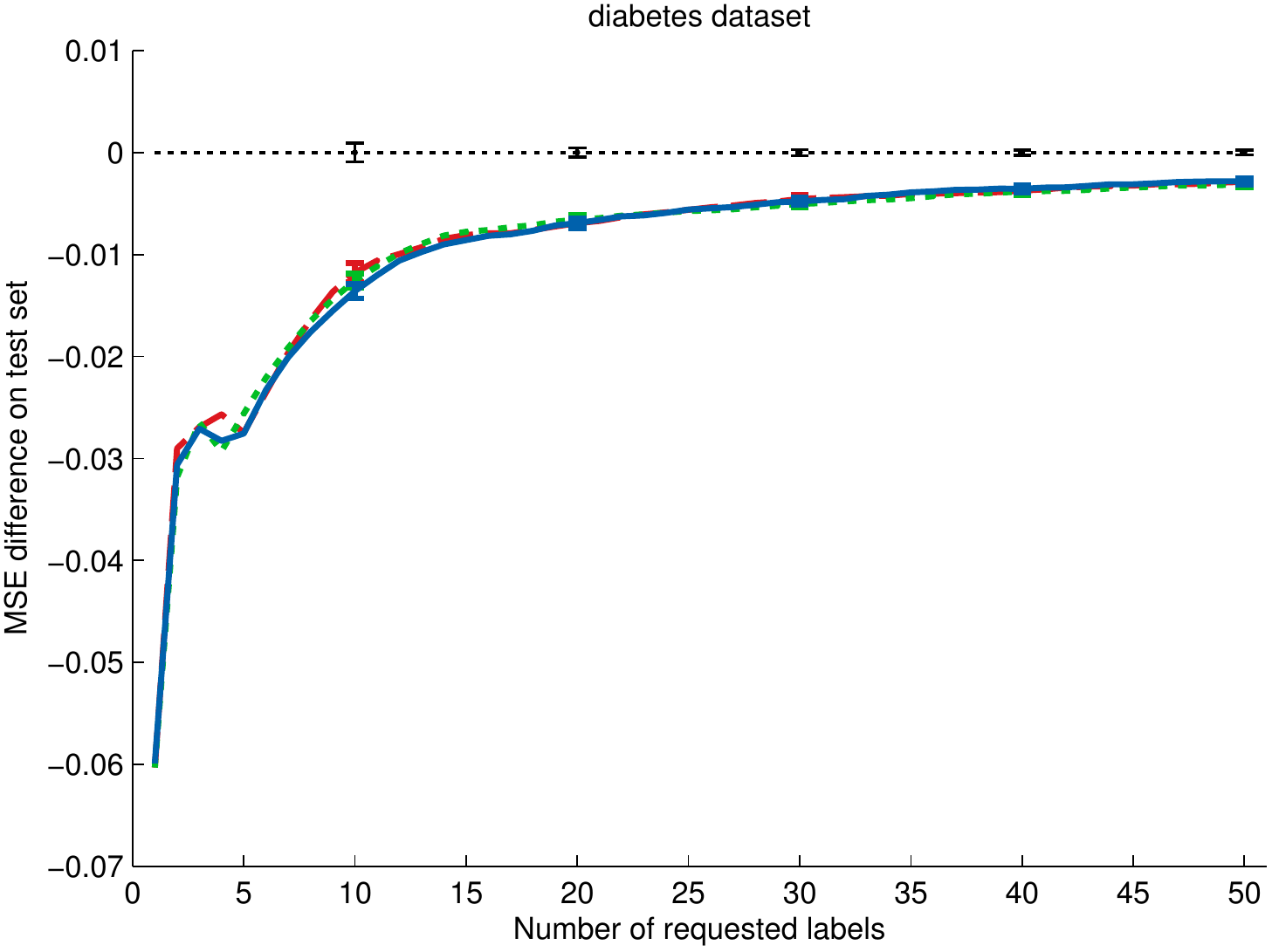}
    \end{subfigure}
		~
		\begin{subfigure}[b]{0.3\textwidth}
        \includegraphics[width=\textwidth]{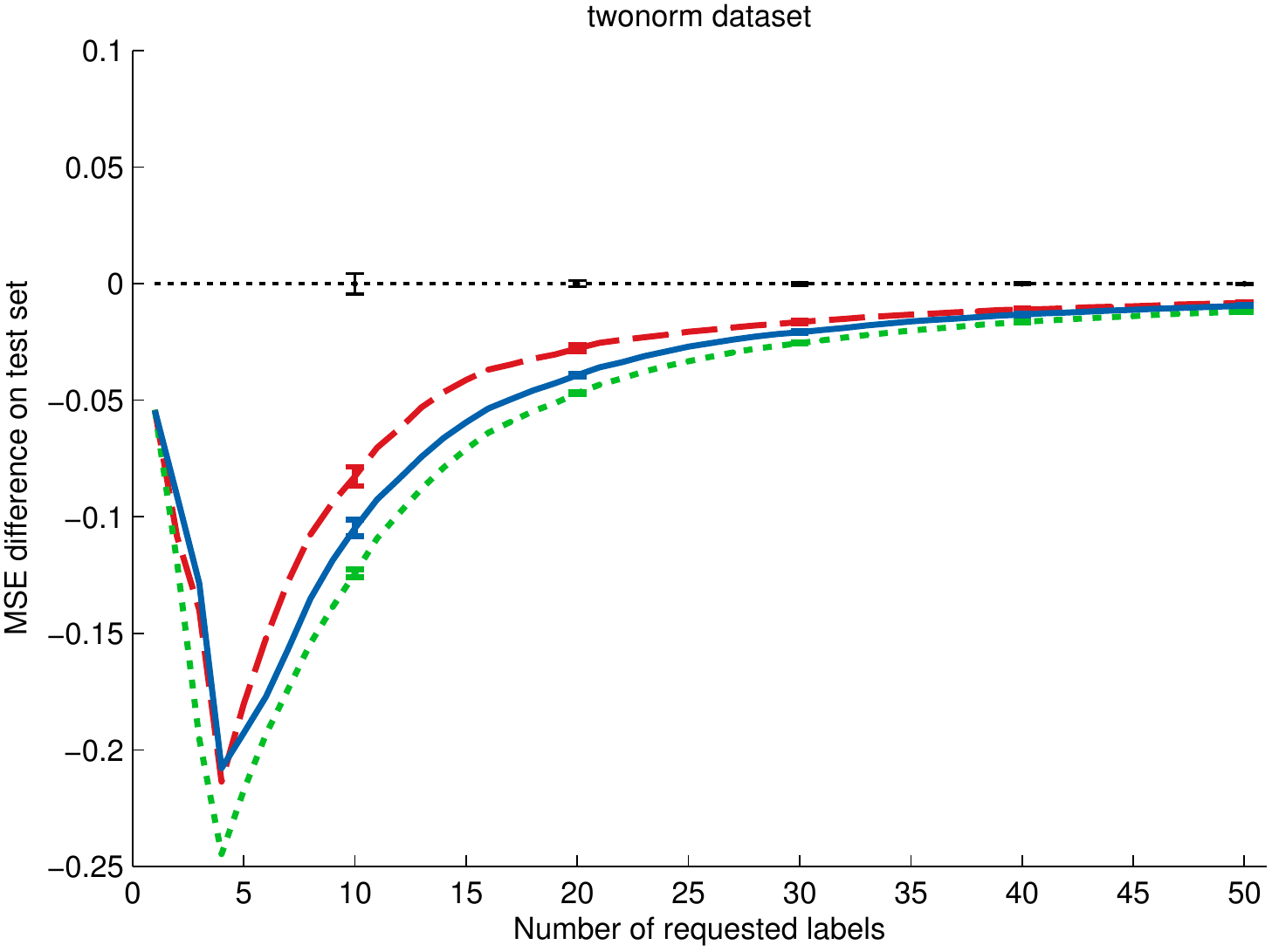}
    \end{subfigure}
		~
		\begin{subfigure}[b]{0.3\textwidth}
        \includegraphics[width=\textwidth]{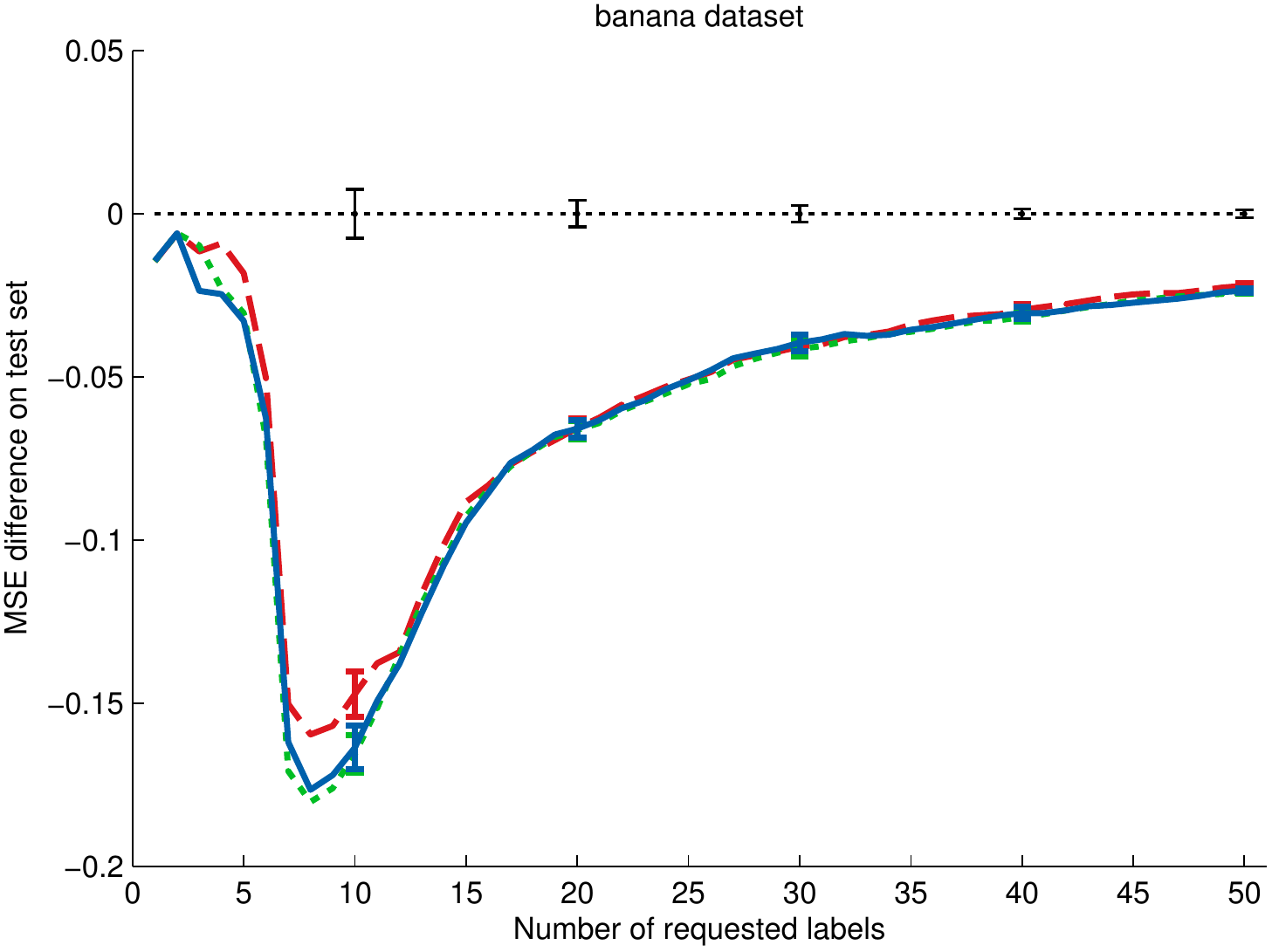}
    \end{subfigure}
		~
		\begin{subfigure}[b]{0.3\textwidth}
        \includegraphics[width=\textwidth]{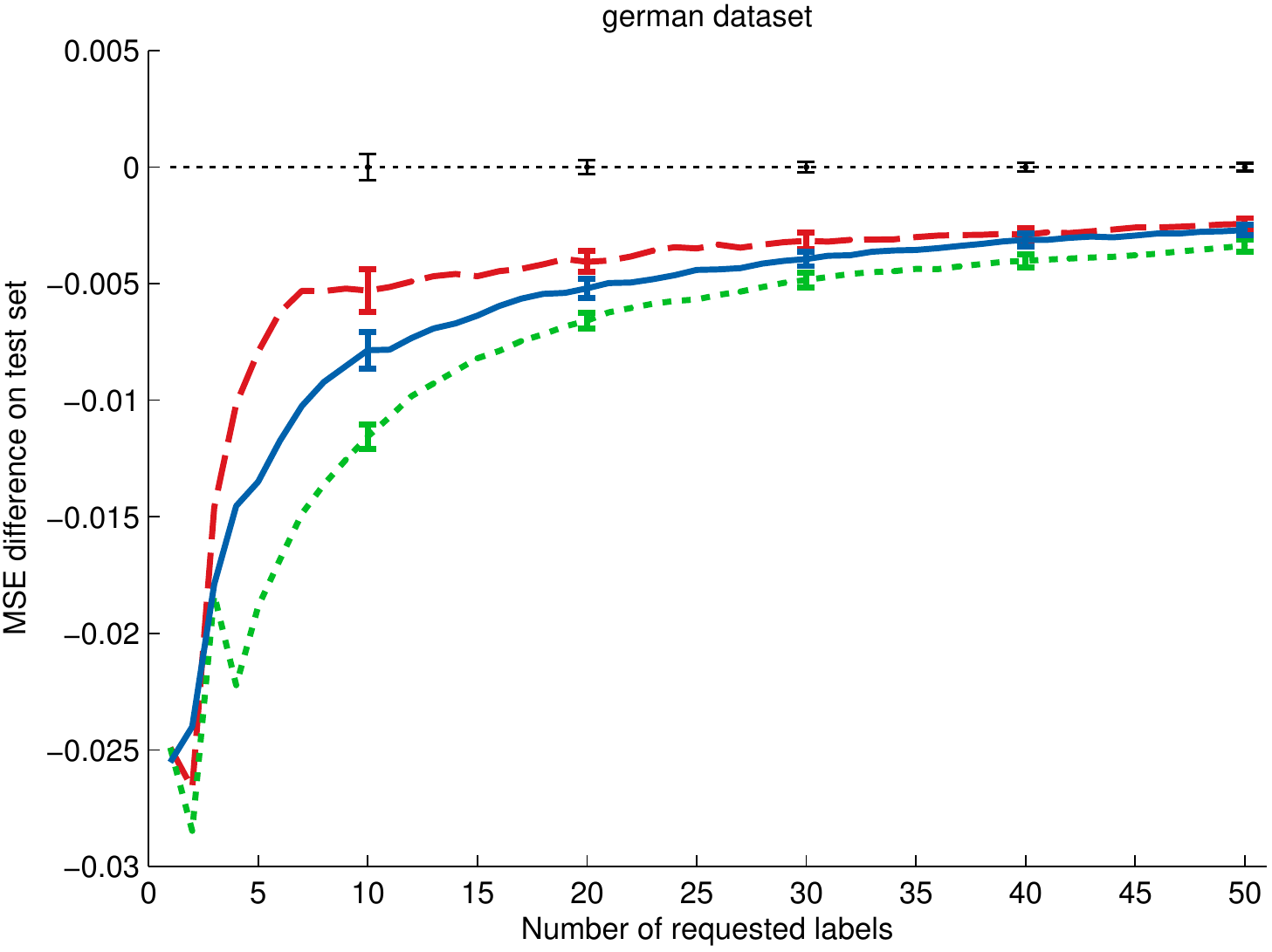}
    \end{subfigure}
		~
		\begin{subfigure}[b]{0.3\textwidth}
        \includegraphics[width=\textwidth]{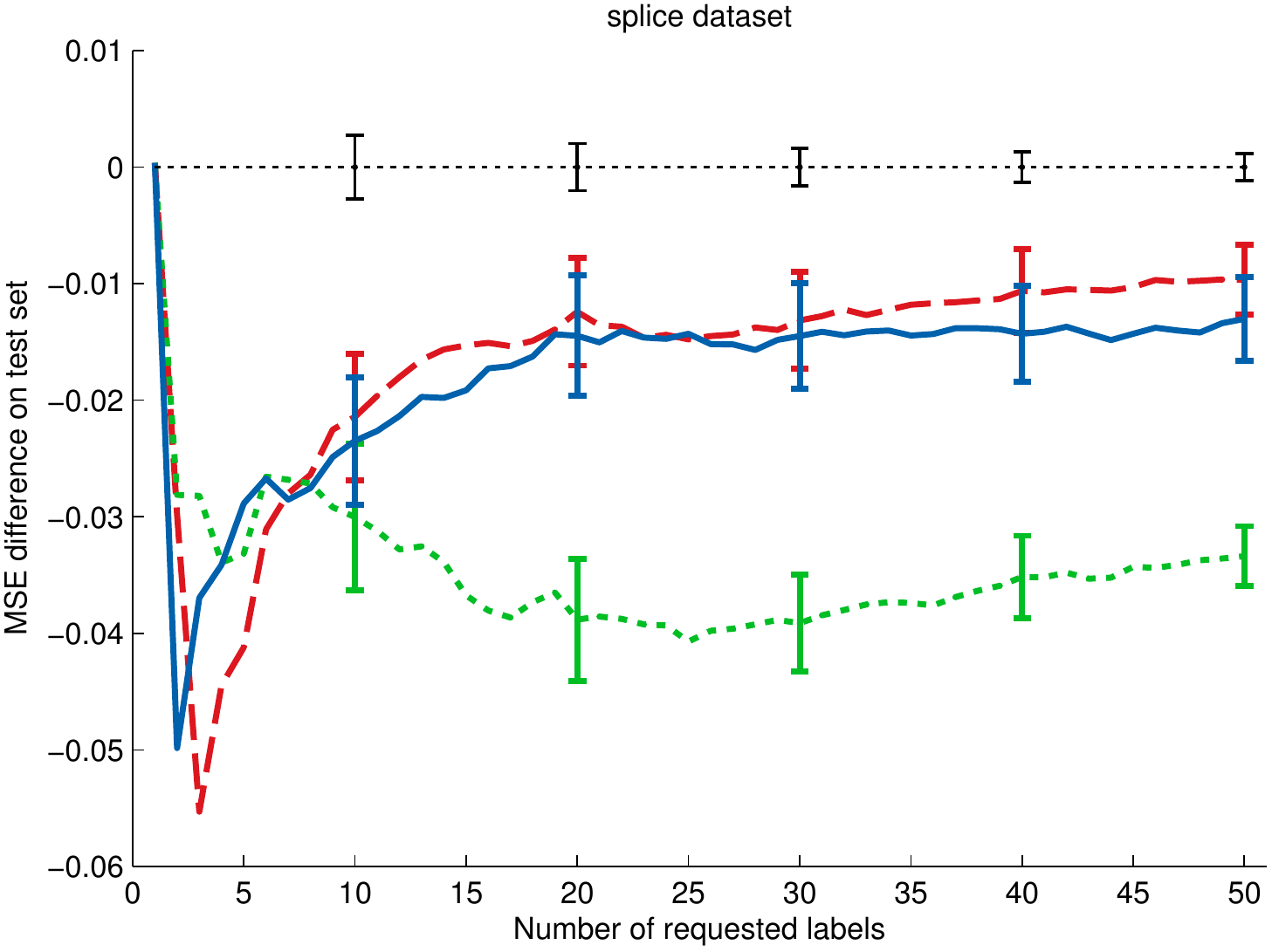}
    \end{subfigure}
		~
		\begin{subfigure}[b]{0.3\textwidth}
        \includegraphics[width=\textwidth]{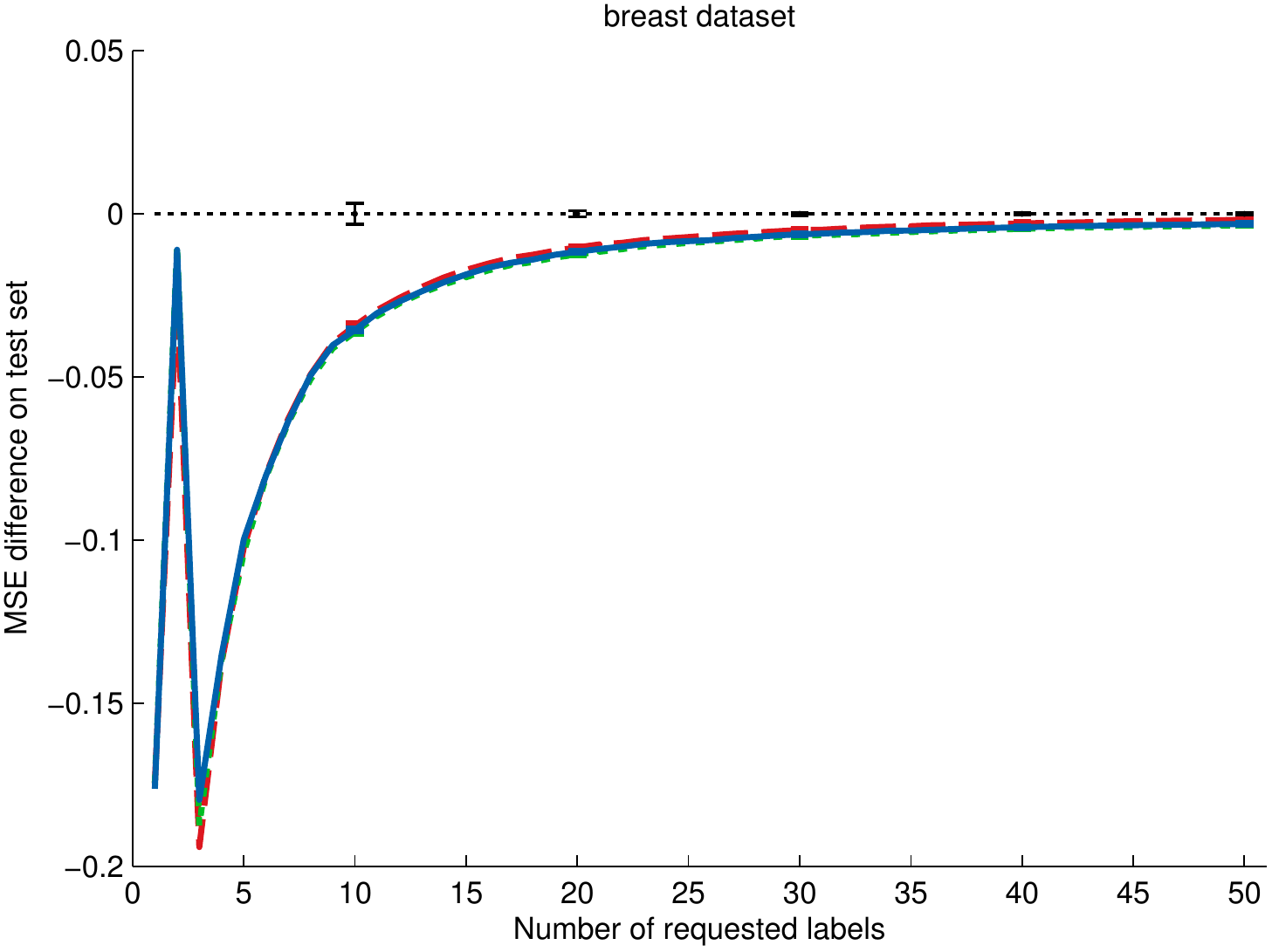}
    \end{subfigure}
    \caption{Results on all benchmark datasets for the realizable setting where $f \in H$.}\label{lc_realizeable}
\end{figure}
~
\begin{figure}[h]
    \centering
    \begin{subfigure}[b]{0.3\textwidth}
        \includegraphics[width=\textwidth]{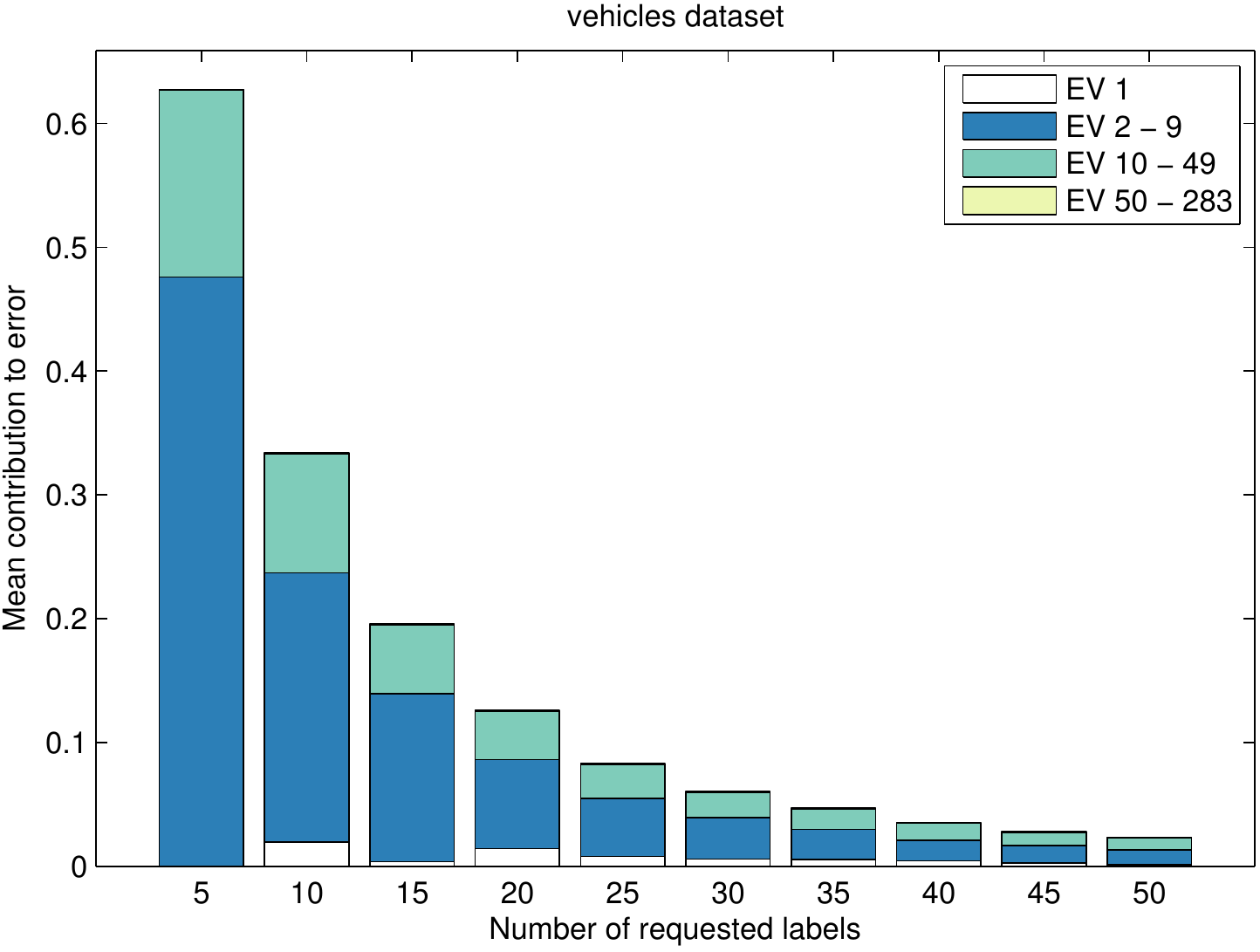}
    \end{subfigure}
    ~ 
    \begin{subfigure}[b]{0.3\textwidth}
        \includegraphics[width=\textwidth]{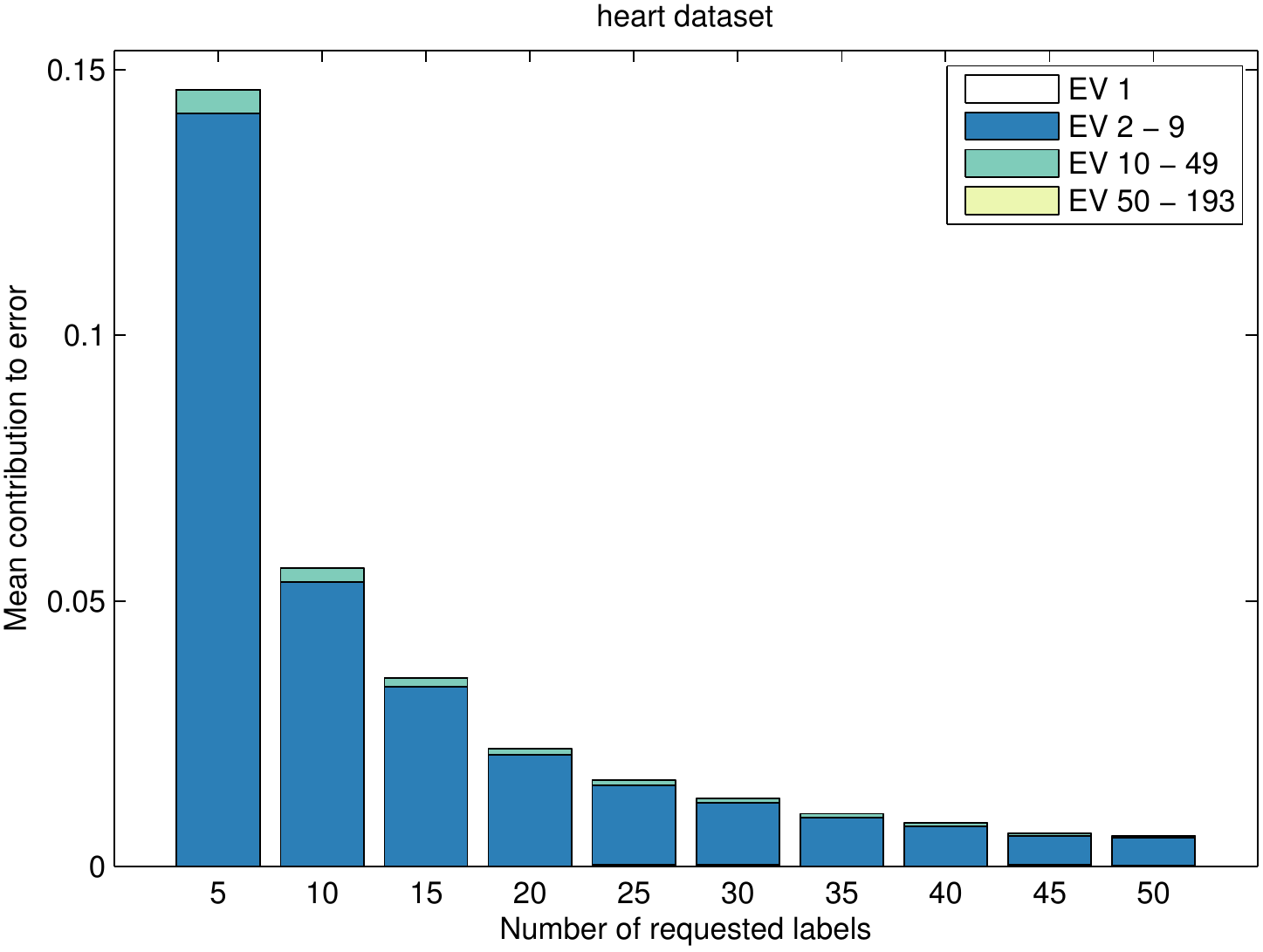}
    \end{subfigure}
		~
		\begin{subfigure}[b]{0.3\textwidth}
        \includegraphics[width=\textwidth]{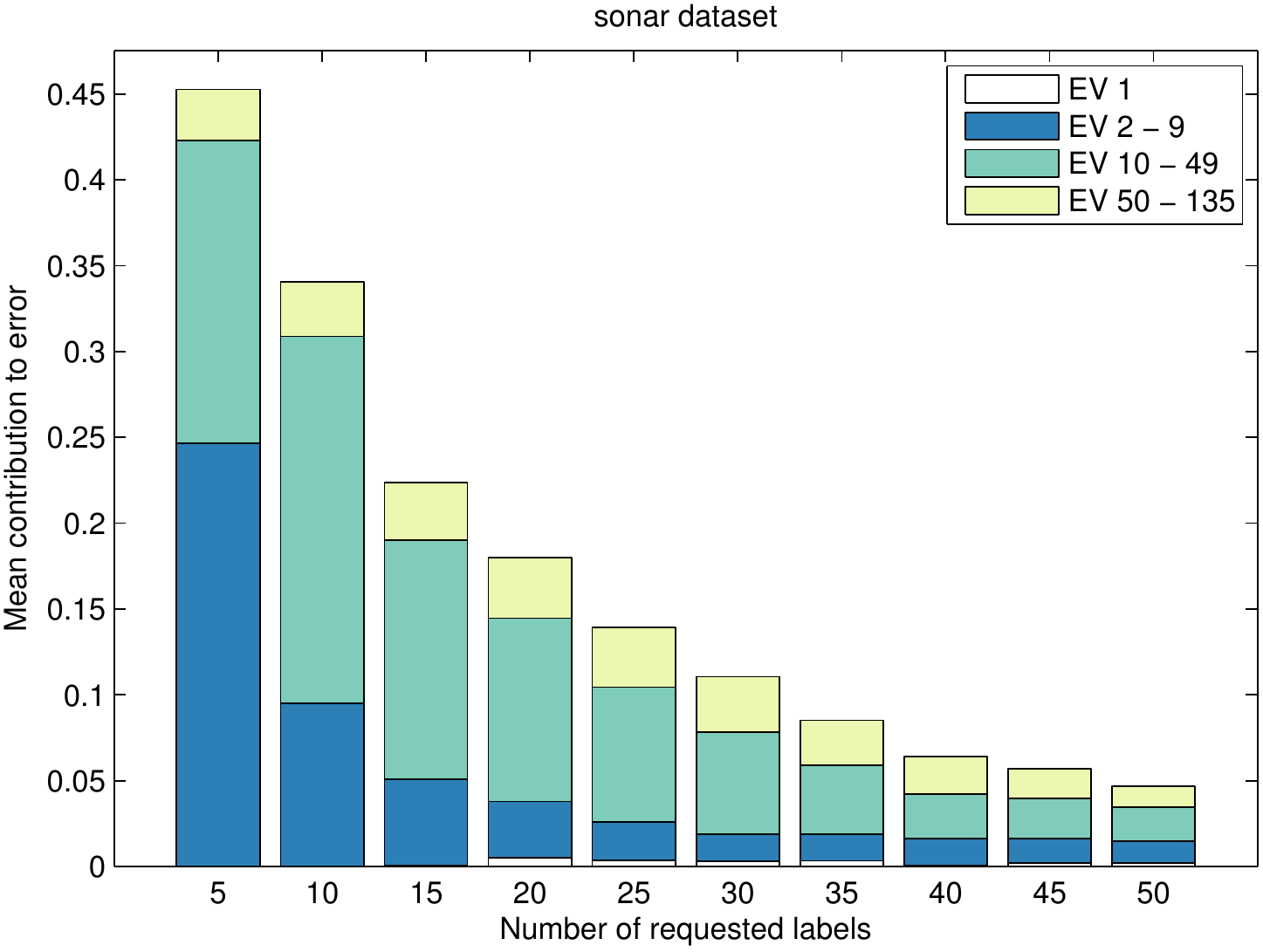}
    \end{subfigure}
		~
		\begin{subfigure}[b]{0.3\textwidth}
        \includegraphics[width=\textwidth]{staaf_new/contribution_5_Random_legend.pdf}
    \end{subfigure}
		~
		\begin{subfigure}[b]{0.3\textwidth}
        \includegraphics[width=\textwidth]{staaf_new/contribution_6_Random_legend.pdf}
    \end{subfigure}
		~
		\begin{subfigure}[b]{0.3\textwidth}
        \includegraphics[width=\textwidth]{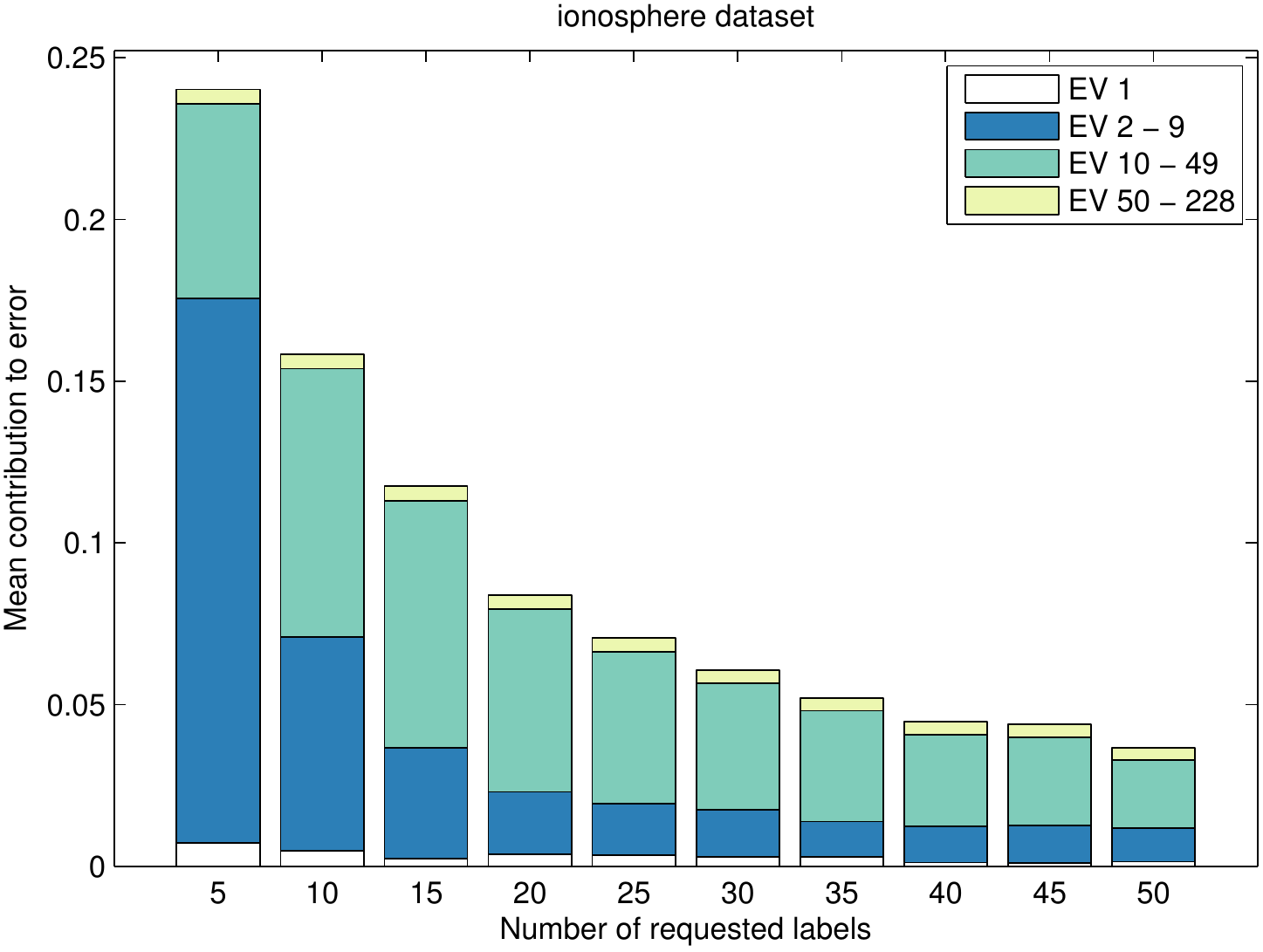}
    \end{subfigure}
		~
		\begin{subfigure}[b]{0.3\textwidth}
        \includegraphics[width=\textwidth]{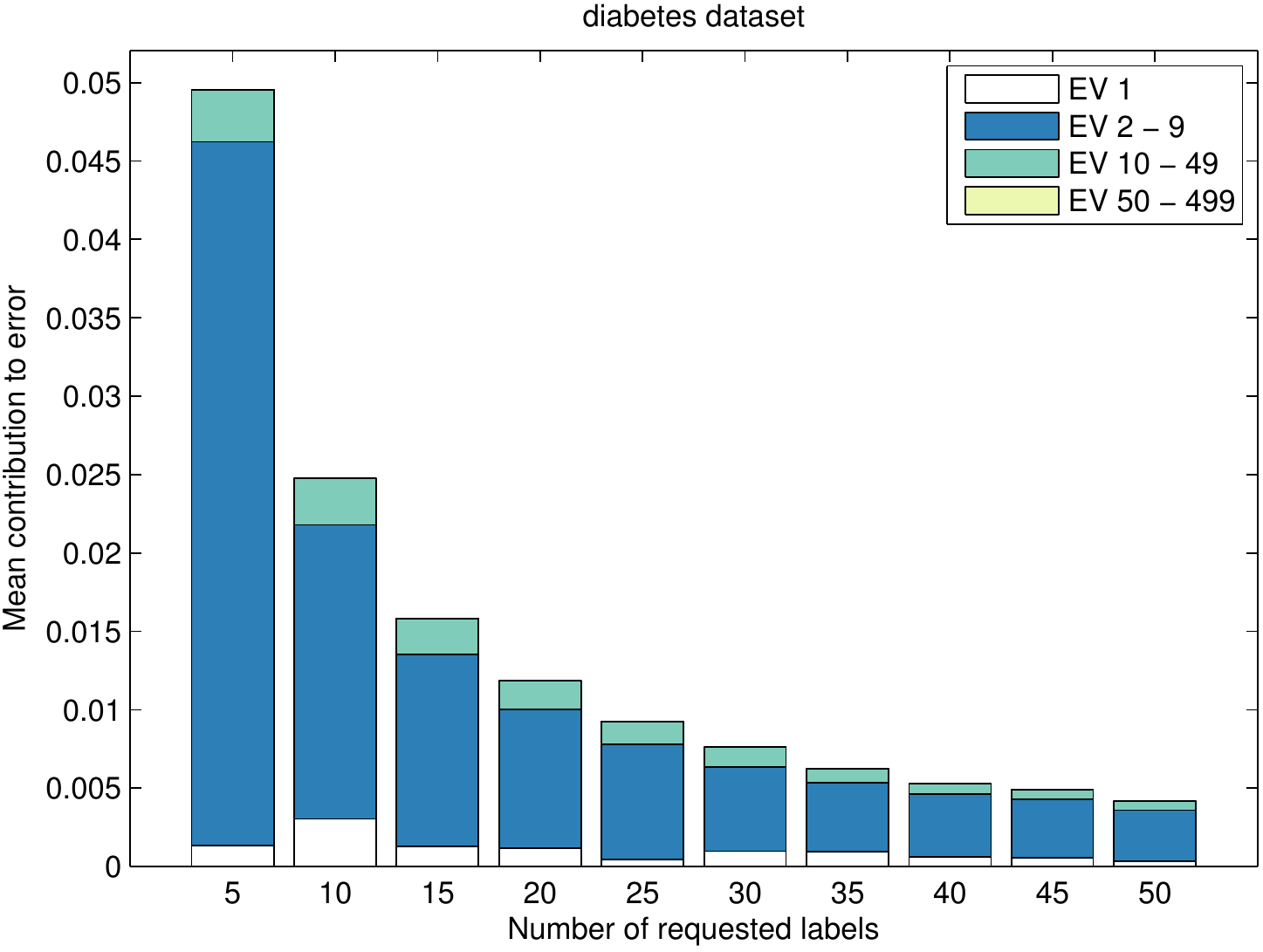}
    \end{subfigure}
		~
		\begin{subfigure}[b]{0.3\textwidth}
        \includegraphics[width=\textwidth]{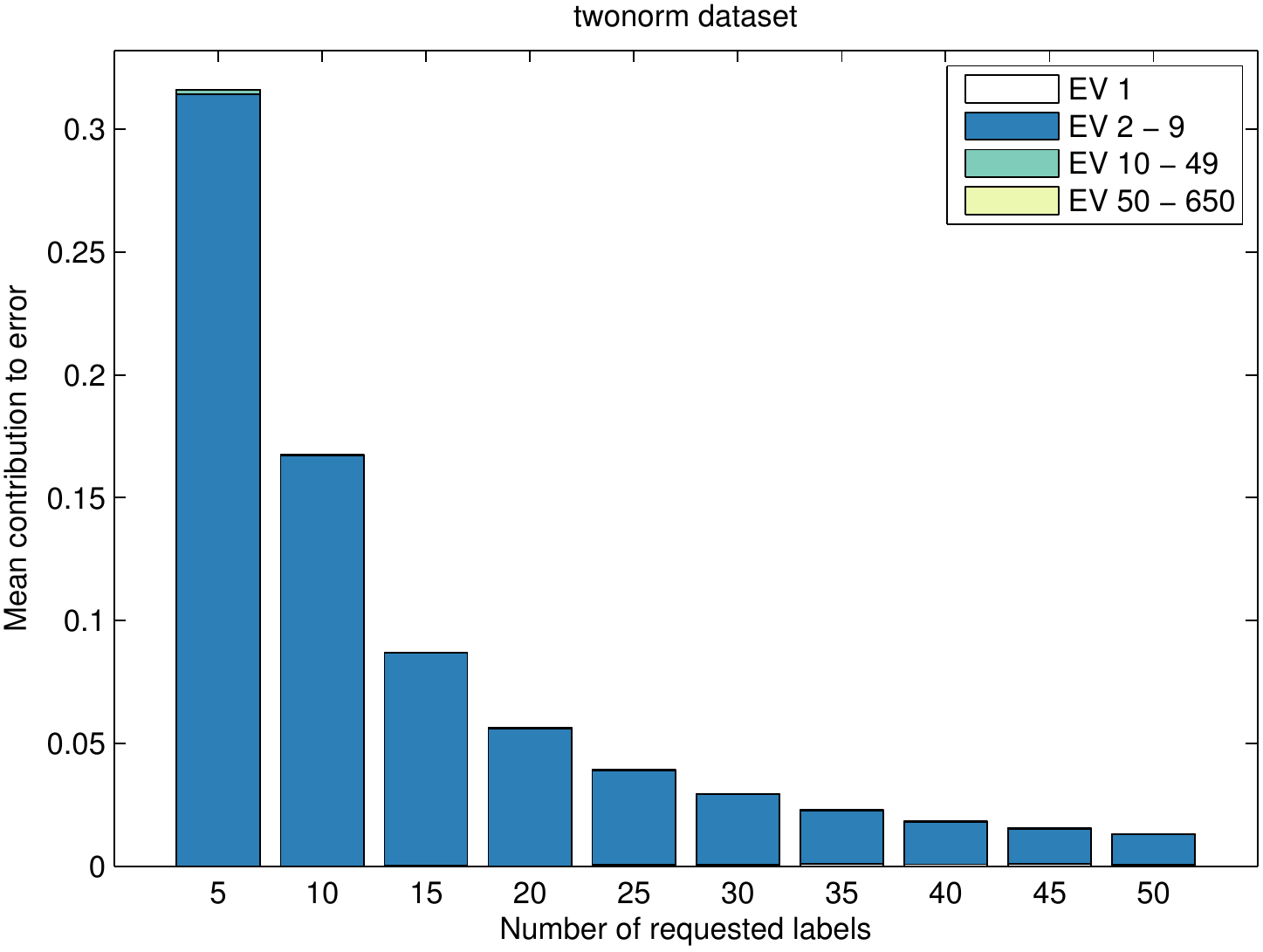}
    \end{subfigure}
		~
		\begin{subfigure}[b]{0.3\textwidth}
        \includegraphics[width=\textwidth]{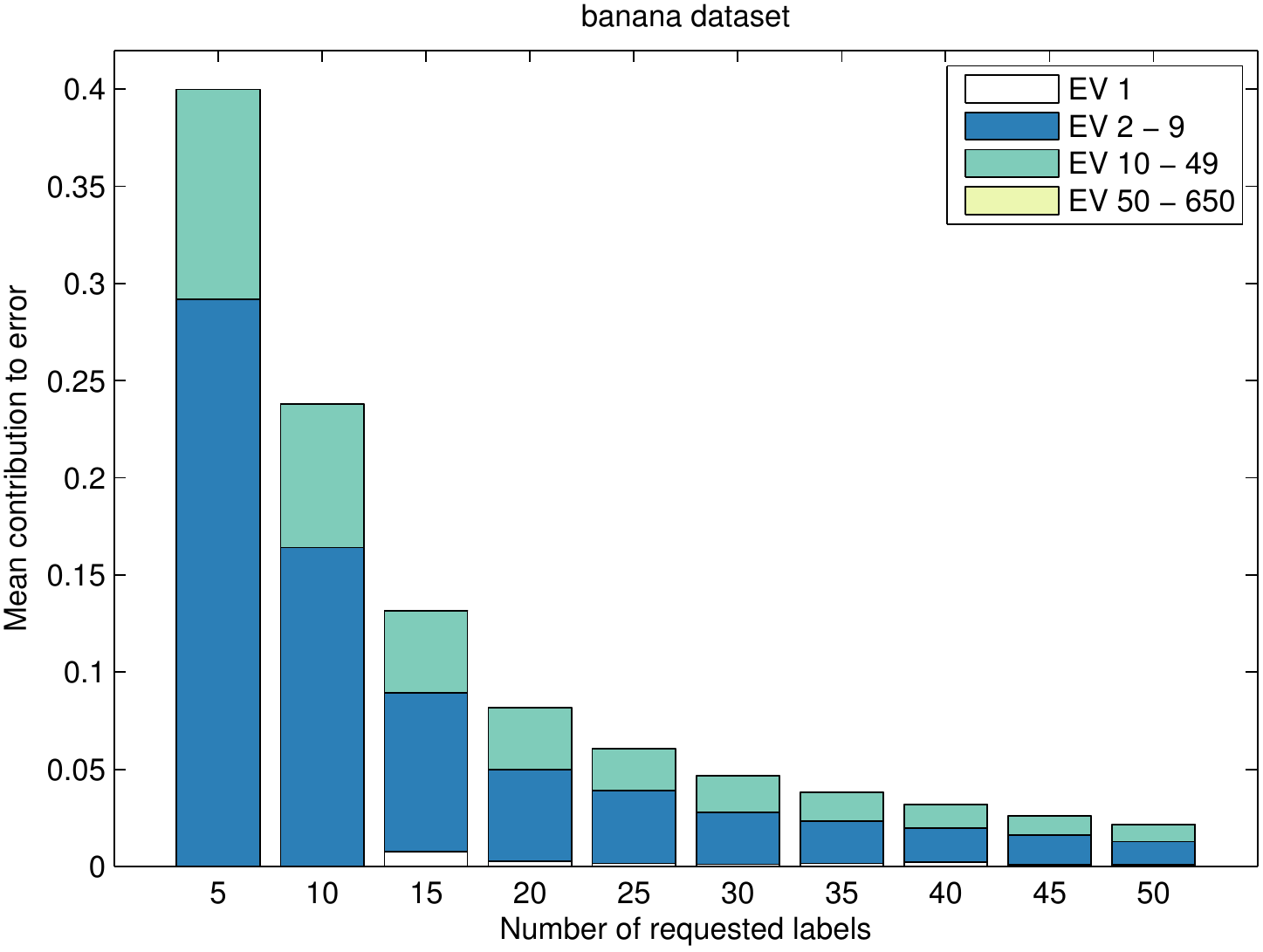}
    \end{subfigure}
		~
		\begin{subfigure}[b]{0.3\textwidth}
        \includegraphics[width=\textwidth]{staaf_new/contribution_13_Random_legend.pdf}
    \end{subfigure}
		~
		\begin{subfigure}[b]{0.3\textwidth}
        \includegraphics[width=\textwidth]{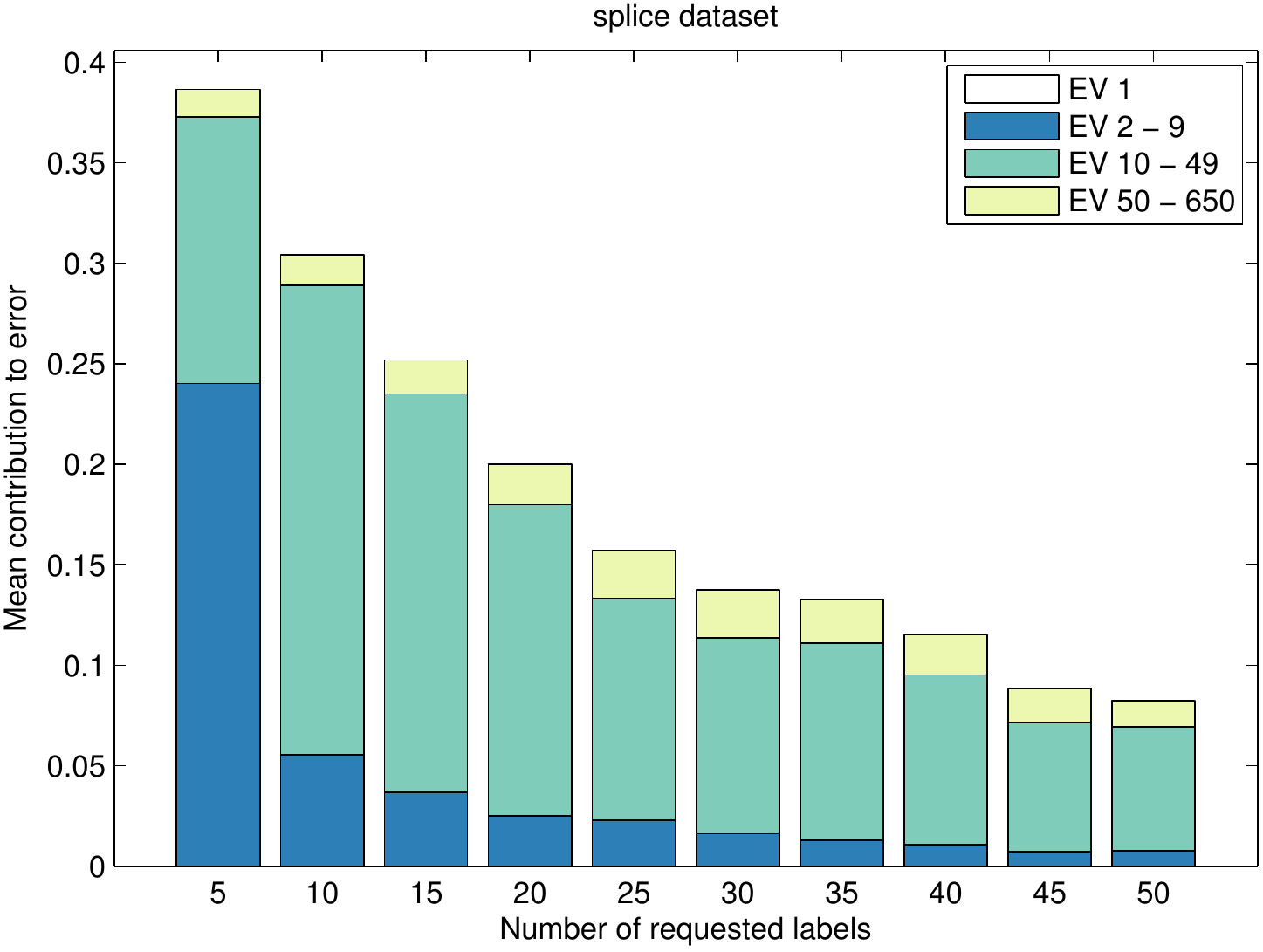}
    \end{subfigure}
		~
		\begin{subfigure}[b]{0.3\textwidth}
        \includegraphics[width=\textwidth]{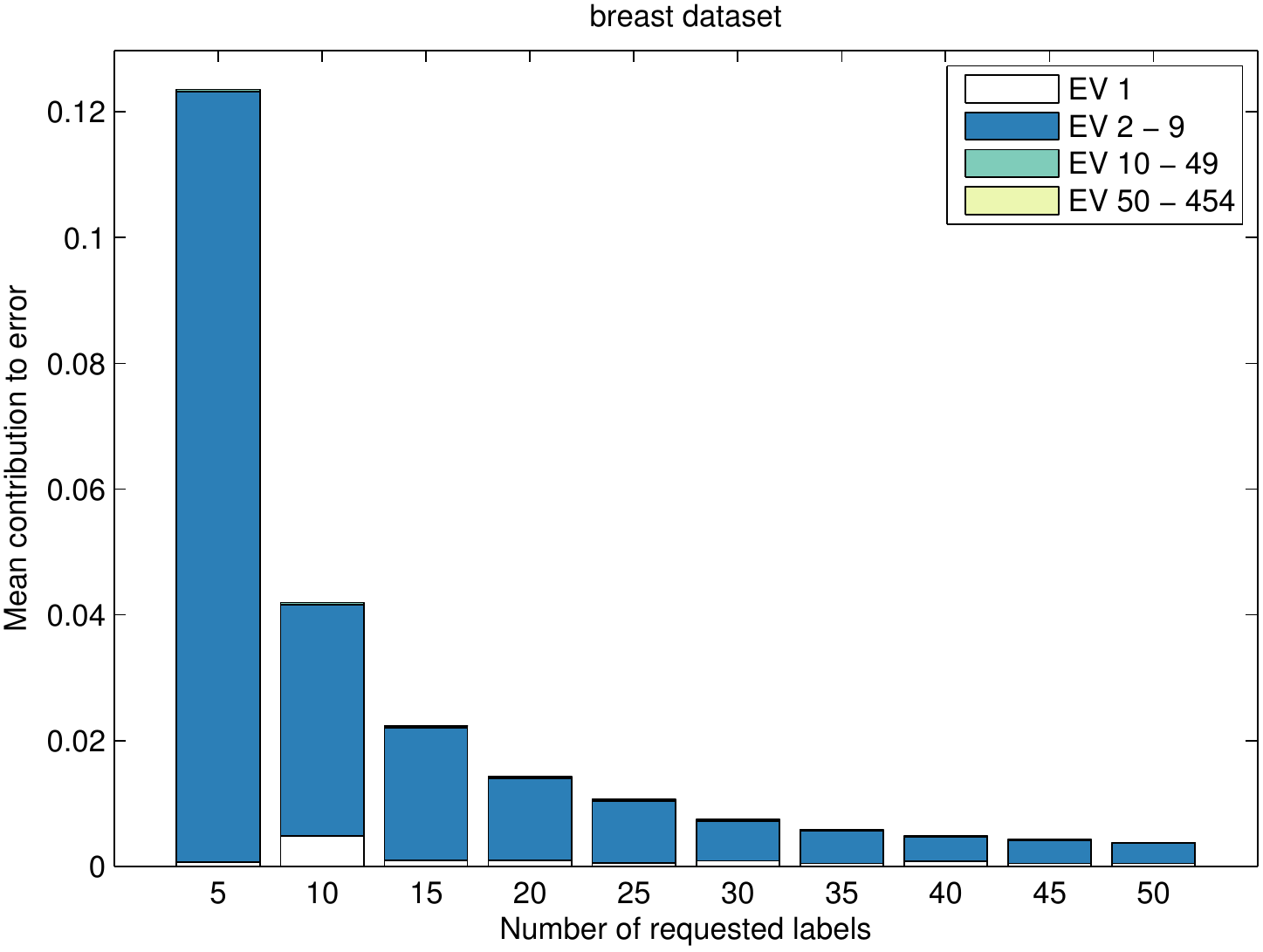}
    \end{subfigure}
    \caption{Error decomposition for all datasets.}
\end{figure}
~
\end{supplement}

\end{document}